\documentclass[Afour,sageh,times]{sagej}

\usepackage{moreverb,url}

\usepackage[bookmarksopen,bookmarksnumbered,citecolor=black,urlcolor=black]{hyperref}
\theoremstyle{definition}
\newtheorem{problem}{Problem}
\usepackage{caption}
\usepackage{subcaption}
\usepackage{graphicx}
\usepackage{array}
\newcolumntype{L}[1]{>{\centering\arraybackslash}m{#1}}

\usepackage{multirow}
\usepackage{threeparttable}
\usepackage{amsmath}
\usepackage[ruled, linesnumbered, vlined]{algorithm2e}
\newcommand{\forcond}{}
\SetArgSty{textnormal}
\newcommand{\llIf}[2]{{\let\par\relax\lIf{#1}{#2}}}
\newcommand{\llElse}[1]{{\let\par\relax\lElse{#1}}}

\newcommand\BibTeX{{\rmfamily B\kern-.05em \textsc{i\kern-.025em b}\kern-.08em
T\kern-.1667em\lower.7ex\hbox{E}\kern-.125emX}}

\def\volumeyear{2019}

\begin{document}

\runninghead{Dai et al.}

\title{Fast-reactive probabilistic motion planning for high-dimensional robots}

\author{Siyu Dai, Andreas Hofmann and Brian C. Williams}

\affiliation{Computer Science and Artificial Intelligence Laboratory, Massachusetts Institute of Technology, USA}

\corrauth{Siyu Dai, 
32 Vassar Street,
Cambridge,
MA,
02139, USA.}

\email{sylviad@mit.edu}

\begin{abstract}

Many real-world robotic operations that involve high-dimensional humanoid robots require fast-reaction to plan disturbances and probabilistic guarantees over collision risks, whereas most probabilistic motion planning approaches developed for car-like robots can not be directly applied to high-dimensional robots. In this paper, we present \emph{probabilistic Chekov (p-Chekov)}, a fast-reactive motion planning system that can provide safety guarantees for high-dimensional robots suffering from process noises and observation noises. Leveraging recent advances in machine learning as well as our previous work in deterministic motion planning that integrated trajectory optimization into a sparse roadmap framework, p-Chekov demonstrates its superiority in terms of collision avoidance ability and planning speed in high-dimensional robotic motion planning tasks in complex environments without the convexification of obstacles. Comprehensive theoretical and empirical analysis provided in this paper shows that p-Chekov can effectively satisfy user-specified chance constraints over collision risk in practical robotic manipulation tasks.

\end{abstract}


\maketitle

\section{Introduction}

Robotic systems deployed in the real world have to contend with a variety of challenges: wheels slip for mobile robots, lidars do not reflect off glass doors, currents and turbulence disturb underwater vehicles, and humans in the environment move in unpredictable manners. However, many state-of-the-art robots, with inevitable uncertainties from various sources including approximate models of system dynamics, imperfect sensors, and stochastic motions caused by controller noise, are not yet ready to handle these challenges. Although nowadays feedback controllers can take care of a large portion of uncertainties during the execution phase, the remaining deviations can still be problematic, especially for robots operating in hazardous environments or systems that collaborate closely with humans. One representative example is a manipulator mounted on an underwater vehicle, which faces not only the disturbances from currents and inner waves, but also the base movements caused by the interaction between manipulators and the vehicle on which they are mounted. A collision accident of such manipulators deployed in underwater scientific exploration tasks can often cost millions of dollars. Another typical example is a domestic assistive robot surrounded by elder people and children, which needs to be very careful about collision avoidance. Therefore, in those tasks, it is important that the motion planner can take uncertainties into account and can react quickly to plan interruptions.

Fast-reactive risk-aware motion planning for high-dimensional robots like humanoid robots, however, is a very challenging task. Unlike car-like robots, a typical robotic manipulator can have seven degrees-of-freedom (DOFs), and this high-dimensionality makes it extremely difficult to quantify uncertainties into collision risks and to make safe motion plans in real time. Existing systems that tackle the risk-aware motion planning problem~\citep{van2012motion, luders2013robust, ono2013probabilistic, sun2015high, chen2017motion, axelrod2018provably, luo2019importance} lack the ability of efficiently handling high-dimensional robots and non-convex environments. In order to address these difficulties, we propose \emph{probabilistic Chekov (p-Chekov)}, a combined sampling-based and optimization-based approach that takes advantage of the fact that most obstacles in a lot of practical motion planning tasks are static and only a small number of objects are dynamic during deployment. In these cases, we can construct sparse roadmaps based on our prior knowledge about the static environment to cache feasible trajectories offline, so that during plan execution, we only need to optimize solution trajectories according to new observations~\citep{dai2018improving, orton2019improving} and adjust plans to satisfy safety requirements~\citep{dai2019chance}. Combining ideas from risk allocation~\citep{ono2008efficient, ono2008iterative} and supervised learning, p-Chekov can effectively reason over uncertainties and provide motion plans that satisfy constraints over the probability of plan failure, i.e. \emph{chance constraints}~\citep{ono2008iterative}.

In this paper, we first provide a comprehensive review of the relevant literature and distinguish our approach in four different aspects, and then describe the problem formulation and test environments. Then in Section~\ref{d-Chekov}, we provide extensive empirical results to demonstrate that our deterministic Chekov approach can overcome the shortcomings of both sampling-based planners and optimization-based planners and achieve fast-reaction for high-dimensional motion planning problems in practical environments. This deterministic Chekov approach forms a core component in p-Chekov that generates nominal trajectories in real-time, and lays an essential foundation for the fast-reaction of the risk-aware planning approach. Section~\ref{p-chekov} illustrates the main technical components in p-Chekov planner, including the estimation of robot state probability distributions during execution, two different approaches for collision probability estimation given robot state distributions, and the allocation and reallocation of risk bounds during planning phase and execution phase. Section~\ref{results} then demonstrates the performance of p-Chekov empirically and compares the performance of the two collision probability estimation approaches. Finally we summarize the main contributions of p-Chekov and discuss potential directions for future research.

We have previously presented some components of deterministic Chekov~\citep{dai2018improving} and p-Chekov~\citep{dai2019chance}. The main contributions of this paper in addition to our previous publications include: 1) a more detailed theoretical and empirical analysis on the \emph{deterministic Checkov} approach proposed by~\cite{dai2018improving} and the \emph{quadrature-based p-Chekov} approach proposed by~\cite{dai2019chance}; 2) an \emph{iterative risk allocation (IRA)} approach for plan improvement during the execution phase of p-Chekov, and its comparison with the planning phase \emph{risk reallocation} approach; 3) an analysis on the performance of different machine learning algorithms for estimating trajectory collision risks; 4) a \emph{learning-based p-Chekov} approach that can overcome quadrature-based p-Chekov's limitations in planning speed and achieve fast-reaction as well as high chance constraint satisfaction rate for real-world high-dimensional robotic motion planning tasks. To the author's best knowledge, learning-based p-Chekov is the first motion planning and execution system that can provide chance-constrained motion plans for high-dimensional robots in complex environments in real time.

\section{Related work}

\subsection{Fast-reactive motion planning}

Approaches for robotic motion planning usually fall into three categories: search-based \citep{stentz1994optimal, koenig2005fast, cohen2010search}, sampling-based \citep{lavalle1998rapidly, bohlin2000path, karaman2011sampling} and optimization-based \citep{kalakrishnan2011stomp, zucker2013chomp, schulman2014motion}. A typical way search-based (A* like) motion planners formulate their algorithms is through discretizing the configuration space into grids and applying search algorithms to find a valid trajectory from the start to the goal. Despite that search-based motion planners can guarantee completeness and optimality, the discretization of the configuration space means the computational cost could be very high for complicated high-dimensional planning tasks. 

Sampling-based planning is another powerful approach that randomly explores a subset of the configuration space (C-space) while keeping track of the search progress. Although sampling-based planners, e.g. rapidly exploring random trees (RRTs), are able to solve some difficult motion planning tasks with the guarantee of probabilistic completeness, their performance in complicated high-dimensional planning tasks is highly restricted by the selection of sampled nodes, and their planning time is often a major concern. In contrast, optimization-based motion planning shows its advantage in planning speed because they operate on the space of trajectories and conduct a fast but local search instead of a global search. However, this also means that the performance of optimization-based planners, especially numerical trajectory optimizers which often suffer from the problem of getting stuck in high-cost local optima, can be very sensitive to the quality of the initial seed trajectory, and deeply infeasible initializations can often cause plan failures. Probabilistic inference has also been applied to robotic motion planning~\citep{mukadam2018continuous}, but similar to optimization-based approaches, inference-based planners are also very sensitive to initializations.

One way to achieve fast motion planning with high success rate is to combine optimization-based motion planners with offline global planners, such as sampling-based probabilistic roadmaps, which can provide optimization-based planners high-quality initializations with the help of offline pre-computation. The approaches \cite{luna2013anytime} and \cite{campana2015simple} proposed can conduct online path shortening for plans generated by sampling-based planners, but the effect of trajectory optimization in their approaches is limited to trajectory smoothing and shortening, whereas real-time obstacle avoidance and differential constraints were not incorporated. \cite{park2015parallel} presented a combined roadmap and trajectory optimization planning algorithm. However, their additional focus on avoiding singularities in redundant manipulators and meeting Cartesian constraints resulted in relatively long planning times.

In this paper, we propose a fast-reactive motion planning frame work for high-dimensional robots that combines obstacle-aware trajectory optimization with sparse probabilistic roadmaps in the C-space. Sparse global roapmaps are the core to fast reaction and can provide motion plans that are guaranteed to be collision-free, while obstacle-aware local optimization helps smoothen and shorten the trajectories without introducing collisions. Covariance Hamiltonian Optimization for Motion Planning (CHOMP)~\citep{zucker2013chomp}, Stochastic Trajectory Optimization for Motion Planning (STOMP)~\citep{kalakrishnan2011stomp}, Incremental Trajectory Optimization for Real-time Replanning (ITOMP)~\citep{park2012itomp} and TrajOpt~\citep{schulman2014motion} are four state-of-the-art trajectory optimization approaches. Our proposed framework could in theory work with any obstacle-aware trajectory optimizer, but in this paper we demonstrate its performance with TrajOpt out of three considerations. First, the convex-convex collision checking method used in TrajOpt can take accurate object geometry into consideration, shaping the objective to enhance the ability of getting trajectories out of collision. In contrast, the distance field method used in CHOMP and STOMP consider the collision cost for each exterior point on a robot, which means two points might drive the objective in opposite direction. Second, the sequential quadratic programming method used in TrajOpt can better handle deeply infeasible initial trajectories than the commonly used gradient descent method~\citep{schulman2013finding}. Third, customized differential constraints, such as velocity constraints and torque constraints, can be incorporated in TrajOpt. This is an important consideration for the p-Chekhov system presented in this paper which aims at building a motion execution system that incorporates system dynamics models and control policies while respecting additional temporal constraints.

\subsection{Chance-constrained motion planning}

Existing motion planners that take uncertainties into consideration include two classes: some are safety-driven and provide motion plans that minimize the collision risks~\citep{van2011lqg, patil2015scaling, patil2014gaussian, xiao2020robot}, and others, also called \emph{chance-constrained motion planners}~\citep{ono2013probabilistic}, seek the optimal plans that can satisfy a user-specified constraint over the probability of collision. In this paper, we focus on providing chance-constrained motion plans for high-dimensional robots in real time.
Many uncertainty-aware motion planners are based on Markov Decision Processes (MDPs)~\citep{thrun2005probabilistic, burlet2004robust, alterovitz2007stochastic}, and an extension of MDP, Partially Observable MDP (POMDP), is often applied to address the sensing uncertainties in robotic motion planning tasks~\citep{kurniawati2008sarsop, van2012motion, luo2019importance}. Despite their wide application, most of them require discretization of the state space. Even for extensions that can handle continuous planning domains, tractability is still a common issue due to the need of partitioning or approximation of the continuous state space~\citep{ono2013probabilistic}. 

Another class of probabilistic planners formulates motion planning into an optimization problem through approaches such as Disjunctive Linear Program (DLP). \cite{blackmore2006probabilistic} introduced a DLP-based approach that can perform obstacle avoidance under uncertainties, \cite{blackmore2010probabilistic} described a Mixed Integer Linear Programming (MILP) formulation of the robust path planning problem which approximates chance constraints with a probabilistic particle-control approach, \cite{ono2013probabilistic} proposed the probabilistic Sulu planner (p-Sulu) which performs goal-directed planning in a continuous domain with temporal and chance constraints, and \cite{lee2013sigma} adopted trajectory optimization in belief space and formulated collision avoidance constraints using sigma hulls. However, since p-Sulu encodes feasible regions with linear constraint approximations, it inevitably suffers from the exponential growth of computation complexity when applied in complicated 3D environments or tasks with multiple agents. Additionally, both linear approximations and sigma hulls place restrictions on robot and environment geometry and also introduce inaccuracies in collision probability estimation.

Uncertainty-aware extensions of search-based~\citep{lenz2015heuristic, chen2017motion} and sampling-based~\citep{luders2010chance, bry2011rapidly, luders2013robust, liu2014incremental, sun2015high} planners are also popular in the motion and path planning field. However, their applications are often limited to car-like robots in simplified environments due to their disadvantages in planning speed and collision probability estimation ability for high-DOF robots in real-world complex environments. When the robot has high dimensionality, the collision checking happens in the 3D workspace, whereas the motion planning happens in the high-dimensional C-space. Mapping the collision-free workspace into the C-space is nontrivial, which hence becomes another barrier for high-dimensional risk-aware motion planning. 


\subsection{Collision risk estimation}

The estimation of trajectory collision probability has been widely investigated in the motion planning field, yet no perfect solution has been proposed due to its inherent difficulties. In order to approach this problem, many approximations have been used, including the discretization of time and the convexification of obstacles. For low-dimensional planning tasks in convex environments, the estimation of collision risks at discrete waypoints is relatively straightforward. In the p-Sulu planner presented by~\cite{ono2013probabilistic}, each boundary of each obstacle is formulated into a linear constraint, and the half-spaces that represent those linear constraints form the collision-free regions. In this way, the waypoint collision probability becomes the probability of violating any of the linear constraints, and can be solved through linear program (LP) solvers. A very different idea is to take advantage of confidence intervals, which are ellipses and ellipsoids for Gaussian distributions~\citep{van2011lqg}. If the configuration space is 2D, then the maximum factor by which the elliptical confidence interval can be scaled before it intersects obstacles gives an indication of the collision probability at that configuration, where the scale factor can be computed as the Euclidean distance to the nearest obstacle in the environment. \cite{patil2012estimating} further investigate this idea and account for the fact that the collision probability at each step along a trajectory is conditioned on the previous steps being collision-free. They propose that the \emph{a priori} state probability distributions for different waypoints along a trajectory can be truncated to better reflect the actual collision probabilities.

However, it is nontrivial to extend the aforementioned approaches to high-dimensional planning tasks. Obstacles defined in workspace can not be directly mapped into a 6-DOF or 7-DOF C-space in closed form~\citep{choset2005principles}, hence the feasible region idea \citep{ono2013probabilistic} and the confidence interval scaling idea \citep{van2011lqg} can not be easily applied. \cite{sun2016safe} pointed out a key relation between workspace geometry and C-space geometry: configuration $\mathbf{q}$ lies on the boundary of a C-space obstacle if and only if the workspace distance between the obstacle and the robot configured at $\mathbf{q}$ is zero. Based on this relation, \cite{sun2016safe} proposed an approach that looks for the point on the boundary of C-space obstacles that is closest to the robot's mean configuration by calculating the gradient of the workspace signed-distance field. Although this approach builds an important bridge between workspace obstacles and C-space obstacles, it relies on the assumption that the geometries of the C-space obstacles are locally convex. Since p-Chekov aims at solving high-dimensional motion planning problems in 3D complex environments where obstacles maintain their original non-convex shapes, we explore two different ideas for estimating waypoint collision risks that overcome the limitations of the aforementioned methods: one relies on a quadrature-based sampling method to mitigate the inaccuracy of random sampling and to avoid the difficulty of mapping between C-space and workspace (quadrature-based p-Chekov), and the other leverages regression methods with function approximators to learn risk distributions through offline sampling and to make predictions during online planning queries (learning-based p-Chekov).

\subsection{Machine learning in motion planning}

Machine learning approaches are still not widely applied in robotic motion planning. Existing applications include guiding the exploration of sampling-based motion planners using nearest neighbor and adaptive sampling~\citep{atramentov2002efficient, arslan2015machine, ichter2018learning}, accelerating collision detection through supervised classification~\citep{pan2016fast, pan2017probabilistic}, and pursuing end-to-end motion planning through learning from demonstration~\citep{wang2016motion, pfeiffer2017perception, ha2018approximate}. To the author's best knowledge, this paper is the first application of learning-based methods on the collision risk estimation problem for probabilistic motion planning systems. We explore the real-time collision risk estimation performance of different machine learning algorithms with different structures in chance-constrained motion planning tasks for robotic manipulators. It is shown that neural networks with appropriate structures can efficiently generate accurate predictions on collision risks in the environments they are trained in. The experiment results in this paper show that p-Chekov with neural networks as collision risk estimation component performs significantly better than the quadrature-based p-Chekov in terms of planning speed.

\section{Problem statement} 

We define a \textit{disturbance} as an unexpected change to task goals, environment, or robot state. It may be due to an actual physical change, or a change in the estimated state of the environment or robot. Here we distinguish between \emph{severe disturbances} and \emph{small disturbances}. Severe disturbances refer to the ones that will cause significant and qualitative plan changes, such as changes of the planning goal, the movement of some obstacles that obstructs the original plan, or a strong external force that results in large deviations from the desired trajectory and the feedback controllers can't get the robot back on track due to actuation limits. On the other hand, small disturbances are mainly caused by process noises and observation noises, and the control inputs within limit can get the robot back to the desired trajectory. In practice, motion planners should account for the risk of potential plan failure caused by small disturbances, and react fast and naturally to severe disturbances which would necessitate plan adjustment. 

P-Chekov solves robotic motion planning problems under uncertainty with a guaranteed success probability, considering temporal, spatial and dynamical constraints. Under process and observation noises, the collision rate during plan executions should not exceed a user specified chance constraint. The resulting motions should be locally optimal or near-optimal according to a specified objective function, which may optimize a variety of characteristics such as path length or control effort. P-Chekov's real-time planning feature is key to providing robots the capability of operating effectively in unstructured and uncertain environments.

\subsection{Model definition} \label{model}

Let $\mathcal{X} = \mathbb{R}^{n_x}$ denote the robot \emph{state space} and $\mathcal{U} = \mathbb{R}^{n_u}$ the system \emph{control input space}, where $n_x$ and $n_u$ are the dimensions of the state space and the control input space respectively. Consider a discretized series of time steps $t = 0, 1, 2, \ldots, T$ with a fixed time interval $\Delta T$, where the number of time steps $T$ is a finite integer. Let $\mathbf{x}_t \in \mathcal{X}$ denote the robot state at time step $t$. We assume applying a control input $\mathbf{u}_t \in \mathcal{U}$ at time step $t$ will bring the robot from state $\mathbf{x}_t \in \mathcal{X}$ to $\mathbf{x}_{t+1} \in \mathcal{X}$, according to a given stochastic dynamics model: 

\begin{equation}
\begin{aligned}
& \mathbf{x}_t = f(\mathbf{x}_{t-1}, \mathbf{u}_{t-1}, \mathbf{m}_t), & & \mathbf{m}_t \sim \mathcal{N}(0, M_t),
\end{aligned}
\end{equation}

\noindent where $\mathbf{m}_t$ is the zero-mean Gaussian distributed process noise at time step $t$ with a given covariance matrix $M_t$. $\mathbf{m}_t$ can be modeled based on the prior knowledge about robot controllers. Function $f$ governs the robot dynamics and is assumed to be either linear or can be well approximated locally by its linearization.

The robot states are observed by taking a measurement at each time step $t$, denoted as $\mathbf{z}_t$. We assume that measurements are provided by noisy sensors according to a stochastic observation model:

\begin{equation}
\begin{aligned}
& \mathbf{z}_t = h(\mathbf{x}_t, \mathbf{n}_t), & & \mathbf{n}_t \sim \mathcal{N}(0, N_t),
\end{aligned}
\end{equation}

\noindent where $\mathbf{n}_t$ is the zero-mean Gaussian distributed observation noise at time step $t$ with a given covariance matrix $N_t$.

For each specific planning task, a start state $\mathbf{x}^\textrm{start}$ and a goal state $\mathbf{x}^\textrm{goal}$ or a convex goal region $\mathcal{X}^\textrm{goal}$ will be given. Let $\mathbf{x}_0 \in \mathcal{X}$ denote the initial state of the robot that follows a Gaussian distribution with mean $\mathbf{x}^\textrm{start}$ and covariance matrix $\mathbf{\Sigma}_{\mathbf{x}_0}$:

\begin{equation}
\mathbf{x}_0 \sim \mathcal{N}(\mathbf{x}^\textrm{start}, \mathbf{\Sigma}_{\mathbf{x}_0}).
\end{equation}

\noindent An initial condition is defined as a combination of $\mathbf{x}^\textrm{start}$ and $\mathbf{\Sigma}_{\mathbf{x}_0}$. A trajectory $\Pi$ is defined as a sequence of nominal robot states and control inputs $(\mathbf{x}^*_0, \mathbf{u}^*_0, \ldots, \mathbf{x}^*_T)$ that satisfies the deterministic dynamics model $\mathbf{x}^*_t = f(\mathbf{x}^*_{t-1}, \mathbf{u}^*_{t-1}, 0)$ for $0 < t \leq T$. We assume that an objective function $J(\Pi)$ will be specified for each planning task, which can implement planning goals such as minimizing trajectory length.

\subsection{Constraint definitions}

A valid solution provided by p-Chekov should satisfy temporal constraints, chance constraints over collision risks, goal state constraints, control input constraints, and system dynamics constraints specified by the robot model. A temporal constraint defines an upper bound $\tau$ on the execution duration of a trajectory: 

\begin{equation}
 T \times \Delta T \leq \tau.
\end{equation}

We assume a joint collision chance constraint with bound $\Delta_c \in [0, 1]$ will be given for each planning task, which specifies the allowed probability of collision failure. Let $C_i$ denote the no-collision constraint for each obstacle $i = 1, \ldots, N$, then the probability of colliding with obstacle $i$ is $P(\overline{C_i})$. The collision chance constraint over an entire trajectory can then be expressed as:

\begin{equation}
 P\Bigg(\bigvee^N_{i=1} \overline{C_i}\Bigg) \leq \Delta_c.
\end{equation}

The control input constraint requires that $\mathbf{u}^*_t \in \mathcal{U}, \forall t = 1, \ldots, T$. The system dynamics constraints require that the robot states at each time step along the trajectory are within the robot state space $\mathcal{X}$, and the state transitions between adjacent time steps satisfy the deterministic system dynamics model:

\begin{equation}
 \mathbf{x}^*_t = f(\mathbf{x}^*_{t-1}, \mathbf{u}^*_{t-1}, 0) \in \mathcal{X}, ~~~ \forall t = 1, \ldots, T.
\end{equation}

\subsection{Problem definition}

Problem \ref{risk-aware-optimization-problem} defines the constrained optimization problem solved by p-Chekov. It aims at finding a feasible trajectory $\Pi$ that minimizes the given objective $J(\Pi)$ while satisfying the chance constraint and temporal constraint. The solution trajectory $\Pi$ should satisfy the initial condition and the robot dynamics model, and the control inputs along the trajectory should fall into the control input space. If a C-space goal pose $\mathbf{x}^\textrm{goal}$ is given, the robot configuration at the final time step should be at $\mathbf{x}^\textrm{goal}$; on the other hand, if a convex goal region of the workspace end-effector pose $\mathcal{X}^\textrm{goal}$ is specified, then the end-effector should be in $\mathcal{X}^\textrm{goal}$ at the end of $\Pi$.

\begin{problem}
\label{risk-aware-optimization-problem}
 \begin{equation}
\begin{aligned}
& \underset{\Pi}{\textrm{minimize}} & & J(\Pi) \\
& \textrm{subject to} & & \mathbf{x}_0 \sim \mathcal{N}(\hat{\mathbf{x}}_0, \mathbf{\Sigma}_{\mathbf{x}_0}) \\
& & & \mathbf{x}_t = f(\mathbf{x}_{t-1}, \mathbf{u}_{t-1}, \mathbf{m}_t), & 0 < t \leq T \\
& & & \mathbf{z}_t = h(\mathbf{x}_t, \mathbf{n}_t), & 0 < t \leq T  \\
& & & \mathbf{m}_t \sim \mathcal{N}(0, M_t), & 0 < t \leq T  \\
& & & \mathbf{n}_t \sim \mathcal{N}(0, N_t), & 0 < t \leq T  \\
& & & \mathbf{x}_t \in \mathcal{X}, & 0 < t \leq T \\
& & & \mathbf{u}_t \in \mathcal{U}, & 0 < t \leq T \\
& & & \mathbf{x}^*_T = \mathbf{x}^\textrm{goal} ~~ \textrm{or} ~~ \mathbf{x}^*_T \in \mathcal{X}^\textrm{goal} \\
& & & P\Bigg(\bigvee^N_{i=1} \overline{C_i}\Bigg) \leq \Delta_c \\
& & & T \times \Delta T \leq \tau \\
\end{aligned}
\end{equation}
\end{problem}
%
%

\section{Experiment setup} \label{implementation}

\begin{figure*}[t]
      \centering
      \caption{Simulation Environments for Motion Planner Evaluation Experiments}
      \label{fig:environments}
      \begin{subfigure}[t]{0.25\textwidth}
      \includegraphics[width=0.95\linewidth]{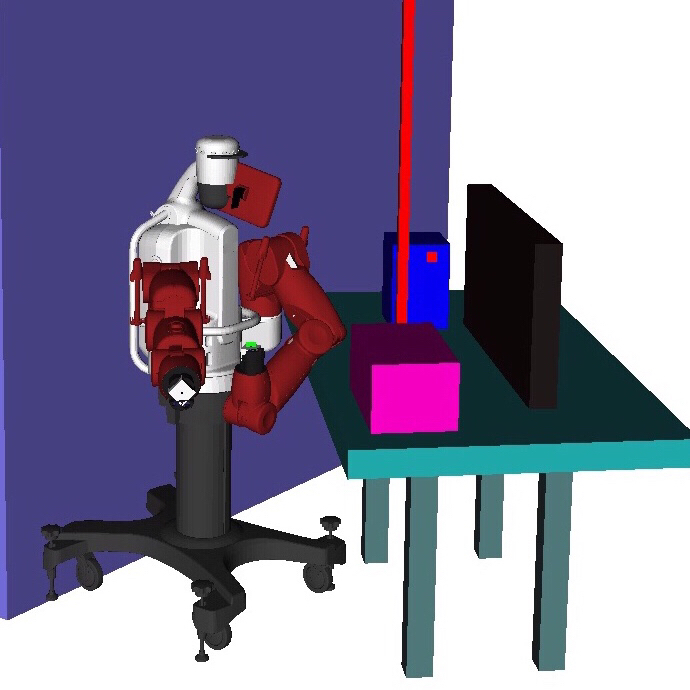}
      \caption{The ``tabletop with a pole'' environment}
      \end{subfigure}%
      \begin{subfigure}[t]{0.25\textwidth}
      \includegraphics[width=0.95\linewidth]{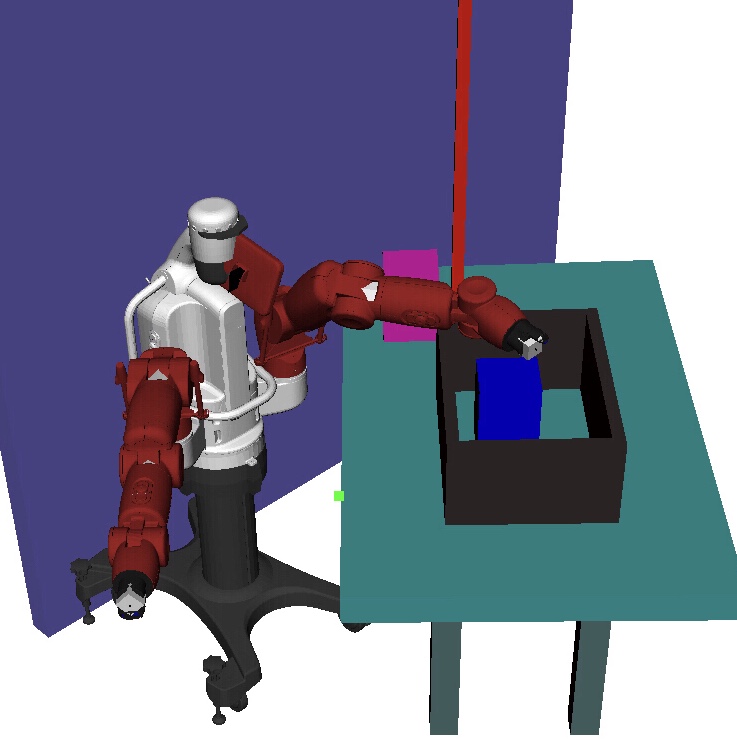}
      \caption{The ``tabletop with a container'' environment}
      \end{subfigure}%
     \begin{subfigure}[t]{0.25\textwidth}
      \includegraphics[width=0.95\linewidth]{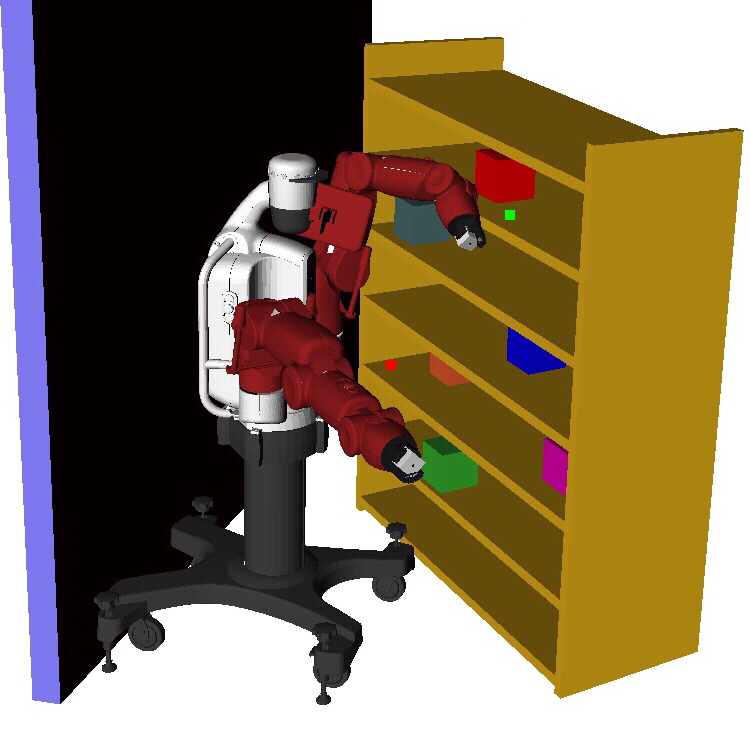}
      \caption{The ``shelf with boxes'' environment}
      \end{subfigure}%
      \begin{subfigure}[t]{0.25\textwidth}
      \includegraphics[width=0.95\linewidth]{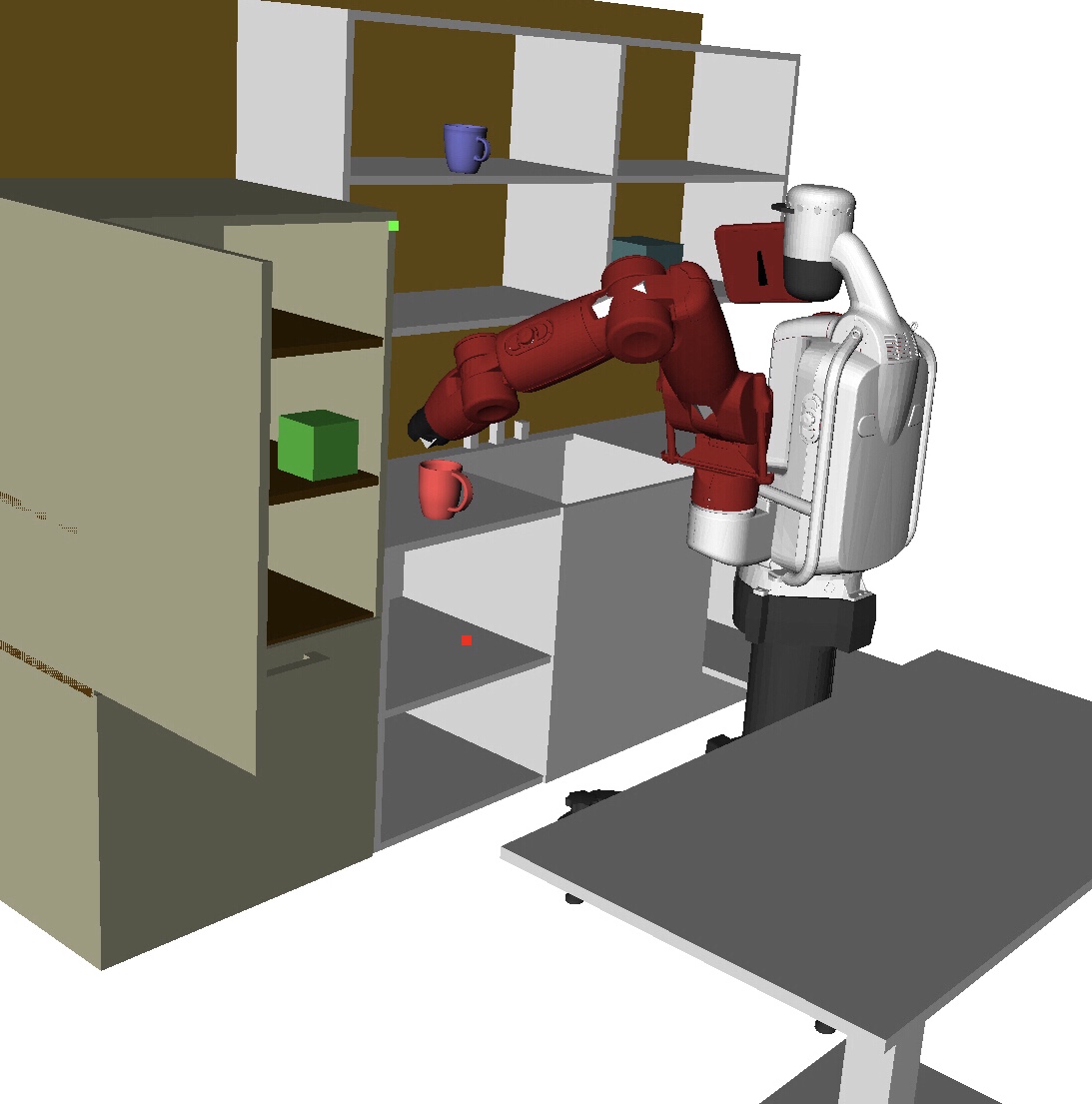}
      \caption{The ``kitchen'' environment}
      \end{subfigure}
   \end{figure*}

In this paper, four practical simulation environments are used in the experiments on different motion planners: a ``tabletop with a pole'' environment, a ``tabletop with a container'' environment, a ``shelf with boxes'' environment and a ``kitchen'' environment~\citep{dai2018improving}, as shown in Figure~\ref{fig:environments}. We choose environments that are representatives of different application domains rather than using an environment with randomly-placed obstacles because our goal is to develop a motion planner that operates quickly and  provides short paths for real world applications. The kitchen environment is adapted from the TrajOpt package, whereas, we designed the remaining three. The ``tabletop with a pole'' environment, shown in Figure \ref{fig:environments}-(a), is a simple tabletop pick-and-place task environment, with a slender pole in the middle of the table and a box on each side of the pole. Empirical results show that planning queries in this environment are relatively easy for all the tested planners. The ``tabletop with a container'' environment is similar, but with a large container on the table with boxes both inside and outside of it, as shown in Figure \ref{fig:environments}-(b). The ``kitchen'' environment, shown in Figure \ref{fig:environments}-(d), models a typical kitchen scenario which is common in household domains. The ``shelf with boxes'' environment, shown in Figure \ref{fig:environments}-(c), is a 7-level shelf environment with boxes on each level of the shelf, which is a common scenario in the logistic application domain. This scenario is known to be hard for all the planners because of the relatively large total number of obstacles and the narrow space between different layers of the shelf. 

For each environment, we generate 5000 feasible planning queries by randomly sampling start and target end-effector pose pairs that are collision-free and kinematically feasible. For each experiment trial, planners are provided with the starting C-space position and the goal end-effector pose. We specify the goal in workspace to give planners the opportunity to find different C-space solutions to the planning problem. We have ensured that all test queries have a feasible solution by executing all the planners on each test case, and re-sampling start and goal poses when no planner could find a solution. The Baxter robot \citep{BaxterRobot} with its 7-DOF left manipulator is used as the experiment testbed. All the experiments shown in this paper are conducted on a 10-core Intel i7 3.0 GHz desktop with 64~GB RAM.

We noticed that the TrajOpt package also provides a ``swept-out volume'' method in addition to their discrete-time collision costs approach which only takes into account waypoint collisions, in order to ensure continuous-time safety when executing the planned trajectories~\citep{schulman2013finding}. However, during our experiments we found out that collisions can still occur on the edges between waypoints even when the continuous-time collision cost is utilized, and it is not obvious how to use TrajOpt's reported collision cost to detect collisions consistently since large cost values can indicate either an actual collision or simply the trajectory being close to some obstacles. Hence for the sake of time-efficiency, we use TrajOpt with discrete-time collision costs, and implement an independent collision checking process to evaluate continuous-time safety on solution trajectories. In particular, we interpolate 100 intermediate points between each pair of adjacent waypoints and check collisions on each of them using OpenRAVE. For our work, we assume this fine-grained discrete-time collision test can approximate continuous-time safety sufficiently well.

\section{Fast-reactive motion planning approach: deterministic Chekov} \label{d-Chekov}

Real-world robotic systems usually cannot spend an unbounded amount of time searching for an optimal motion plan -- a plan that might soon be invalidated by the next sensor reading or a slipping wheel. The problem of moving a robot safely and efficiently in uncertain environments, however, is a challenging one. In this section, we propose a roadmap-based fast-reactive motion planning approach called \emph{deterministic Chekov} in order to address this issue. We first describe the deterministic Chekov approach in Section~\ref{rm-seeds}, then present a systematic evaluation of several popular motion planners in typical manipulation scenarios in Section~\ref{limitation}, including the basic version of TrajOpt~\citep{schulman2013finding} with straight-line C-space initializations, BasicRRT from OpenRAVE, as well as LazyPRM \citep{bohlin2000path}, PRM* \citep{karaman2011sampling} and RRT* \citep{karaman2011sampling} from the Open Motion Planning Library (OMPL). The evaluation exposes issues including long planning time and high failure rate, thus in Section~\ref{roadmap} we demonstrate the performance of our approach which addresses the aforementioned issues. All experiments are conducted in the four simulation environments introduced in Section~\ref{implementation} with 5000 test queries each.



\subsection{Deterministic Chekov: the roadmap-based fast-reactive motion planner}  \label{rm-seeds}

The deterministic Chekov motion planner achieves a fast-reactive capability through constructing a sparse probabilistic roadmap and storing the all-pair-shortest-path solutions between each pair of nodes offline~\citep{dai2018improving}. The roadmap represents the static collision-free space and is re-used across planning instances. We construct very sparse probabilistic roadmaps with a small number of nodes (1000 nodes for a 7-dimensional C-space) so that the online queries can be fast. For each pair of nodes in the roadmap, $k$ shortest paths ($k\geq1$) are calculated and stored offline, so that when dynamic obstacles invalidate some of the edges in the roadmap, the probability of finding a collision-free trajectory for the planning task can be enhanced as we increase $k$. Because we hope to consider the entire solution space rather than the very sparse one provided by the roadmap, we combine this offline roadmap with an online obstacle-aware optimizer in order to improve trajectory smoothness and achieve fast reaction to disturbances. The key ideas of this fast motion planning approach are the reuse of the offline cached all-pair-shortest-path solutions of sparse roadmaps during online queries and the combination with fast obstacle-aware online trajectory optimization.

The core of the roadmap framework for deterministic Chekhov is a simplified probabilistic roadmap (PRM) variant combined with a cache of all-pair-shortest-path solutions. The roadmaps are constructed by randomly sampling points in C-space until a pre-defined number of collision-free points have been sampled. In the test environments introduced in Section~\ref{implementation}, the sampling is uniform over the four most proximal joints of the manipulator, and fixed values are assigned to the remaining joints for all nodes. Then, each node is connected to the $n$ nearest neighbors for which collision-free edges exist. This approach is taken to more completely cover the workspace with random samples in the C-space. Tests were conducted to observe the failure rates of roadmaps in different environments relative to the number of randomly sampled points in the roadmap. As the number of randomly sampled points increased, we observed significant improvement in how often the roadmap was connected to in all environments, particularly in the ``shelf with boxes'' environment. For the tests in this paper, the roadmaps start out with 1000 collision-free nodes and $n=10$ is used. The resulting roadmap is pruned of any nodes and edges disconnected from the largest subgraph. For the environments tested, no more than five of the 1000 points were disconnected from the main subgraph. Then an all-pair-shortest-path solution set is constructed for the pruned roadmap and stored for rapid online queries. During online planning, the start and goal poses are connected to the nearest nodes in the roadmap and the shortest path between these two nodes is added to the solution trajectory. This solution trajectory is then fed into the trajectory optimizer as the initialization in order to generate a smooth final solution trajectory quickly.

For the purposes of evaluating the key aspects of our approach, we have assumed that all obstacles in the test environments are static. We focus here on static rather than dynamic obstacles because static obstacles occupy the majority of the workspace in many practical applications. Dynamic obstacles could be handled through storing redundant roadmap paths and by coupling these paths with fast online obstacle-aware optimization. In addition, the incremental Chekov approach introduced by~\cite{orton2019improving} is an extension to the deterministic Chekov approach illustrated in this paper that highlights the handling of dynamic obstacles in the environment. Incorporating incremental Chekov's ability of handling dynamic obstacles into the chance-constrained motion planning framework described in this paper is a potential direction of our future research.

\subsection{Limitation of existing motion planners}  \label{limitation}

Sampling-based motion planners can operate stand-alone but are usually not fast enough for real-time high-dimensional planning tasks, and some of them (like PRM and PRM*) cannot incorporate constraints on robot dynamics. On the other hand, optimization-based motion planners locally optimize a seed trajectory and their performance is very sensitive to initializations. This section provides a systematic empirical study on four popular sampling-based planners and one optimization-based planner, TrajOpt, comparing their performance in terms of failure-rate, length of solutions and average planning time. Note that the runtime upper bound for sampling-based planners are set to 300s in this experiment so that the optimal planners (RRT* and PRM*) are provided with enough time to optimize the solutions. This also means the runtime for RRT* and PRM* will always be around 300s because optimal planners keep optimizing their solutions until timeout.

TrajOpt formulates the kinematic motion planning problem as non-convex optimization over a $T \times K$-dimensional vector, where $T$ is the number of time steps and $K$ is the number of DOFs. Every trajectory in TrajOpt consists of $T$ waypoints, where the number $T$ is set by the user. We ran 16 sets of tests, each with an increasing total number of waypoints, and observed that TrajOpt runtime increased approximately linearly with number of waypoints while the collision rate dropped quickly with more waypoints. In our tests on TrajOpt with straight-line seed trajectories, we found that setting $T=30$ provided a good balance between low collision rates and algorithm runtimes. Henceforth, in this section, we use 30 total waypoints (including the start and target waypoints).

\begin{table}
\caption{Evaluation of Existing Sampling-based and Trajectory Optimization Motion Planners}
\label{table_planners}
\small\sf\centering
\begin{threeparttable}
\begin{tabular}{L{1.4cm}|L{1.38cm} L{1.38cm} L{1.38cm} L{1.66cm}}
\toprule
Environ-ments & Planners\tnote{1} & Failure Rate & Average Runtime (s)\tnote{2} & 
Average Path Length (rad)\\
\hline
\multirow{5}{1.4cm}{\centering Tabletop with a Pole}
& RRT & 2.30\% & 17.88 & 0.77 \\ 
& LazyPRM & 0.22\% & 7.32 & 1.76 \\ 
& RRT* & 5.32\% & 300.19 & 0.63 \\ 
& PRM* & 1.00\% & 300.71 & 0.79 \\
& TrajOpt & 17.38\% & 0.56 & 0.71 \\ 
\hline
\multirow{5}{1.4cm}{\centering Tabletop with a Container}
& RRT & 19.50\% & 44.90 & 0.92 \\ 
& LazyPRM & 1.11\% & 15.04 & 1.92 \\ 
& RRT* & 0.86\% & 300.29 & 0.80 \\ 
& PRM* & 1.28\% & 300.73 & 1.04 \\ 
& TrajOpt & 35.96\% & 1.33 & 1.14 \\ 
\hline
\multirow{5}{1.4cm}{\centering Shelf with Boxes} 
& RRT & 10.00\% & 63.86 & 1.06 \\ 
& LazyPRM & 16.94\% & 63.85 & 2.08 \\ 
& RRT* & 26.78\% & 300.37 & 0.93 \\ 
& PRM* & 24.34\% & 300.79 & 1.16 \\ 
& TrajOpt & 32.06\% & 1.59 & 1.51 \\ 
\hline
\multirow{5}{1.4cm}{\centering Kitchen}
& RRT & 12.28\% & 45.95 & 0.78 \\ 
& LazyPRM & 0.85\% & 18.03 & 1.67 \\
& RRT* & 0.51\% & 300.27 & 0.71 \\ 
& PRM* & 1.33\% & 300.89 & 0.87 \\ 
& TrajOpt & 8.80\% & 0.74 & 0.94 \\ 
\bottomrule
\end{tabular}
\begin{tablenotes}
\footnotesize
 \item[1] For each planner, the data shown are averaged from 5000 planning queries in each environment.
 \item[2] The runtime upper-bound is set to 300s. RRT* and PRM* always use the full amount of time -- the small deviation from 300s shown in the table is due to small timing errors during simulation.
 \end{tablenotes}
 \end{threeparttable}
\end{table}

Table~\ref{table_planners} summarizes the performance of the five tested motion planners. The reported failure rate includes failures in finding a solution and failures in passing our independent collision test after returning a solution. From the shown failure rates we can see that in most environments, TrajOpt fails more frequently in finding collision-free solutions than other planners. The four sampling-based planners can find collision-free solutions for most of the queries in the relatively simple ``tabletop with a pole'' environment, but fails much more in the complicated ``shelf with boxes'' environment, especially the optimal planners RRT* and PRM*. Despite sampling-based planners' relatively high success rate, the ``average runtime'' column in Table~\ref{table_planners} shows their limitations. In terms of the average path length, optimal planners have noticeable advantages in finding shorter solutions, especially in more complicated environments. Among the remaining planners, LazyPRM tends to return longer solutions, which is reasonable due to the intrinsic mechanism of lazy searching algorithms. TrajOpt's performance in path length is comparable to sampling-based planners, especially in relatively easy environments. In conclusion, although sampling-based planners are good at avoiding collision, they are too slow to be applied in most practical real-time motion planning tasks. In contrast, TrajOpt shows advantage in terms of runtime, but the high collision rate makes it an unsatisfactory planner in practice.

\subsection{Deterministic Chekov performance} \label{roadmap}

Due to TrajOpt's sensitivity to initialization conditions, we propose that the performance of TrajOpt can be dramatically improved if we pass in collision-free trajectories as seeds instead of using C-space straight-line seeds. Therefore, we use the sampling-based planners from Table~\ref{table_planners} as well as the Chekov roadmap to provide initializations to TrajOpt and evaluate the performance of the combined planners. Here we consider only the queries where the sampling-based planner successfully found a collision-free solution, and evaluate TrajOpt's runtime, solution trajectory length and collision rate. 

We first demonstrate in Table~\ref{table_roadmap} the performance of the Chekov roadmap planner alone in the four environments introduced in Section~\ref{implementation}. Compared with Table~\ref{table_planners}, Table~\ref{table_roadmap} shows that our roadmap performs comparably or better than all other tested sampling-based planners in terms of failure rate. In the most difficult environment, only RRT was able to produce a solution more often than our roadmap planner. In addition to failure rate, our roadmap planner's average runtime is substantially better than other sampling-based planners' in all cases. It is faster by more than an order of magnitude in most observed cases, which is a result of caching the all-pair-shortest-path solution sets offline. For path length, the roadmap planner performs worse than the optimal planners and RRT, but better than LazyPRM. In general with roadmap-based planners, the sparsity of the roadmap restricts its ability to obtain short paths. With only 1000 nodes, we consider the roadmaps we are using to be relatively sparse for the 7D C-space. Since these paths will be used as seeds for TrajOpt and their lengths are well within an order of magnitude of one another, the discrepancies in path length are not a concern here. 

\begin{table}
\caption{Chekov Roadmap Performance in All Environments}
\label{table_roadmap}
\small\sf\centering
\begin{threeparttable}
\begin{tabular}{L{1.5cm} L{1.3cm} L{1.35cm} L{1.65cm} L{1.35cm}}
\toprule
Environ-ments\tnote{1} & Failure Rate\tnote{2} & Average Runtime (s) & Average Path Length (rad) & Best Average\tnote{3} (rad)\\
\hline
Tabletop with a Pole & 0.18\% & 0.14 & 1.24 & 0.63 \\ \hline
Tabletop with a Container & 0.76\% & 0.18 & 1.32 & 0.80 \\  \hline
Kitchen & 1.92\% & 0.38 & 1.29 & 0.71 \\ \hline
Shelf with Boxes & 12.06\% & 0.39 & 1.30 & 0.93 \\
\bottomrule
\end{tabular}
\begin{tablenotes}
\footnotesize
 \item[1] In each environment, roadmap performance is tested on 5000 planning tasks and the data shown in this table are averaged from the 5000 results.
 \item[2] For these roadmaps, failure occurs when no collision-free straight-line connection was found to an existing point on the roadmap from the start or goal pose of a test case.
 \item[3] Best average is the shortest average path length between all tested sampling-based planners in that environment.  Shown here to provide context for the roadmap performance.
 \end{tablenotes}
 \end{threeparttable}
\end{table}

TrajOpt requires the number of waypoints in the solution trajectory to be the same as in the seed. Therefore, if we pass in seeds directly from sampling-based planners without any pre-processing, the number of waypoints in different queries will fluctuate drastically. As mentioned in Section~\ref{limitation}, TrajOpt's runtime increases approximately linearly as the number of waypoints increases, which means the variation of waypoint numbers will influence runtime. Additionally, seeds taken directly from the sampling-based planners with a fewer number of waypoints might result in higher collision rates after shortening and smoothing through TrajOpt compared to those with more waypoints. This is because such seed trajectories usually have longer edges in-between waypoints and are more likely to be very close to obstacles. Hence, before passing the seeds into TrajOpt, we interpolate them by setting a upper bound of 0.16 rad for the distance between adjacent waypoints. This pre-processing dramatically reduced the collision rate of TrajOpt solutions and narrowed down the variance of TrajOpt's runtime among different cases. Although the average TrajOpt runtime is increased due to the increased number of waypoints after interpolation, it is still under 1s in most environments. 

\begin{table}
\caption{TrajOpt Seeded with Sampling-based Planner Solution compared to Chekov Roadmap Solution}
\label{table_trajopt-roadmap}
\small\sf\centering
\begin{threeparttable}
\begin{tabular}{L{1.15cm}|L{1.1cm} L{1cm} L{1.1cm}|L{1cm} L{0.95cm} L{0.9cm}}
\toprule
\multirow{5}{1cm}{\centering Environ-ments} & \multirow{5}{1cm}{\centering Seed Planners} & \multirow{5}{1cm}{\centering Average TrajOpt Runtime (s)} & \multirow{5}{1cm}{\centering Average Seed Length (rad)} & 
\multicolumn{3}{c}{Seed + TrajOpt Planner} \\ \cline{5-7}
& & & & Average Runtime (s)\tnote{1} & Average Path Length (rad) & Colli-sion Rate\tnote{2} \\
\hline
\multirow{5}{1cm}{\centering Tabletop with a Pole} 
& RRT & 0.63 & 0.77 & 18.51 & 0.70 & 1.29\% \\ 
& LazyPRM & 0.98 & 1.76 & 8.30 & 1.28 & 0.12\% \\
& RRT* & 0.29 & 0.63 & 300.48 & 0.54 & 0.02\% \\ 
& PRM* & 0.36 & 0.79 & 301.07 & 0.64 & 0.10\% \\ 
& Chekov & 0.45 & 1.24 & 0.59 & 0.82 & 0.06\% \\ 
\hline
\multirow{5}{1cm}{\centering Tabletop with a Contain-er} 
& RRT & 1.02 & 0.92 & 45.92 & 0.85 & 2.18\% \\ 
& LazyPRM & 1.55 & 1.92 & 16.59 & 1.44 & 0.96\% \\ 
& RRT* & 0.44 & 0.80 & 300.73 & 0.70 & 0.90\% \\ 
& PRM* & 0.49 & 1.04 & 301.22 & 0.84 & 1.12\% \\ 
& Chekov & 0.52 & 1.32 & 0.70 & 1.02 & 0.90\% \\ 
\hline
\multirow{5}{1cm}{\centering Shelf with Boxes} 
& RRT & 0.92 & 1.06 & 64.87 & 0.98 & 4.20\% \\ 
& LazyPRM & 1.36 & 2.08 & 65.21 & 1.60 & 1.57\% \\ 
& RRT* & 0.46 & 0.93 & 300.83 & 0.81 & 1.17\% \\ 
& PRM* & 0.67 & 1.16 & 301.46 & 0.95 & 1.98\% \\ 
& Chekov & 0.61 & 1.30 & 1.00 & 1.02 & 1.98\% \\
\hline
\multirow{5}{1cm}{\centering Kitchen} 
& RRT & 0.99 & 0.78 & 46.95 & 0.72 & 0.52\% \\ 
& LazyPRM & 1.28 & 1.67 & 19.31 & 1.11 & 0.35\% \\ 
& RRT* & 0.45 & 0.71 & 300.72 & 0.62 & 0.37\% \\ 
& PRM* & 0.54 & 0.87 & 301.43 & 0.70 & 0.46\% \\ 
& Chekov & 0.70 & 1.29 & 1.08 & 0.86 & 0.73\% \\ 
\bottomrule
\end{tabular}
\begin{tablenotes}
\footnotesize
 \item[1] Sum of seed planner runtime and TrajOpt runtime averaged from 5000 test cases.
 \item[2] Continuous-time collision rate.
 \end{tablenotes}
 \end{threeparttable}
\end{table}

\begin{figure*}
  \centering
    \framebox{\includegraphics[width=0.81\linewidth]{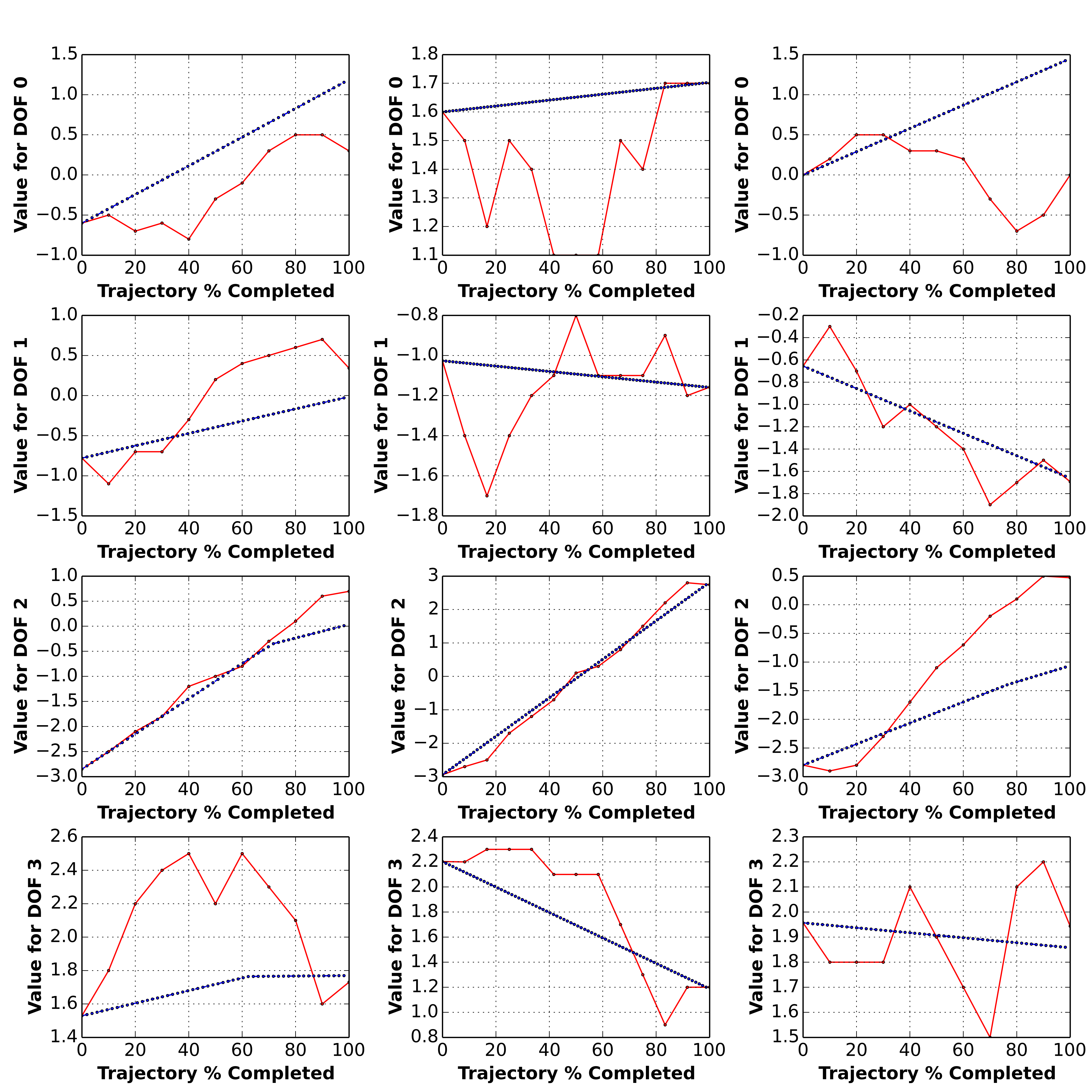}}
    \caption{Roadmap seed trajectories shown with corresponding trajectories optimized by TrajOpt to illustrate improvement on the seed~\citep{dai2018improving}. The solid lines are the roadmap seeds and the dashed lines are the outputted trajectories by TrajOpt when provided those seeds.}
    \label{figure-trajopt}
\end{figure*}

Table~\ref{table_trajopt-roadmap} provides a comparison between the combined ``sampling-based + TrajOpt'' planners with existing sampling-based planners and with our Chekov roadmap planner. Comparing the TrajOpt runtime column in Table~\ref{table_trajopt-roadmap} and the straight-line seed TrajOpt runtime in Table~\ref{table_planners}, we see that TrajOpt's runtime usually decreases when provided with a collision-free seed. Specifically, in the cases where TrajOpt with a straight-line seed failed to push the trajectory out of collision, we found a 50\% - 70\% runtime drop after provided with sampling-based planners' solutions as initializations. Although a small percentage of cases end up in collision after TrajOpt's smoothing and optimization, a significant improvement in average C-space path length is observed if we compare the ``average path length'' column in Table~\ref{table_planners} and in Table~\ref{table_trajopt-roadmap}. However, the ``average runtime'' for combinations with existing sampling-based motion planners indicates that it is infeasible to use them as seed planners in real-time motion planning tasks. In contrast, the ``Chekov roadmap + TrajOpt'' combination shows an average run-time for about 1~s in all four tested environments.

When the roadmap planner produces a solution, TrajOpt in turn produces a collision-free trajectory more than 98\% of the time. Additionally, these optimized trajectories are on average more than 10\% shorter than their corresponding seed trajectories. Figure \ref{figure-trajopt} shows the four proximal joints for three different trajectories to help visualize the improvements TrajOpt is making on the seed trajectories. The solid lines are the roadmap seeds and the dashed lines are the outputted trajectories by TrajOpt when provided those seeds. From Figure \ref{figure-trajopt} we can see that TrajOpt fulfilled the task of smoothing and shortening the sub-optimal trajectories produced by the Chekov roadmap. In Table~\ref{table_trajopt-roadmap}, the difference in average runtime of the different seed planner coupled with TrajOpt is most notable for highlighting the performance improvements provided by our roadmap planner, but runtime as a metric does not reveal the whole picture for many of these planners.  As noted earlier, the optimal planners like RRT* will always use the full allotted time but may have a good non-optimal solution far sooner than that.  Also, in our test cases, LazyPRM constructs its roadmap online for one time use and then searches for a path in that roadmap.  In general, a PRM does not lend itself to single-query problems.  Our roadmap planner precomputes the roadmap and all-pair-shortest-path solutions, but is also essentially a PRM.  It would be interesting to compare the performance of our roadmap planner to faster RRT variants, but it is clear to us that the speed provided by querying precomputed solutions from a PRM of some form outweighs any optimization to be had in online search.

Overall, our roadmap planner performs as well as if not better than the off-the-shelf sampling-based planners we tested. Average runtime is where we saw the greatest improvement when using our roadmap planner to provide seed solutions rather than using other traditional sampling-based planners, which is promising given that one of our main goals is to establish a fast-reactive motion planning and execution system for high-dimensional robots. Although we are currently not using dynamic obstacles in our experiments, our average online planning time leaves us optimistic that our planner will be able to handle disturbances in planning tasks with fast reaction.

\section{Chance-constrained motion planning approach -- probabilistic Chekov}  \label{p-chekov}

\begin{figure}
 \begin{center}
 \includegraphics[width=1.02\linewidth]{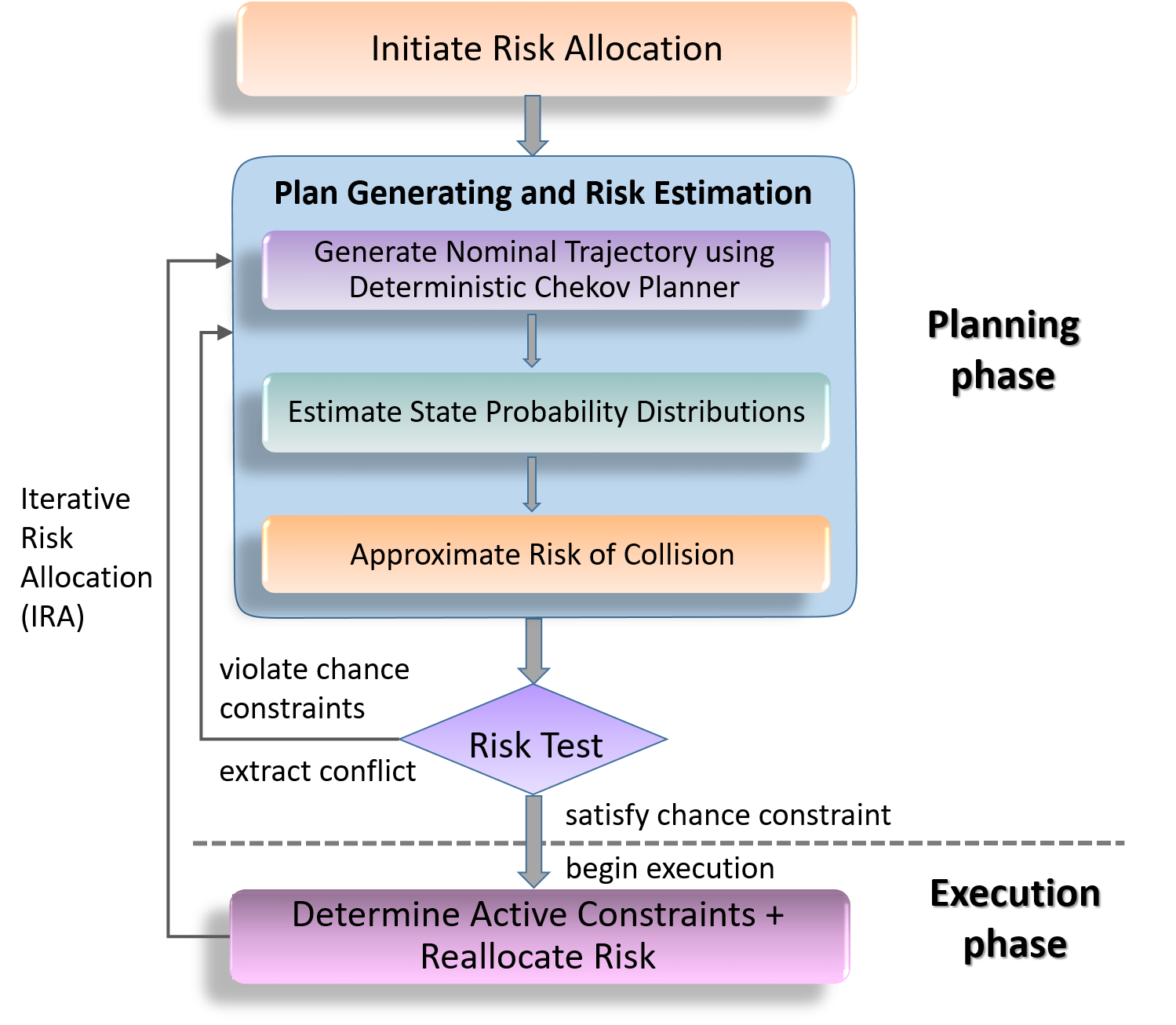}
 \caption{System diagram for p-Chekov~\citep{dai2019chance}}
 \label{fig:diagram}
 \end{center}
\end{figure}

This section introduces the probabilistic Chekov (p-Chekov) risk-aware motion planning and execution system that accounts for the potential uncertainties during execution while making plans and returns solutions that can satisfy user-specified chance constraints over plan failure. Figure \ref{fig:diagram} shows the system diagram of p-Chekov, which can be divided into a planning phase and an execution phase. The goal in the planning phase is to find a feasible solution trajectory along which the estimated risk of collision is smaller than or equal to the given joint chance constraint $\Delta_c$. Since this initial solution is not guaranteed to be optimal and can sometimes be overly conservative, p-Chekov will keep improving it in an anytime manner during the execution phase in order to achieve better utility.

In p-Chekov, time is discretized into fixed-interval time steps, and the collision risk at each waypoint is considered separately through risk allocation. When the planning phase starts, p-Chekov first uniformly distributes the joint chance constraint into the allowed collision risk bounds for each waypoint along the trajectory. Provided with a risk allocation, p-Chekov then uses the deterministic Chekov approach described in Section~\ref{rm-seeds} to generate a nominal trajectory that is feasible and collision-free under deterministic dynamics. Given the estimated model of controller and sensor noises during execution, p-Chekov then estimates the \emph{a priori} probability distribution of robot states along this nominal trajectory, which will be introduced in Section~\ref{LQG-MP}. With this state distribution information, p-Chekov provides two different approaches for estimating the probability of collision at each waypoint along this trajectory: a quadrature-based sampling approach and an offline-trained function approximation approach, which will both be explained in Section~\ref{quadrature_sampling}. After that, we can compare the allocated risk bound and the estimated probability of collision at each waypoint along the nominal solution trajectory, shown as the ``risk test'' step in Figure~\ref{fig:diagram}. If the nominal trajectory fails to pass the risk test, the robot configurations at the waypoints where the estimated risk of collision exceeds the allocated risk bound will be viewed as conflicts. Before p-Chekov goes back to the ``plan generating and risk estimation'' stage in Figure~\ref{fig:diagram},
constraints associated with the conflict configurations and conflict waypoints will be added so that deterministic Chekov can be guided to find safer nominal trajectories.
Additionally, p-Chekov will also reallocate the waypoint risk bounds, as will be explained in detail through Algorithm~\ref{algo:reallocation} in Section~\ref{risk_reallocation}. This risk reallocation takes risk bounds from the waypoints where they are underutilized to the ones where they are violated, so that feasible trajectories can be found in fewer iterations. With the above new constraints, p-Chekov will then replan and improve the nominal trajectory from the previous iteration. This cycle will keep going until the solution trajectory satisfies the chance constraints at all waypoints or the iteration number hits its upper bound.

When the nominal trajectory passes the risk test, p-Chekov will transition to the execution phase, where it optimizes the solution it found in the planning phase while the robot is executing the trajectory. This plan refinement is based on the iterative risk allocation (IRA) algorithm, which would be illustrated in Section~\ref{execution_phase_IRA}. Through gradually reallocating the risk from inactive constraints to active constraints, IRA provides less conservative risk allocations and allows for higher quality motion plans. After that, p-Chekov will go back to the ``plan generating and risk estimation'' stage with zero penalty hit-in distance and find a new feasible solution which satisfies the new risk allocation. When it finds a valid plan, the robot will keep executing based on the updated plan. This risk reallocation and plan refinement process is conducted iteratively, which will help the planner to converge to a locally optimal solution if given enough number of iterations.

\subsection{Approach for estimating robot state probability distributions}  \label{LQG-MP}

The ``estimate state probability distributions'' component in Figure~\ref{fig:diagram} plays a significant role of estimating the robot state probability distributions along the nominal trajectory during execution based on the given noise level. In this paper, we present a linear-quadratic Gaussian motion planning (LQG-MP) approach~\citep{van2011lqg} which can act as this state probability estimator in p-Chekov. LQG-MP connects control theory and probabilistic motion planning by taking into account the controllers and sensors that will be used during execution and characterizing the \emph{a priori} probability distributions of robot states when making motion plans. In our implementation of LQG-MP, it is assumed that a discrete-time Kalman filter~\citep{gelb1974applied} and a finite-horizon discrete-time LQR controller~\citep{bertsekas1995dynamic} will be used during execution, and the deviation from the desired trajectory during execution is small enough so that the control effort needed to bring the robot back on track will not exceed the controller limit. In order to achieve optimal control policies according to the \emph{separation theorem}~\citep{luenberger1979introduction}, it is also assumed that both process noises and observation noises have Gaussian distributions. In real-world scenarios, noises often accumulate from inconsistent, random sources, and based on the Central Limit Theorem~\citep{hoeffding1948central}, thus the Gaussian noise assumption are appropriate in many applications. However, developing a state probability estimator that can incorporate non-Gaussian noises to replace LQG-MP is a very interesting future work direction.

We express the system model in terms of the deviations from the desired trajectory $\Pi = (\mathbf{x}^*_0, \mathbf{u}^*_0, \ldots, \mathbf{x}^*_T, \mathbf{u}^*_T)$:

\begin{equation}
\label{deviation}
 \begin{aligned}
  & \bar{\mathbf{x}}_t = \mathbf{x}_t - \mathbf{x}^*_t, \\
  & \bar{\mathbf{u}}_t = \mathbf{u}_t - \mathbf{u}^*_t, \\
  & \bar{\mathbf{z}}_t = \mathbf{z}_t - h(\mathbf{x}^*_t, 0). \\
 \end{aligned}
\end{equation}

\noindent Since robot motions will be controlled to closely follow the planned trajectory during execution, it is reasonable to linearize the system dynamics model and observation model as:

\begin{equation}
\begin{aligned}
& \bar{\mathbf{x}}_t = A_t\bar{\mathbf{x}}_{t-1} + B_t\bar{\mathbf{u}}_{t-1} + V_t\mathbf{m}_t, & & \mathbf{m}_t \sim \mathcal{N}(0, M_t),  \\
& \bar{\mathbf{z}}_t = H_t\bar{\mathbf{x}}_t + W_t\mathbf{n}_t, & & \mathbf{n}_t \sim \mathcal{N}(0, N_t).
\end{aligned}
\end{equation}

\noindent where $A_t$, $B_t$, $V_t$, $H_t$ and $W_t$ are the Jacobian matrices of $f$ and $h$ along the desired trajectory $\Pi$. 

In LQG, since the true state $\bar{\mathbf{x}}_t$ is unknown, the state estimation $\tilde{\mathbf{x}}_t$ from the Kalman filter is used to determine the control input at each time step during the trajectory execution.  This is reasonable because the separation theorem tells us that observer design and controller design can be separated into two independent processes with the guarantee of LQG optimality. During the execution of the whole desired trajectory, optimal state estimations based on the Kalman filter and optimal control policy computations based on LQR take turns and cycles until the execution is complete, so as to optimize the execution and track the desired trajectory.

If we denote the Kalman gain as $L_t$ and the controller gain as $K_t$, then the evolution of the true state $\bar{\mathbf{x}}_t$ and the estimated state $\tilde{\mathbf{x}}_t$ at each time step $t$ can be predicted as follows~\citep{van2011lqg}:

\begin{equation}
\begin{aligned}
  \begin{bmatrix}
   \bar{\mathbf{x}}_t \\
   \tilde{\mathbf{x}}_t \\
  \end{bmatrix}
= & 
  \begin{bmatrix}
   A_t       & B_tK_t \\
   L_tH_tA_t & A_t + B_tK_t - L_tH_tA_t \\
  \end{bmatrix}
\begin{bmatrix}
 \bar{\mathbf{x}}_{t-1} \\
   \tilde{\mathbf{x}}_{t-1} \\
\end{bmatrix}
 \\ & +
\begin{bmatrix}
   V_t       & 0 \\
   L_tH_tV_t & L_tW_t \\
  \end{bmatrix}
  \begin{bmatrix}
   \mathbf{m}_t \\
   \mathbf{n}_t \\
  \end{bmatrix},
\end{aligned}
\end{equation}

\noindent where

\begin{equation}
  \begin{bmatrix}
   \mathbf{m}_t \\
   \mathbf{n}_t \\
  \end{bmatrix}
\sim \mathcal{N}(\mathbf{0}, 
\begin{bmatrix}
 M_t & 0 \\
 0   & N_t \\
\end{bmatrix}).
\end{equation}

If we define 

\begin{equation}
\begin{aligned}
 & \mathbb{X}_t \triangleq 
 \begin{bmatrix}
   \bar{\mathbf{x}}_t \\
   \tilde{\mathbf{x}}_t \\
  \end{bmatrix}, \\
 & E_t = 
 \begin{bmatrix}
   A_t       & B_tK_t \\
   L_tH_tA_t & A_t + B_tK_t - L_tH_tA_t \\
  \end{bmatrix}, \\
 & F_t = 
 \begin{bmatrix}
   V_t       & 0 \\
   L_tH_tV_t & L_tW_t \\
  \end{bmatrix}, \\
 & G_t = 
 \begin{bmatrix}
 M_t & 0 \\
 0   & N_t \\
\end{bmatrix}, \\
 \end{aligned}
\end{equation}

\noindent and initialize the variances for estimate states with 0 and the variances for true states with $\Sigma_0$, then the variance matrix $C_t$ for $\mathbb{X}_t$ can be expressed as:

\begin{equation}
\begin{aligned}
 & C_t = E_tC_{t-1}E^T_t + F_tG_tF^T_t, & ~~ & C_0 = \begin{bmatrix} \Sigma_0 & 0 \\ 0 & 0 \\ \end{bmatrix}.
 \end{aligned}
\end{equation}

Therefore, the matrix of true states and estimated states $\mathbb{X}_t$ has the distribution:

\begin{equation}
 \mathbb{X}_t \sim \mathcal{N}(\mathbf{0}, C_t).
\end{equation}

\noindent Substitute into Equation \ref{deviation}, we can get the \emph{a priori} distributions of the true states and control inputs during the execution of the desired trajectory:

\begin{equation}
 \begin{bmatrix}
  \mathbf{x}_t \\ \mathbf{u}_t
 \end{bmatrix}
\sim \mathcal{N}(
\begin{bmatrix}
 \mathbf{x}^*_t \\ \mathbf{u}^*_t
\end{bmatrix}
,  \Lambda_tC_t\Lambda^T_t),
\end{equation}

\noindent where

\begin{equation}
 \Lambda_t = \begin{bmatrix} I & 0 \\ 0 & K_{t+1} \end{bmatrix}.
\end{equation}

With these \emph{a priori} distributions of robot states, we can then evaluate the probability of collision along the desired trajectory to find feasible solutions that can satisfy the given chance constraint.

\subsection{Collision probability estimation approach}  \label{CP}

Continuous-time collision risk is difficult to represent, so most collision probability estimation approaches divide the entire trajectory into discrete waypoints, estimate the collision probability at each waypoint, and then use additive or multiplicative approximations to represent the collision risk along the entire trajectory. In the additive approach, Boole's inequality tells us that:

\begin{equation}
 P\Bigg(\bigvee^T_{t=1} \overline{S_t}\Bigg) \leq \sum_{t=1}^T P\Big(\overline{S_t}\Big),
\end{equation}

\noindent  where $S_t$ is the no-collision constraint at waypoint $t$. Therefore,

\begin{equation}
\sum_{t=1}^T P\Big(\overline{S_t}\Big) \leq \Delta
\end{equation}

\noindent is the sufficient condition for

\begin{equation}
 P\Bigg(\bigvee^T_{t=1} \overline{S_t}\Bigg) \leq \Delta,
\end{equation}

\noindent where $\Delta$ is the joint chance constraint for all the waypoints along a given trajectory. Similarly, the multiplicative approach assumes independence between the collision probabilities at different waypoints, and uses

\begin{equation}
 1 - \prod_{t=1}^T\Bigg(1 - P\Big(\overline{S_t}\Big)\Bigg) \leq \Delta
\end{equation}

\noindent to approximate

\begin{equation}
 P\Bigg(\bigvee^T_{t=1} \overline{S_t}\Bigg) \leq \Delta.
\end{equation}

Note that neither of these two approaches can account for edge collisions between waypoints. In addition, both approximations have made strong assumptions about the complex high-dimensional correlation between collisions at different waypoints. Specifically, the additive approach assumes that the collisions at different waypoints are mutually exclusive, whereas the multiplicative approach assumes they are independent from each other. If we don't take edge collisions into consideration, then the additive approach is always guaranteed to be conservative, but the multiplicative approach only has the conservativeness guarantee when $T$ approaches infinity. In p-Chekov, we adopt the additive discretization approximation and estimate the collision risk at individual waypoints, then alleviate the conservative shortcoming of additive approaches through risk reallocation. We present two different approaches for estimating waypoint collision risk in Section~\ref{quadrature_sampling} and Section~\ref{ml_approach} respectively: a quadrature-based approach and a learning-based approach. Note that both approaches still inherit the conservativeness from additive risk approximation and tend to fail at finding feasible solutions in environments with narrow spaces, e.g. the shelf with boxes environment introduced in this paper. How to take the mutual correlation between waypoint collisions into consideration and relax the conservativeness issue in p-Chekov is a very interesting direction for future research.

\subsubsection{Quadrature-based collision probability estimation}  \label{quadrature_sampling}
 
Given the state probability distribution around a nominal configuration, the collision probability can be approximated by sampling from this distribution and checking the percentage of configurations that are in collision. However, as with all Monte Carlo methods, this approach would suffer from inaccuracy when the sample size is small and high computational cost when the sample size is large. We tested the speed of the Flexible Collision Library (FCL) collision checker~\citep{pan2012fcl} p-Chekov uses in the ``kitchen'' environment with 55 obstacles, and results show that 100 collision checks take about 0.2 s. Although FCL is one of the fastest collision checking tools, it is still infeasible for p-Chekov to be a real-time motion planner if we use simple Monte Carlo in the 7-dimensional C-space. Therefore, an intelligent sampling method that can closely approximate the collision probability with only a small number of samples is very important~\citep{dai2019chance}.

This Monte Carlo collision probability estimation approach is essentially estimating the expectation of a collision function:

\begin{equation*}
\begin{aligned} 
  c(\mathbf{x}_t) = \begin{cases} \mbox{0,} & \mbox{if } \mathbf{x}_t \mbox{ is collision free} \\ \mbox{1,} & \mbox{if } \mathbf{x}_t \mbox{ is in collision} \end{cases}
\end{aligned} 
\end{equation*}

\noindent along the distribution $\mathbf{x}_t \sim \mathcal{N}(\hat{\mathbf{x}}_t, \mathbf{\Sigma}_{\mathbf{x}_t})$ estimated in Section~\ref{LQG-MP}, where $\mathbf{x}_t \in \mathbb{R}^{n_x}$ is the nominal configuration at time step~$t$. Since expectations can be written as integrals, non-random numerical integration methods (also called quadratures~\citep{hildebrand1987introduction}) can be applied to solve this problem. Assume $\mathbf{x}_t$ is $d$-dimensional and let $x_t^i$ denote its $i$th component whose distributions are independent from each other. This assumption is reasonable because correlated noise components can be transformed through robot state space coordinate transformation so that the covariance matrices will become diagonal. Since $\mathbf{x}_t$ is Gaussian distributed, we can write $x_t^i \sim \mathcal{N}(\mu_i, \sigma_i^2)$. Then, based on the conditional distribution rule of multivariate normal distribution~\citep{eaton1983multivariate}, the probability density function of $\mathbf{x}_t$ can be expressed as:

\begin{equation}
\begin{aligned}
 & p(\mathbf{x}_t) = p(\mathbf{x}_t^{1:d}) = p(x_t^1)p(\mathbf{x}_t^{2:d}|x_t^1) = p(x_t^1)p(\mathbf{x}_t^{2:d}), \\
 & x_t^1 \sim \mathcal{N}(\mu_1, \sigma_1^2), \\
 & \mathbf{x}_t^{2:d} \sim \mathcal{N}(\mathbf{\mu}_{2:d}, \mathbf{\Sigma}_{2:d}), \\
\end{aligned}
\end{equation}

\noindent where $\mathbf{\mu}_{2:d}$ and $\mathbf{\Sigma}_{2:d}$ denote the mean and variance of $\mathbf{x}_t^{2:d}$ respectively. Then we can write the expectation of the collision function as:

\begin{equation}
 \mathbb{E}(c(\mathbf{x}_t)) = \int_{-\infty}^\infty p(x_t^1) \int_{\mathbb{R}^{n_x - 1}} p(\mathbf{x}_t^{2:d}) c(\mathbf{x}_t) d\mathbf{x}_t^{2:d} dx_t^1.
\end{equation}

\noindent Let $g(x_t^1) = \int_{\mathbb{R}^{n_x - 1}} p(\mathbf{x}_t^{2:d}) c(\mathbf{x}_t) d\mathbf{x}_t^{2:d}$ and apply the probability density function of Gaussian distributions, we have:

\begin{equation}
\begin{aligned}
\label{expectation}
 \mathbb{E}(c(\mathbf{x}_t)) & = \int_{-\infty}^\infty p(x_t^1)g(x_t^1)dx_t^1 \\
 & = \int_{-\infty}^\infty \frac{1}{\sigma_1 \sqrt {2\pi }} \exp\Big(- \frac{(x_t^1 - \mu_1)^2}{2 \sigma_1^2}\Big)g(x_t^1)dx_t^1.
\end{aligned}
\end{equation}

Gauss-Hermite quadrature approximates the value of an integral by calculating the weighted sum of the integrand function at a finite number of reference points, i.e.

\begin{equation}
 \int_{-\infty}^\infty e^{-y^2} h(y) dy \approx \sum_{j = 1}^n w_j h(y_j),
\end{equation}

\noindent where $n$ is the number of sampled points, $x_j$ are the roots of the Hermite polynomial $H_n(x)$ and the associated weights $w_j$ are given by~\cite{abramowitz1964handbook}:

\begin{equation}
 w_j = \frac{2^{n-1}n!\sqrt{\pi}}{n^2[H_{n-1}(y_j)]^2}.
\end{equation}

\noindent A quadrature rule with $n$ sampled points is called a $n$-point rule.

$\mathbb{E}(c(\mathbf{x}_t))$ in its form in Equation \ref{expectation} still doesn't correspond to the Hermite polynomial, therefore we conduct the following variable change:

\begin{equation}
\label{variable}
 y_1 = \frac{x_t^1 - \mu_1}{\sqrt{2}\sigma_1} \Leftrightarrow x_t^1 = \sqrt{2}\sigma_1y_1 + \mu_1.
\end{equation}

\noindent Applying Equation \ref{variable} to Equation \ref{expectation} yields:

\begin{equation}
 \mathbb{E}(c(\mathbf{x}_t)) = \int_{-\infty}^\infty \frac{1}{\sqrt{\pi}} e^{-(y_1)^2}g(\sqrt{2}\sigma_1 y_1 + \mu_1) dy_1.
\end{equation}

\noindent If we iteratively conduct this Gauss-Hermite quadrature approximation procedure from $x_t^1$ through $x_t^d$, we will be able to approximate the value of $\mathbb{E}(c(\mathbf{x}_t))$ through:

\begin{equation}
\label{equation:estimation}
\begin{aligned}
 \mathbb{E}(c(\mathbf{x}_t))\approx & ~\pi^{-\frac{d}{2}} \sum_{j_1=1}^{n_1}  \sum_{j_2=1}^{n_2} \ldots \sum_{j_d=1}^{n_d} \Bigg(\prod_{i=1}^d w_{i,j_i}\Bigg) g(\sqrt{2}\sigma_1 y_{1, j_1} \\ 
 &  + \mu_1, \sqrt{2}\sigma_2 y_{2, j_2} + \mu_2, \ldots , \sqrt{2}\sigma_d y_{d, j_d} + \mu_d). \\
\end{aligned}
\end{equation}

In one-dimensional space, a $n$-point rule yields $2n$ parameters and it is possible to integrate polynomials of degree up to $2n-1$ without error. For $a < x < b$ and $h(x)$ with $2n$ continuous derivatives, the error in a Gauss rule is:

\begin{equation}
 \frac{(b-a)^{2n+1}(n!)^4}{(2n+1)[(2n)!]^3}h^{(2n)}(x).
\end{equation}

\noindent Note that although quadrature methods are well tuned to one-dimensional problems, extending them to multi-dimensional problems through iterated one-dimensional integrals still can't escape the ``curse of dimensionality''~\citep{bellman1957dynamic}. The result of a $d$-dimensional quadrature rule can not be better than the worst of the rules we use in each dimension. If we use the same $n$-point one-dimensional quadrature rule for each of the $d$-dimensions, then we need $N = n^d$ function evaluations. If the one-dimensional rule has error $O(n^{-r})$, then the combined rule has error

\begin{equation}
 |\hat{I} - I| = O(n^{-r}) = O(N^{-r/d}).
\end{equation}

\noindent Even a modestly large $d$ can give a very inaccurate result~\citep{owen2014monte}. Additionally, the collision function $c(\mathbf{x}_t)$ we are trying to evaluate is not smooth, which adds to the inaccuracy of the approximations through this quadrature-based sampling method. Consequently, this quadrature-based collision probability estimation approach is a relatively rough one. 

To achieve fast online motion planning for a 7-DOF manipulator, in p-Chekov, the number of quadrature points at each dimension shouldn't be too large. Table \ref{quadrature:table} shows the abscissas and weights of the two- and three-point Gauss-Hermite quadrature rules. We hypothesize that the two-point rule will generate more conservative risk estimations, because the three-point rule places higher weights on the mean values, which in p-Chekov are the nominal configurations that are guaranteed to be collision-free. Empirical results proved that using the two-point rule is safer and also much faster, thus it is used in our implementation.

\begin{table}[h]
\caption{Gauss-Hermite Quadrature Abscissas and Weights}
\label{quadrature:table}
\small\sf\centering
\begin{tabular}{L{2cm}  L{2cm}  L{2cm}} \toprule
 $n$   & $x_i$ & $w_i$ \\ \hline
\rule{0pt}{15pt} 2     & $\pm \frac{1}{2} \sqrt{2}$   & $\frac{1}{2} \sqrt{\pi}$    \\ \hline
\rule{0pt}{15pt}  \multirow{2}{2cm}{\centering 3} & 0  & $\frac{2}{3} \sqrt{\pi}$ \\ 
\rule{0pt}{15pt}   & $\pm \frac{1}{2} \sqrt{6}$ & $\frac{1}{6} \sqrt{\pi}$ \\ \bottomrule 
\end{tabular}
\end{table}

Note that the state probability distribution computed from LQG-MP doesn't fully reflect the true probability distribution of joint states due to joint limits. In p-Chekov, this issue is addressed by using the upper or lower bound of joint values instead of the actual point sampled from the estimated distribution when the joint limit is exceeded. We choose to handle it this way because in practice, deviations from the desired configuration are usually caused by internal or external disturbance to robot joints. When the disturbance tends to push one of the joints towards a point which exceeds its upper bound, this joint will end in its upper bound position instead of exceeding the joint limit. 

This sampling-based approach of estimating the collision probability based on Gauss-Hermite quadrature theory can be summarized in Algorithm~\ref{algorithm:cp}.  Given a nominal trajectory and the corresponding state probability distributions, it first calculates the abscissas and weights for each DOF of the target manipulator through applying the $n$-point Gauss-Hermite quadrature rule to each one-dimensional Gaussian distribution. After checking joint limits, it stores the sampled abscissas and weights in a $NodeList$ (line 7-11), and then evaluates the collision risks of the robot configured at all the combinations of these DOF values according to Equation~\ref{equation:estimation}. Algorithm~\ref{algorithm:cp} iteratively conducts this quadrature-sampling and collision probability evaluating procedure for each waypoint along the nominal trajectory, and then returns the collision probabilities as a list $\mathbf{r}$. These probabilities are then compared with the allocated risk bound at each waypoint in order to determine whether the joint chance constraint for the whole trajectory is satisfied. A detailed description of the risk allocation approach will be provided in Section~\ref{IRA}.

\begin{algorithm}
 \caption{GHCollisionProbabilityEstimation}
 \label{algorithm:cp}
 \small
 \DontPrintSemicolon
 \KwIn{\\ $\Pi$: \text{desired trajectory}
 \\ $\mathcal{D}$: \text{robot state distribution along desired trajectory}
 \\ $\mathcal{R}, \mathcal{E}$: \text{robot and environment collision models respectively}
 \\ $dof$: \text{robot degrees of freedom}
 \\ $n$: \text{number of samples used in quadrature rule}
 \\ $l_u, l_l$: \text{upper and lower limits of active joints respectively}}
 \KwOut{
 \\ $\mathbf{r}$: \text{collision risk at each waypoint along desired trajectory} }
 
 Initialize $\mathbf{r}$ to a list of zeros \\
 \For{\forcond $i=1, 2, \ldots$, len($\Pi)$}{
 Initialize $NodeList$ to an empty set \\
 \For{\forcond $d=1, 2, \ldots, dof$}{
 $(\mu, \sigma) \leftarrow \mathcal{D}[i, d]$   \tcc*[r]{Draw from $\mathcal{D}$ at the $i$th waypoint $d$th joint} 
 $(nodes, weights) \leftarrow$ QuadratureSampling($\mu, \sigma, n$) \\
 \For{\forcond $node$ in $nodes$}{
 \llIf{$node > l_u[d]$}{$node \leftarrow l_u[d]$} \\
 \llIf{$node < l_l[d]$}{$node \leftarrow l_l[d]$} }
 Append $(nodes, weights)$ to $NodeList$ }
 Estimate $\mathbf{r}(i)$ by taking nodes from $NodeList$, checking collision with $\mathcal{E}, \mathcal{R}$, and averaging the collision number
 }
\end{algorithm}

\subsubsection{Learning-based collision probability estimation}  \label{ml_approach}

The quadrature-based sampling approach introduced in Section~\ref{quadrature_sampling} mitigates the inaccuracy of random sampling and avoids the difficulty of mapping between C-space and workspace. Although it can significantly reduce the number of samples required for collision risk estimation at each time step in the trajectory, its computation time in high-dimensional planning space still obstructs its application in real-time motion planning tasks. Even though only two quadrature nodes per dimension are used to estimate the collision risk for each waypoint, the total number of collision tests conducted online is still very big when the manipulator have 7 DOFs ($2^7 \times n_{waypoints}$ collision tests for each nominal trajectory). Additionally, two-node quadratures have very limited ability of approximating non-smooth functions, whereas the collision functions here are highly non-smooth. Therefore, quadrature-based p-Chekov inevitably suffers from errors when approximating the collision risk, and the efficiency and accuracy of risk estimation becomes its bottleneck that restricts its application in uncertainty-sensitive real-time manipulation planning tasks. Therefore, this section introduces machine learning approaches into the collision risk estimation component of p-Chekov in order to improve its efficiency and accuracy. 

We hypothesize that if we take enough samples containing nominal configurations with their probability distributions and risks of collision from the environment that the robot will be interacting with in order to train a regression model offline, then this model can act as the ``Approximate Risk of Collision'' component in Figure~\ref{fig:diagram} in the online planning phase which makes accurate predictions given a nominal trajectory and the state distributions outputted by the ``LQG-MP'' component. In order to test this hypothesis, 60000 data points are collected in each of the tabletop environments, each of which contains a nominal joint configuration that is randomly sampled from the uniform distribution defined by the manipulator's joint limit, a randomly sampled standard deviation whose range is decided according to the real experiment data from quadrature-based p-Chekov tests, and a collision risk scalar that is viewed as the ``ground-truth'' risk associated with this configuration distribution. This collision risk is estimated using a simple Monte Carlo method: randomly sample 100000 nodes from the Gaussian distribution defined by the nominal joint configuration and the standard deviation, and compute the average collision rate. The nominal configuration together with its standard deviation forms the input vector to the regression algorithm, and the collision risk is its label. 

This paper compares the performance of three different classes of regressors in the Scikit Learn~\citep{scikit-learn} package (kernel ridge regressor, random forest regressor, and Gaussian process regressor) as well as neural networks through the Keras~\citep{chollet2015keras} interface with TensorFlow~\citep{tensorflow2015-whitepaper} back engine. Kernel ridge regression~\citep{murphy2012machine} learns a linear function in the space induced by the respective kernel and the data, and minimizes the objective:

\begin{equation}
 J = ||y - w^T X||^2 + \alpha ||w||^2,
\end{equation}

\noindent where $X$ is the input vector, $y$ is the true label, $w$ is the weight vector given by the regressor and $\alpha$ is the parameter that determines the regularization strength. In the kernel ridge regression tests in this paper, the performance of three different classes of kernels are compared: radial basis function (RBF) kernel, polynomial kernel and Matern kernel. In RBF kernels, each element in the kernel matrix between datasets $X$ and $Y$ is computed by:

\begin{equation}
 K(x, y) = exp(-\gamma ||x - y||^2)
\end{equation}

\noindent for each pair of rows $x$ in $X$ and $y$ in $Y$. Therefore, the parameter $\gamma$ represents how far the influence of a single training example reaches, with low values meaning ``far'' and high values meaning ``close''. In polynomial kernels, $degree$ is a parameter that represents the order of polynomials used in the kernel. A $degree-d$ polynomial kernel is defined as:

\begin{equation}
 K(x, y) = (x^T y + c)^d,
\end{equation}

\noindent where $c \geq 0$ is a free parameter trading off the influence of higher-order versus lower-order terms in the kernel. Matern kernel is defined by:

\begin{equation}
 K(x, y) = \frac{1}{2^{\nu - 1} \Gamma (\nu)}(\frac{2\sqrt{\nu} ||x - y||}{\theta})^\nu H_\nu (\frac{2\sqrt{\nu} ||x - y||}{\theta}),
\end{equation}

\noindent where the length scale parameter $\theta$ is similar to the $\gamma$ in RBF kernels, $\Gamma$ is the Gamma function, the $\nu$ parameter controls the smoothness of the learned function, and $H_\nu$ is the modified Bessel function of the second kind of order $\nu$. When $\nu$ approaches infinity, the Matern kernel becomes equivalent to the RBF kernel, and when $\nu=0.5$ it's equivalent to the absolute exponential kernel. 

Random forest regression~\citep{liaw2002classification} constructs an ensemble of decision trees using a different bootstrap sample of the data for each tree (also called bagging), and selects a random subsets of the features at each candidate split in the decision tree learning process. Gaussian process regression~\citep{rasmussen2006gaussian} defines a collection of random variables, any finite number of which have a joint Gaussian distribution, and then conducts probabilistic inference directly in the function space. Here we choose to use Matern kernels in the Gaussian process regression tests. Artificial neural network is another powerful tool for conducting supervised regression on large datasets. Section~\ref{learning_results} compares the performance of different regression methods and shows that neural networks with appropriate configurations have the best performance in this collision risk regression task, thus we apply them to p-Chekov and compare their performance with the quadrature-based p-Chekov.

\subsection{Risk allocation approach}  \label{IRA}

Using discretizations to estimate trajectory collision probability inevitably faces sensitiveness to the location and number of discrete waypoints. \cite{janson2018monte} addresses this issue by introducing a Monte Carlo Motion Planning (MCMP) approach which solves the deterministic motion planning problem with inflated obstacles and then adjusts the inflation so that the solution trajectory is exactly as safe as desired. However, since MCMP inflates obstacles in the whole planning scene with the same amount, it doesn't account for the different collision probabilities at different locations along the trajectory due to different robot configurations and velocities. Furthermore, MCMP require obstacles with simple geometries, which limits the application of this approach to simple low-dimensional motion planning tasks. \cite{ono2008iterative} addresses the conservative shortcoming of the additive discretization approach through an iterative risk allocation (IRA) algorithm, which divides the whole chance-constrained optimization problem into two stages and seeks the optimal risk allocation that allows for a feasible solution. 

Inspired by the concept of risk allocation and bi-stage motion planning, p-Chekov decomposes the joint chance constraint into individual risk bounds at each time step, and then compares the estimated collision risk with the corresponding risk bound to determine whether the joint chance constraint is satisfied or violated. The planning phase algorithm of p-Chekov starts with a uniform risk allocation and aims at finding a feasible trajectory that can satisfy this specific risk allocation. However, a feasible solution that satisfies this uniform risk allocation might not exist or might need too many iterations to find, thus p-Chekov uses a risk reallocation approach during the planning phase to intelligently speed up the process of finding an initial feasible solution. Since this initial solution can sometimes be overly conservative and highly sub-optimal due to the additive discretization assumption, in the execution phase p-Chekov iteratively improves the trajectory by optimizing the upper stage risk allocation. The risk allocation approaches in p-Chekov planning phase and execution phase are presented in Section \ref{risk_reallocation} and Section \ref{execution_phase_IRA} respectively.

\subsubsection{P-Chekov planning phase risk reallocation}  \label{risk_reallocation}

Risk allocation decomposes a joint chance constraint $\Delta$ by allocating risk bounds $\delta_i$ to individual constraints, where $\sum_1^N \delta_i = \Delta$. The planning phase of p-Chekov starts with a uniform risk allocation and a nominal trajectory from deterministic Chekov which is collision-free in the static environment with no noise. When provided with process noises and observation noises, the collision risk estimation component described in Section~\ref{CP} gives us the collision risk at each waypoint along the nominal trajectory. If the allocated risk bounds are violated at some waypoints, besides adding more constraints to those waypoints, p-Chekov also reallocates the risk bounds to allow for higher collision risks at the violated waypoints. This risk reallocation procedure, as shown in Algorithm~\ref{algo:reallocation}, not only reduces the number of iterations to get initial feasible solutions but also produces less conservative trajectories.

The planning phase risk reallocation relies on the classification of different constraints. Denote the estimated collision risk at waypoint $i$ as $r_i$, and the allocated risk bound as $\delta_i$. When $r_i$ exceeds $\delta_i$, we define the chance constraint at the $i$th waypoint as a violated constraint, otherwise it is viewed as satisfied. Satisfied constraints are divided into active constraints and inactive constraints by introducing a risk tolerance parameter $\eta$. If the difference between $\delta_i$ and $r_i$ is larger than the risk tolerance, we view this underutilized chance constraint as inactive. Otherwise, the constraint is viewed as active. In short, the classification of constraints at different waypoints is as follows:

\begin{equation}
\label{constraints_classification}
 \begin{aligned}
  & \textrm{Constraint Violated:} &  & \delta_i - r_i < 0 \\
  & \textrm{Constraint Satisfied:} & ~~ & \begin{cases} \textrm{Active:} & 0 < \delta_i - r_i < \eta \\
  \textrm{Inactive:} & \delta_i - r_i > \eta \\
  \end{cases}
 \end{aligned}
\end{equation}

Algorithm \ref{algo:reallocation} first identifies inactive constraints where the risk bounds are underutilized, and then takes part of their risk bounds out based on the risk reallocation parameter $\alpha$ (line 1-7). After that, it calculates the total residual risk $\delta_{residual}$ (the total unallocated chance constraint) and the total excessive risk $TotalViolation$ (sum of the risk violation on each violated constraint). It then reallocates $\delta_{residual}$ to each violated constraint proportional to the excessive risk at this waypoint $(r_{p_j} - \delta_{p_j})$ (line 10 - 12). 
The key idea of this risk reallocation method is to take risk from inactive constraints and give it to those violated constraints. This is different from the IRA algorithm introduced by~\cite{ono2008efficient}. IRA requires an initial feasible solution that satisfies the uniform allocation to start with, and reallocates risk from inactive constraints to active constraints. Since IRA doesn't help to find the initial feasible solution, it is only applicable to p-Chekov's execution phase but not the planning phase.

\begin{algorithm}
 \caption{RiskReallocation}
 \label{algo:reallocation}
 \small
 \DontPrintSemicolon
 \KwIn{\\ $r_i$: \text{estimated collision risks at each waypoint;} $i=1, 2, \ldots, N$
 \\ $\delta_i$: \text{risk allocations at each waypoint;} $i=1, 2, \ldots, N$
 \\ $p_j$: \text{waypoint indices where risk allocation is violated}
 \\ $\alpha$: \text{risk reallocation parameter}
 \\ $\Delta$: \text{joint chance constraint for the whole trajectory}
 \\ $\eta$: \text{risk tolerance}
 \\}
 \KwOut{
 \\ $\delta_i^{new}$: \text{new risk allocations for each waypoint;} $i=1, 2, \ldots, N$
 \\}
 
 \For{\forcond $i=1, 2, \ldots, N$}{
 \eIf{$\delta_i - r_i > \eta$}{
 $\delta_i^{new} \leftarrow \alpha\delta_i + (1 - \alpha)r_i$
 }{$\delta_i^{new} \leftarrow \delta_i$}
 }
 $\delta_{residual} = \Delta -\sum_{i=0}^N \delta_i^{new}$  \\
 $TotalViolation \leftarrow $ Sum of excessive risk for all waypoints where collision risk violates the allocated risk bound  \\
 \For{\forcond $j=1, 2, \ldots, N_{violated}$}{
 $\delta_{p_j}^{new} \leftarrow \delta_{p_j} + \delta_{residual}(r_{p_j} - \delta_{p_j})/TotalViolation$
 }
\end{algorithm}

\subsubsection{P-Chekov execution phase iterative risk allocation} \label{execution_phase_IRA}

In the execution phase, p-Chekov adapts the IRA algorithm to improve the trajectory in an anytime fashion, as described in Algorithm~\ref{algo:IRA}. This IRA-based approach takes as input the estimated collision risks and allocated risk bounds for each waypoint from the planning phase, and then determines active constraints using the ActiveContraint() function shown in Algorithm \ref{algo:ac}. It then takes part of the allocated risks for inactive constraints and reallocates them to the active constraints (line 6 - 12). After this risk reallocation procedure, we run the planning phase algorithm with the new risk allocation, and then compare the utility of the new solution $J(\Pi)$ with that of the previous solution $J(\Pi_{previous})$. The algorithm terminates when the improvement is too small. Otherwise, we say IRA effectively improved the solution and repeat this procedure.

\begin{algorithm}
 \caption{IterativeRiskAllocation}
 \label{algo:IRA}
 \small
 \DontPrintSemicolon
 \SetKw{Break}{break}
 \KwIn{\\ $r_i$: \text{estimated collision risks from planning phase;} $i=1, 2, \ldots, N$
 \\ $\delta_i$: \text{risk allocations from planning phase;} $i=1, 2, \ldots, N$
 \\ $\alpha$: \text{risk reallocation parameter}
 \\ $\Delta$: \text{joint chance constraint for the whole trajectory}
 \\ $\eta$: \text{risk tolerance}
 \\ $\epsilon$: \text{convergence tolerance}
 \\}
 \KwOut{\\ $\Pi$: \text{a solution trajectory}
 \\}
 
 $J \leftarrow \infty$ \\
 \While{$|J(\Pi) - J(\Pi_{previous}|) < \epsilon$}{
 $\Pi_{previous} \leftarrow \Pi$ \\
 $N_{active}, ~\mathbf{r} = $ActiveConstraint($\boldsymbol{\delta}$, $\mathbf{r}$) \\
 \eIf{$0<N_{active}<N$}{
 \For{\forcond $i=1, 2, \ldots, N$}{
 \llIf{$\delta_i - r_i > \eta$}{$\delta_i \leftarrow \alpha\delta_i + (1 - \alpha)r_i$}
 }
 $\delta_{residual} = \Delta -\sum_{i=0}^N \delta_i^{new}$  \\
 \ForEach{\forcond $j$ where constraint is active at $j$th waypoint}{
 $\delta_j \leftarrow \delta_j + \delta_{residual}/N_{active}$
 }
 Run p-Chekov planning phase algorithm with new $\boldsymbol{\delta}$ and get new $\mathbf{r}$ associated with the new solution trajectory $\Pi$\\
 }{\Break}
 }
\end{algorithm}

\begin{algorithm}
 \caption{ActiveConstraint}
 \label{algo:ac}
 \small
 \DontPrintSemicolon
 \SetKw{Break}{break}
 \SetKwFunction{ActiveConstraint}{ActiveConstraint}
 \SetKwProg{Fn}{Function}{:}{\KwRet{$N_{active}, ~\mathbf{r}_{previous}$}}
 \Fn{\ActiveConstraint{$\boldsymbol{\delta}$, $\mathbf{r}$}}{
 $N_{active} \leftarrow 0$ \\
 \While{$N_{active} == 0$}{
 $\mathbf{r}_{previous} \leftarrow \mathbf{r}$\\
 \llIf{No constraint to relax}{\Break}\\
 Relax the safety constraint for each waypoint by $d_{step}$ \\
 Find new solution $\Pi$ with planning phase algorithm and re-evaluate collision risk $\mathbf{r}$ \\
 \For{\forcond $i=1, 2, \ldots, N$}{
 \llIf{$r_i > \delta_i$}{$N_{active} \leftarrow N_{active}+1$}}
 }
 }
\end{algorithm}

P-Chekov's execution phase risk allocation optimization approach differs from the original IRA algorithm introduced by~\cite{ono2008efficient} in terms of the active constraint determination method. The way original IRA defines active constraints is the same as the constraint classification method for satisfied constraints in Equation~\ref{constraints_classification}. Here in p-Chekov, however, we use a constraint relaxation approach to find active constraints, as shown in Algorithm \ref{algo:ac}. When ActiveConstraint() is called, it relaxes a small part of the safety constraint for each waypoint and runs the planning phase algorithm again to calculate the new $r_i$ for each waypoint and conducts a risk test (line 7). If some of the risk bounds are violated, they will be viewed as active constraints (line 8 - 10). Otherwise, Algorithm \ref{algo:ac} repeats line 4 - 10 until it detects active constraints.

\subsection{Detailed p-Chekov algorithm illustration} \label{p-algorithm}

Algorithm~\ref{algo:pChekov} summarizes the p-Chekov motion planning and execution system. Line 1 - 5 illustrates the deterministic Chekov planner, which first calls the roadmap planner to find a seed trajectory between the start and the goal. If the roadmap planner fails to find a seed, it returns failure. Otherwise, it calls the trajectory optimizer to locally optimize this seed trajectory. Given this nominal trajectory from the deterministic planner, line 6 calls the state probability distribution estimation algorithm, and line 7 calls one of the collision probability estimation approaches introduced in Section~\ref{CP}. With the estimated collision risk and risk allocation, line 8 conducts a risk test to see whether the risk bounds are satisfied at all waypoints. If all the risk bounds are satisfied, Algorithm~\ref{algo:pChekov} goes to the execution phase and calls the execution phase IRA algorithm (Algorithm~\ref{algo:IRA}) to improve the solution trajectory. Otherwise, the configurations at the violated waypoints will be added as conflicts and new safety constraints will be added at these waypoints. A new risk allocation will be calculated by Algorithm~\ref{algo:reallocation}, and a new solution will be computed from the deterministic planner. This plan improvement procedure will iterate until the chance constraint is satisfied.

\begin{algorithm}[t]
 \caption{P-Chekov}
 \label{algo:pChekov}
 \small
 \DontPrintSemicolon
 \SetKw{Break}{break}
 \SetKw{Return}{return}
 \KwIn{\\ $start$, $goal$: \text{start and goal configuration of the query}
 \\ $\mathcal{R}, \mathcal{E}$: \text{robot and environment collision models respectively}
 \\ $M_t$: \text{covariance matrix of process noises}
 \\ $N_t$: \text{covariance matrix of observation noises}
 \\ $\alpha$: \text{risk reallocation parameter}
 \\ $\Delta$: \text{joint chance constraint for the whole trajectory}
 \\ $\eta$, $\epsilon$: \text{risk tolerance and convergence tolerance}
 \\ $d_{step}$: \text{step size for penalty hit-in distance increase}
 \\}
 \KwOut{\\ $\Pi$: \text{a solution trajectory}
 \\}
 
 $seed$ = RoadmapFindSolution($start, goal$) \\
 \eIf{$seed$ is not $None$}{
 Initialize risk allocation $\boldsymbol{\delta}$ with uniform allocation \\
 Initialize list of conflicts $Clist$ to be empty \\
 $\Pi$ = Optimizer($seed$, $Clist$) \\
 $\mathcal{D} = $StateEstimation($\Pi$, $M_t$, $N_t$) \\
 $\mathbf{r}$ = CollisionProbabilityEstimation($\Pi, \mathcal{D}, \mathcal{R}, \mathcal{E}$) \\
 $violation$ = RiskTest($\mathbf{r}, \boldsymbol{\delta}$) \\
 \While{$violation$ is $True$}{
 \ForEach{\forcond waypoint $i$ where risk bound is violated}{
 Add the configuration at waypoint $i$ to $Clist$ \\
 }
 $\boldsymbol{\delta}$ = RiskReallocation($\mathbf{r}, \boldsymbol{\delta}, \alpha, \Delta, \eta$) \\
 $\Pi$ = Optimizer($\Pi$, $Clist$) \\
 $violation$ = RiskTest($\mathbf{r}, \boldsymbol{\delta}$) \\
 }
 Chance constraint satisfied, start execution \\
 $\Pi$ = IterativeRiskAllocation($\mathbf{r}, \boldsymbol{\delta}, \Pi, \alpha, \Delta, \eta, \epsilon$) \\
 Execute the updated trajectories from IRA \\
 \Return Success
 }
 {\Return Failure}
\end{algorithm}

The main innovation of p-Chekov includes the fast-reactive deterministic Chekov planner that can generate nominal trajectories for high-dimensional robots in real-time, as well as the idea of risk allocation which plays the role of extracting conflicts and guiding the deterministic planner to approach to a feasible solution whose execution failure rate is bounded by the chance constraint. In addition, the application of quadrature-rule and supervised learning techniques in collision risk estimation is key to the speed of p-Chekov's convergence to a feasible solution trajectory. 

\section{Chance-constrained motion planning experiments}   \label{results}

To demonstrate p-Chekov's performance, 500 pairs of start and goal poses in each of the two tabletop environments introduced in Section~\ref{implementation}, the ``tabletop with a pole'' environment and the ``tabletop with a container'' environment, are used for simulation experiments. Note that the second environment not only has the narrow spaces inside the container which are difficult for chance-constrained motion planners, but also include difficult test cases where the robot joints are close to their limits. Section~\ref{modeling} describes the dynamics and observation models used in the experiments, Section~\ref{risk_aware_results} shows the performance of p-Chekov with collision estimation module based on the Guass-Hermite quadrature rule, and Section~\ref{learning_results} demonstrates the experiments on p-Chekov with the learning-based collision estimation module.

\subsection{Experiment modeling} \label{modeling}

We simplify manipulator dynamics into a discrete-time linear time-invariant dynamics model and use accelerations as control inputs at each time step. All the joints are assumed to be fully actuated and independent from each other, corrupted by process noise $\mathbf{m}_{t, j} \sim \mathcal{N}(0, M_{t, j})$, where $j = 1, 2, \ldots, 7$ denotes the degree of freedom (DOF) index, and

\begin{equation}
 M_{t, j} = \begin{bmatrix}
        \sigma_{x, j}^2  &  0 \\
        0  &  \sigma_{v, j}^2 \\
       \end{bmatrix}.
\end{equation}

\noindent Using the linearization from Equation~\ref{deviation}, we have:

\begin{equation}
 \bar{\mathbf{x}}_{t, j} = \begin{bmatrix}
                 1 & \Delta T \\
                 0 & 1 \\
                \end{bmatrix}
 \bar{\mathbf{x}}_{t-1, j} + \begin{bmatrix}
                    \Delta T^2/2  \\ \Delta T \\
                    \end{bmatrix}
 \bar{\mathbf{u}}_{t-1, j} + \mathbf{m}_{t, j},
\end{equation}

\noindent where $\bar{\mathbf{x}}_{t, j}$ includes the position and velocity of the $j$th joint at time step $t$. We consider two different system observation models in this paper: a joint configuration observation model and an end-effector pose observation model.

\subsubsection{Joint configuration observation model}

One natural way of formulating the system observation model is to observe the joint values directly through joint encoders. We assume the value of each joint is corrupted by Gaussian observation noises from the corresponding joint encoder, and the noise at each joint is independent from each other. With this fully observable model, all the joints are decoupled from each other in both the dynamics model and the observation model, which helps reduce the computation complexity of state probability distribution estimation. The observation model at each joint can be expressed as:

\begin{equation}
\begin{aligned}
 & \bar{\mathbf{z}}_{t, j} = \bar{\mathbf{x}}_{t, j} + \mathbf{n}_{t, j}, & \mathbf{n}_{t, j} \sim \mathcal{N}(0, N_{t, j}),
\end{aligned}
\end{equation}

\noindent where $N_{t, j}$ is the noise covariance matrix of the $j$th joint encoder.

\subsubsection{End-effector pose observation model}

Although the joint configuration observation model is very straightforward, in practice the joint encoder noises are usually not the most significant source of errors. In comparison, camera observations are often less accurate due to the inaccuracy of camera itself and the uncertainties from the object it is mounted to. For manipulators mounted on mobile robots, for example, their head camera is often an important source of observations. However, unexpected movements of the mobile base caused by arm movements or external disturbances can often lead to inaccurate estimations of the relative position between the manipulator and the object to be grasped in a pick-and-place task. In addition, in underwater manipulation tasks, vehicle movements are inevitable due to movements of the manipulator and disturbances from ocean currents. In this case, the observations of the spatial relationship between obstacles and the manipulator from cameras mounted on the vehicle will inevitably be corrupted. As a result, it is of more practical significance to incorporate camera observations compared to using the fully observable joint configuration observation model. 

Ideally, observations of the whole manipulator should be evaluated. However, this is nontrivial since it requires modeling the forward kinematics mapping of all the points on each link. In addition, directly modeling the observation noises for the relative spatial relationship between the entire manipulator and workspace objects is also difficult. Thus as a start, an end-effector observation model is introduced in this section to approximate the real-world camera observations. The transformation matrix between workspace objects and the end-effector can be expressed as:

\begin{equation}
 T_{obj}^{ee} = T_{obj}^{cam} \cdot T_{cam}^{ee},
\end{equation}

\noindent where $T_{obj}^{cam}$ is the transformation from the workspace object to the camera frame, and $T_{cam}^{ee}$ is the transformation from the camera frame to the end-effector. Therefore, the noises for observing $T_{obj}^{ee}$ can be transformed into observation noises for $T_{cam}^{ee}$ through the transformation matrix $T_{obj}^{cam}$. Then $T_{cam}^{ee}$ can be transformed into $T_{ee}^{cam}$ through matrix inversion. Therefore, we can approximate the observation noises through corrupted observations of the end-effector pose from the camera.

The observations of the end-effector can be expressed in C-space through the nonlinear relationship:

\begin{equation}
\begin{aligned}
& \mathbf{z}_t = h(\mathbf{x}_t, \mathbf{n}_t), & & \mathbf{n}_t \sim \mathcal{N}(0, N_t),
\end{aligned}
\end{equation}

\noindent where $h(\mathbf{x}_t, 0)$ is the forward kinematics, $\mathbf{n}_t$ is the observation noise, and $N_t$ is the covariance matrix of the observation noise. The linearization of this observation model around a nominal configuration $\mathbf{x}^*_t$ can be expressed as:

\begin{equation}  
\label{Jacobian}
\begin{aligned}
& \mathbf{z}_t - h(\mathbf{x}^*_t, 0) = J_t(\mathbf{x}_t - \mathbf{x}^*_t) + W_t\mathbf{n}_t,
\end{aligned}
\end{equation}

\noindent where

\begin{equation}
 \begin{aligned}
  & J_t = \frac{\partial h}{\partial \mathbf{x}}(\mathbf{x}^*_t, 0). \\
 \end{aligned}
\end{equation}

\noindent Since $h(\mathbf{x}_t, 0)$ is the forward kinematics, $J_t$ is the end-effector Jacobian matrix at the nominal configuration $\mathbf{x}^*_t$. In this way, the linearized system observation matrix becomes the Jacobian matrix, which is usually easy to obtain during computation. Again using the linearization from Equation~\ref{deviation}, the end-effector pose observation model given in Equation~\ref{Jacobian} can be rewritten as:

\begin{equation}
\label{observation}
\begin{aligned}
& \bar{\mathbf{z}}_t = J_t\bar{\mathbf{x}}_t + W_t\mathbf{n}_t, & & \mathbf{n}_t \sim \mathcal{N}(0, N_t)
\end{aligned}
\end{equation}

Compared with the joint configuration observation model, this end-effector observation model no longer decouples different joints, thus it will inevitably require more computation time in the state probability distribution estimation step. In addition, since this is a partially observable model, estimated noise variances will grow as the robot executes along the desired trajectory. Hence we expect that it will be more difficult for p-Chekov to find solutions that satisfy the chance constraint using this end-effector observation model. Section~\ref{risk_aware_results} will compare quadrature-based p-Chekov's performance with these two different observation models empirically, whereas Section~\ref{learning_results} will focus only on the results for the end-effector observation model when testing the learning-based p-Chekov since it is the more realistic yet challenging one.

\subsection{Quadrature-based p-Chekov experiment results}    \label{risk_aware_results}

We focus on evaluating the initial feasible solution returned by the p-Chekov planning phase algorithm in Section~\ref{planning_results}, and then look at the improvement the execution phase IRA algorithm induces in Section~\ref{execution_results}. Baxter's specification indicates that its worst case accuracy of joints is $\pm 0.25$ degree, which is about $\pm 0.0044$ rad. Hence in the experiments in this paper, the standard deviation of noises during execution is set to 0.0044~rad. The collision risk of a returned solution trajectory is evaluated with 100 noisy executions.

To assess the chance constraint satisfaction performance of p-Chekov, we provide the definition of \emph{chance constraint satisfied test cases}. If p-Chekov works perfectly, the 100 independent executions for a particular solution trajectory should all have their probability of collision equal to the chance constraint. For example, if the chance constraint allows for a 10\% collision probability, the probability of collision happening during an execution should be 10\%. Then the number of failures out of the 100 executions follows a binomial distribution with the number of independent experiments $n = 100$ and the probability of occurrence in each experiment $p = 0.1$. The cumulative probability distribution function of binomial distributions can be expressed as:

\begin{equation}
\label{eq:binomial}
 F(k; n, p) = \textrm{Pr}(X \leq k) = \displaystyle\sum_{i}^{k} {n\choose i} p^i(1-p)^{n-i}
\end{equation}

\noindent For $n=100$ and $p=0.1$, we can calculate from Equation~\ref{eq:binomial} that the probability of having less than or equal to 10 failures out of 100 executions is only about 56\%. Similarly, if the chance constraint is 5\%, then the probability of having less than or equal to 5 failures in 100 executions is about 59\%. However, to better represent the actual collision risk of solutions returned by p-Chekov, we want the classification error for \emph{chance constraint satisfied test cases} to be small, so that we are confident to say the test case has violated the chance constraint when there are more than the corresponding number of executions end up in collision. If we define chance constraint satisfied test cases as the ones where the collision rate out of 100 executions is lower than or equal to 1.5 times of the chance constraint, Equation~\ref{eq:binomial} shows that for $p=0.1$ the classification accuracy is about 94\%, and for $p=0.05$ the accuracy is around 86\%. Consequently, we decide to use 1.5 times of the chance constraint as the boundary between chance constraint satisfied cases and chance constraint violated cases. 

Since theoretically p-Chekov only has probabilistic guarantees for waypoints instead of the entire trajectory, we distinguish between continuous-time and discrete-time chance constraint satisfaction performances. If the 100 noisy executions of a test case shows that the average continuous-time (or waypoint) collision rate is within 1.5 times of the collision chance constraint, then we say this test case satisfies the continuous-time (or discrete-time) chance constraint. Only the continuous-time satisfaction is the true criterion for success, but we use discrete-time performance to show the impact of edge collisions, i.e. the collisions in between waypoints.

\subsubsection{Planning phase experiment results}  \label{planning_results}

Table~\ref{table:00044noise} and Table~\ref{table:tabletop2} show quadrature-based p-Chekov's performance with different chance constraints using joint configuration observation model in the ``tabletop with a pole'' environment and the ``tabletop with a container" environment respectively. The first six rows of Table~\ref{table:00044noise} and~\ref{table:tabletop2} compare deterministic Chekov and the quadrature-based p-Chekov planning phase algorithm. As expected, p-Chekov doesn't perform as well as deterministic Chekov in terms of planning time and the average length of execution trajectories, because p-Chekov usually pushes the solution away from the locally optimal solution deterministic Chekov returned in order to ensure safety. However, the overall collision rate (averaged over 500 test cases with 100 noisy executions each) shows the superiority of p-Chekov solutions in the presence of noises. From Table~\ref{table:00044noise} we can see that the overall collision rate is reduced by more than 20\% compared with deterministic solutions, while the average path length is only increased by 0.3 rad. Since the ``tabletop with a container'' environment is much more complicated due to the narrow spaces, p-Chekov's performance shown in Table~\ref{table:tabletop2} is much worse compared to Table~\ref{table:00044noise}. Despite the difficulty in this environment, p-Chekov can reduce the collision rate by about 30\% with both chance constraints.

\begin{table}
\caption{Quadrature-based P-Chekov in Tabletop with a Pole Environment with Joint Observation and Various Chance Constraints}
\label{table:00044noise}
\footnotesize\sf\centering
\begin{threeparttable}
\begin{tabular}{L{1.66cm}|L{1.2cm}|L{2.94cm}|L{0.77cm} L{0.77cm}}
\toprule
 \multicolumn{3}{c|}{Chance Constraint}  &  10\%  & 5\%    \\ 

\hline
\multirow{2}{1.7cm}{\centering Planning Time (s)} 
& \multicolumn{2}{c|}{deterministic Chekov} & 1.10  & 1.38  \\ 
& \multicolumn{2}{c|}{p-Chekov} & 5.09 & 6.44  \\ 

\hline
\multirow{2}{1.7cm}{\centering Overall Collision Rate\tnote{1}} 
& \multicolumn{2}{c|}{deterministic Chekov} & 33.82\%  & 33.84\%  \\ 
& \multicolumn{2}{c|}{p-Chekov} & 7.70\% & 7.63\%  \\ 

\hline
\multirow{2}{1.7cm}{\centering Average Path Length (rad)\tnote{2}} 
& \multicolumn{2}{c|}{deterministic Chekov} & 0.51  & 0.51  \\ 
& \multicolumn{2}{c|}{p-Chekov} & 0.54  & 0.54  \\

\hline
\multirow{14}{1.7cm}{\centering P-Chekov Performance} 
& \multicolumn{2}{m{4.4cm}|}{continuous chance constraint satisfaction rate\tnote{3}} & 91.57\% & 91.57\%  \\ \cline{2-5}
& \multirow{3}{1.3cm}{continuous satisfied cases\tnote{4}} & average iteration number & 1.41 & 1.47 \\ 
&  & average collision rate & 0.02\%  & 0.01\%  \\ 
&  & average risk reduction\tnote{9} & 0.32 & 0.32  \\ \cline{2-5}
& \multirow{3}{1.3cm}{continuous violated cases\tnote{5}} & average iteration number & 2.82  & 2.93  \\
&  & average collision rate & 86.89\%  & 86.30\%  \\ 
&  & average risk reduction & -0.35  & -0.33   \\ \cline{2-5}

& \multicolumn{2}{m{4.4cm}|}{discrete chance constraint satisfaction rate\tnote{6}} & 93.57\% & 92.77\%  \\ \cline{2-5}
& \multirow{3}{1.3cm}{discrete satisfied cases\tnote{7}} & average iteration number & 1.47 & 1.52 \\ 
&  & average collision rate & 0.04\%  & 0.01\% \\ 
&  & average risk reduction & 0.23 & 0.22  \\ \cline{2-5}
& \multirow{3}{1.3cm}{discrete violated cases\tnote{8}} & average iteration number & 2.50  & 2.64  \\
&  & average collision rate & 84.79\%  & 83.14\%  \\ 
&  & average risk reduction & -0.40  & -0.35   \\ 
\bottomrule
\end{tabular}
\begin{tablenotes}
\footnotesize
 \item[1] Average collision rate over 100 noisy executions for all 500 test cases.
 \item[2] Average length of actual execution trajectories.
 \item[3] Percentage of test cases where the average continuous-time collision rate over 100 noisy executions satisfies the chance constraint.
 \item[4] P-Chekov performance over the test cases where the chance constraint is satisfied by continuous-time collision rate (viewed as success cases).
 \item[5] P-Chekov performance over the test cases where the chance constraint is violated by continuous-time collision rate (viewed as failure cases).
 \item[6] Percentage of test cases where the average waypoint collision rate over 100 noisy executions satisfies the chance constraint.
 \item[7] P-Chekov performance over the test cases where the chance constraint is satisfied by waypoint collision rate.
 \item[8] P-Chekov performance over the test cases where the chance constraint is violated by waypoint collision rate.
 \item[9] The difference between the average collision rate of p-Chekov solutions and that of deterministic Chekov solutions.
 \end{tablenotes}
 \end{threeparttable}
\end{table}

\begin{table}
\caption{Quadrature-based P-Chekov in Tabletop with a Container Environment with Joint Observation and Various Chance Constraints}
\label{table:tabletop2}
\footnotesize\sf\centering
\begin{threeparttable}
\begin{tabular}{L{1.66cm}|L{1.2cm}|L{2.94cm}|L{0.77cm} L{0.77cm}}
\toprule
 \multicolumn{3}{c|}{Chance Constraint}  &  10\%  & 5\%   \\ 

\hline
\multirow{2}{1.7cm}{\centering Planning Time (s)} 
& \multicolumn{2}{c|}{deterministic Chekov} & 1.29  & 1.61 \\ 
& \multicolumn{2}{c|}{p-Chekov} & 15.47 & 19.80 \\ 

\hline
\multirow{2}{1.7cm}{\centering Overall Collision Rate} 
& \multicolumn{2}{c|}{deterministic Chekov} & 66.56\%  & 66.75\%  \\ 
& \multicolumn{2}{c|}{p-Chekov} & 36.29\% & 36.83\%   \\ 

\hline
\multirow{2}{1.7cm}{\centering Average Path Length (rad)} 
& \multicolumn{2}{c|}{deterministic Chekov} & 0.63  & 0.63  \\ 
& \multicolumn{2}{c|}{p-Chekov} & 0.76  & 0.77  \\

\hline
\multirow{7}{1.7cm}{\centering P-Chekov Performance} 
& \multicolumn{2}{m{4.4cm}|}{continuous chance constraint satisfaction rate\tnote{3}} & 61.30\% & 60.49\%  \\ \cline{2-5}
& \multirow{3}{1.3cm}{continuous satisfied cases} & average iteration number & 2.74 & 2.87 \\ 
&  & average collision rate & 0.09\%  & 0.05\%  \\ 
&  & average risk reduction & 0.54 & 0.54 \\ \cline{2-5}
& \multirow{3}{1.3cm}{continuous violated cases} & average iteration number & 6.28  & 6.38  \\
&  & average collision rate & 93.64\%  & 92.66\%  \\ 
&  & average risk reduction & -0.08  & -0.07  \\ \cline{2-5}

& \multicolumn{2}{m{4.4cm}|}{discrete chance constraint satisfaction rate\tnote{6}} & 69.65\% & 69.04\%  \\ \cline{2-5}
& \multirow{3}{1.3cm}{discrete satisfied cases} & average iteration number & 3.58 & 3.78 \\ 
&  & average collision rate & 0.05\%  & 0.04\%  \\ 
&  & average risk reduction & 0.50 & 0.50 \\ \cline{2-5}
& \multirow{3}{1.3cm}{discrete violated cases} & average iteration number & 5.31  & 5.36  \\
&  & average collision rate & 92.80\%  & 92.09\%  \\ 
&  & average risk reduction & -0.14  & -0.13  \\ 
\bottomrule
\end{tabular}
 \end{threeparttable}
\end{table}

The remaining rows of Table~\ref{table:00044noise} and~\ref{table:tabletop2} focus on the chance constraint satisfaction performance of p-Chekov. From both tables we can see that the discrete and continuous chance constraint satisfaction performances are very close, which means edge collisions in these experiments don't have significant influence. Comparing p-Chekov's performance in the continuous chance constraint satisfied cases and violated cases, we can see that the satisfied cases take much fewer iterations than the violated cases and also have much lower average collision rate. In addition, in the satisfied cases p-Chekov successfully reduces the average collision rate by 0.3 - 0.5, meanwhile in the violated cases the collision risk actually increased. This means in the violated cases, p-Chekov is failing to find safe solutions that satisfy the chance constraint, and might get the trajectories close to other objects while pushing them away from some obstacles. For the chance constraint satisfied cases, in contrast, the collision rate is much lower than the chance constraint, meaning that p-Chekov is overly conservative. This is potentially caused by the conservative quadrature-based collision probability estimation approach and the conservative risk allocations. Section~\ref{execution_results} will show p-Chekov's performance with the execution phase IRA algorithm, which aims at providing less conservative solutions.

If we compare p-Chekov's performance with different chance constraints, we can see that the overall collision rate is not necessarily going down when the chance constraint decreases. One possible cause for this is, when the chance constraint is getting tighter, more test cases become infeasible, thus the solutions p-Chekov found for those cases are likely to end up in collision. This shows the importance of filtering out infeasible test cases in order to better see p-Chekov's performance, which we will present later in this section.

\begin{figure}[t]
      \centering
      \begin{subfigure}[t]{0.5\textwidth}
      \includegraphics[width=1.08\linewidth]{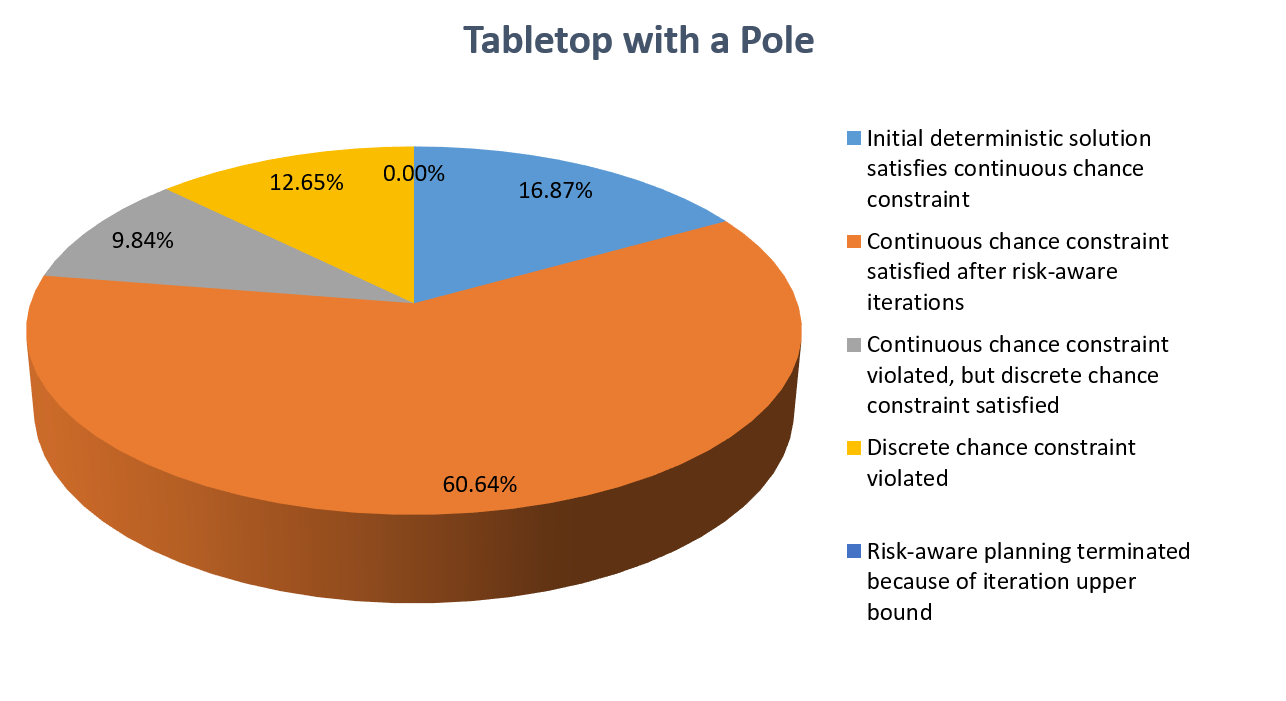}
      \end{subfigure}
      \begin{subfigure}[t]{0.5\textwidth}
      \includegraphics[width=1.08\linewidth]{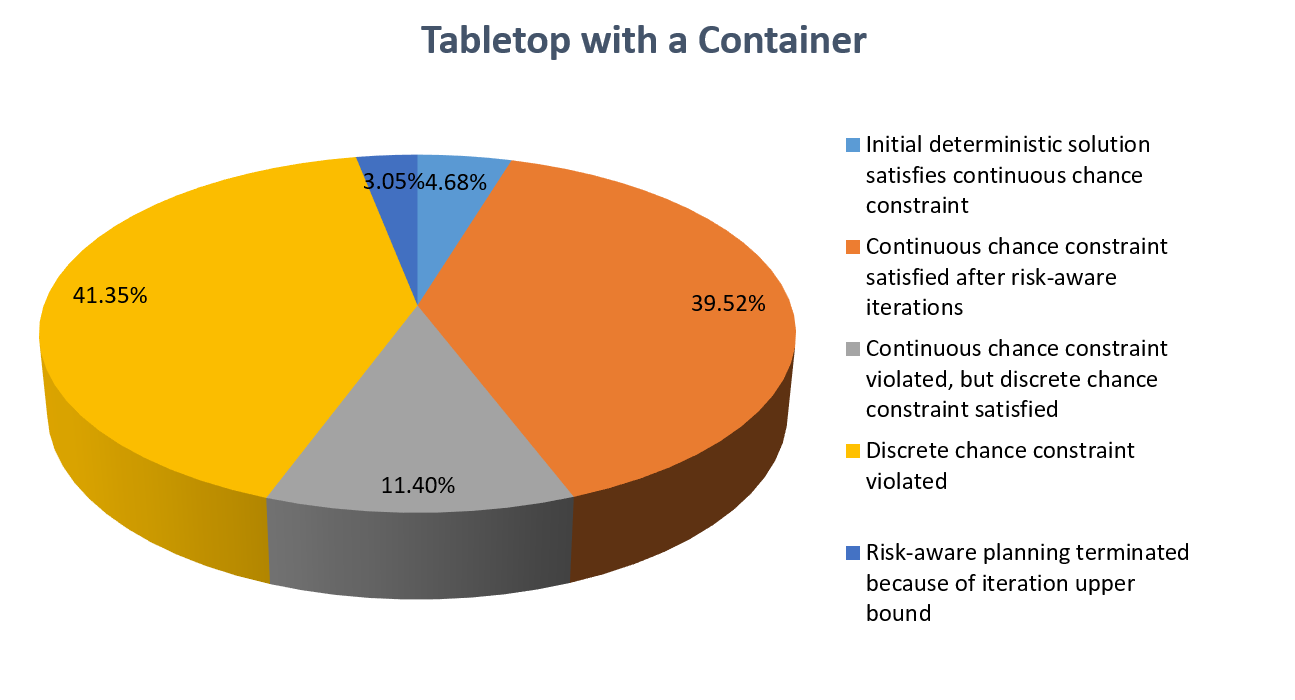}
      \end{subfigure}
      \caption{Quadrature-based p-Chekov statistics breakdown for experiments with end-effector observation, 0.0044 noise standard deviation and 10\% chance constraint}
      \label{fig:breakdown_1}
   \end{figure}

As described in Section~\ref{modeling}, the partially-observable end-effector observation model is more difficult but more realistic in practical applications. We now investigate p-Chekov's performance with this end-effector observation model.
Figure~\ref{fig:breakdown_1} shows the experiment result breakdown in the ``tabletop with a pole'' environment and the ``tabletop with a container'' environment respectively, with 10\% chance constraint and noise level 0.0044 rad. The test cases are divided into five groups: (1) chance constraint is satisfied by the initial deterministic Chekov solution, (2) continuous-time collision rate satisfies the chance constraint, (3) continuous-time collision rate violates the chance constraint but discrete-time collision rate satisfies it, (4) discrete-time collision rate violates the chance constraint but the p-Chekov algorithm terminated before it hits its iteration number upper bound, and (5) p-Chekov terminates because it hit the iteration limit.

\begin{table*}
\caption{Quadrature-based P-Chekov in Two Environments with End-effector Observation, 10\% Chance Constraint and Various Noise Levels}
\label{table:ee_01}
\footnotesize\sf\centering
\begin{threeparttable}
\begin{tabular}{L{2.2cm}|L{2.2cm}|L{3.2cm}|L{1cm} L{1cm} L{1cm}|L{1cm} L{1cm} L{1cm}}
\toprule
 \multicolumn{3}{c|}{Environment}  & \multicolumn{3}{c|}{Tabletop with a Pole}  & \multicolumn{3}{c}{Tabletop with a Container}  \\ \hline
 \multicolumn{3}{c|}{Noise Standard Deviation (rad)}  &  0.0044  & 0.0022  & 0.0011 &  0.0044  & 0.0022  & 0.0011 \\ 

\hline
\multirow{2}{2.2cm}{\centering Planning Time (s)} 
& \multicolumn{2}{c|}{deterministic Chekov} & 1.13  & 1.36 & 1.11 & 1.31  & 1.52 & 1.24 \\ 
& \multicolumn{2}{c|}{p-Chekov} & 24.17 & 16.08 & 8.38  & 49.60 & 34.67 & 17.47 \\ 

\hline
\multirow{2}{2.2cm}{\centering Overall Collision Rate} 
& \multicolumn{2}{c|}{deterministic Chekov} & 35.88\%  & 34.27\% & 33.40\%  & 66.95\%  & 65.77\% & 65.04\% \\ 
& \multicolumn{2}{c|}{p-Chekov} & 21.46\% & 17.42\% & 14.12\%   & 53.11\% & 48.82\% & 43.51\%  \\ 

\hline
\multirow{2}{2.2cm}{\centering Average Path Length (rad)} 
& \multicolumn{2}{c|}{deterministic Chekov} & 0.51  & 0.51 & 0.51 & 0.63  & 0.63 & 0.63 \\ 
& \multicolumn{2}{c|}{p-Chekov} & 0.75  & 0.63 & 0.57  & 1.11  & 0.91 & 0.76 \\

\hline
\multirow{14}{2.2cm}{\centering P-Chekov Performance} 
& \multicolumn{2}{m{5.4cm}|}{continuous chance constraint satisfaction rate} & 77.51\% & 81.33\% & 85.34\%  & 44.20\% & 49.08\% & 54.79\% \\ \cline{2-9}
& \multirow{3}{2.2cm}{continuous satisfied cases} & average iteration number & 5.41 & 3.57 & 2.25 & 6.90 & 4.81 & 3.50  \\ 
&  & average collision rate & 0.13\%  & 0.05\% & 0.02\%  & 0.20\%  & 0.11\% & 0.10\% \\ 
&  & average risk reduction & 0.30 & 0.29 & 0.30  & 0.46 & 0.48 & 0.48 \\ \cline{2-9}
& \multirow{3}{2.2cm}{continuous violated cases} & average iteration number & 9.48  & 6.63 & 5.01 & 12.12  & 9.11 & 6.26 \\
&  & average collision rate & 91.70\%  & 93.05\% & 96.21\%  & 95.01\%  & 95.02\% & 95.25\% \\ 
&  & average risk reduction & -0.37  & -0.37 & -0.40  & -0.12  & -0.12 & -0.11  \\ \cline{2-9}

& \multicolumn{2}{m{5.4cm}|}{discrete chance constraint satisfaction rate} & 87.35\% & 86.55\% & 88.96\% & 55.60\% & 59.27\% & 62.32\% \\ \cline{2-9}
& \multirow{3}{2.2cm}{discrete satisfied cases} & average iteration number & 5.91 & 3.80 & 2.43 & 7.87 & 5.74 & 3.83 \\ 
&  & average collision rate & 0.12\%  & 0.06\% & 0.05\% & 0.16\%  & 0.33\% & 0.14\% \\ 
&  & average risk reduction & 0.25 & 0.22 & 0.21 & 0.47 & 0.45 & 0.43 \\ \cline{2-9}
& \multirow{3}{2.2cm}{discrete violated cases} & average iteration number & 8.99  & 6.32 &4.39 & 12.20  & 8.84 & 6.27 \\
&  & average collision rate & 74.10\%  & 85.76\% & 93.02\%  & 88.27\%  & 92.54\% & 94.60\%  \\ 
&  & average risk reduction & -0.26  & -0.41 & -0.48  & -0.11  & -0.17 & -0.16  \\ 
\bottomrule
\end{tabular}
 \end{threeparttable}
\end{table*}

\begin{table*}
\caption{Quadrature-based P-Chekov in Tabletop with a Pole Environment with End-effector Observation, 0.0044 Noise Level and 10\% Chance Constraint}
\label{table:ee_02}
\footnotesize\sf\centering
\begin{threeparttable}
\begin{tabular}{L{3.5cm}|L{2.2cm}|L{3.2cm}|L{1.2cm} L{1.2cm} L{1.2cm} L{1.2cm}}
\toprule
 \multicolumn{3}{c|}{Step Size of the Collision Penalty Hit-in Distance Increase (m)}  &  0.03  & 0.04  & 0.05 & 0.06  \\ 

\hline
\multirow{2}{3.5cm}{\centering Planning Time (s)} 
& \multicolumn{2}{c|}{deterministic Chekov} & 1.35  & 1.39 & 1.13 & 1.35 \\ 
& \multicolumn{2}{c|}{p-Chekov} & 51.92 & 37.00 & 24.17 & 26.08 \\ 

\hline
\multirow{2}{3.5cm}{\centering Overall Collision Rate} 
& \multicolumn{2}{c|}{deterministic Chekov} & 35.89\%  & 35.87\% & 35.88\% & 36.05\% \\ 
& \multicolumn{2}{c|}{p-Chekov} & 21.95\% & 21.82\% & 21.46\% & 23.48\% \\ 

\hline
\multirow{2}{3.5cm}{\centering Average Path Length (rad)} 
& \multicolumn{2}{c|}{deterministic Chekov} & 0.51  & 0.51 & 0.51 & 0.51 \\ 
& \multicolumn{2}{c|}{p-Chekov} & 0.72  & 0.74 & 0.75 & 0.78 \\

\hline
\multirow{14}{3.5cm}{\centering P-Chekov Performance} 
& \multicolumn{2}{p{5.4cm}|}{continuous chance constraint satisfaction rate} &  76.91\% &  76.71\% & 77.51\% & 74.50\% \\ \cline{2-7}
& \multirow{3}{2.2cm}{continuous satisfied cases} & average iteration number & 8.04 & 6.61 & 5.41 & 4.41 \\ 
&  & average collision rate & 0.11\%  & 0.17\% & 0.13\% & 0.18\% \\ 
&  & average risk reduction & 0.30 & 0.30 & 0.30 & 0.29 \\ \cline{2-7}
& \multirow{3}{2.2cm}{continuous violated cases} & average iteration number & 13.87  & 11.38 & 9.48 & 8.57 \\
&  & average collision rate & 92.28\%  & 92.32\% & 91.70\% & 90.13\% \\ 
&  & average risk reduction & -0.38  & -0.37 & -0.37 & -0.34 \\ \cline{2-7}

& \multicolumn{2}{p{5.4cm}|}{discrete chance constraint satisfaction rate} &  85.94\% &  84.74\% & 87.35\% & 83.73\% \\ \cline{2-7}
& \multirow{3}{2.2cm}{discrete satisfied cases} & average iteration number & 8.93 & 7.04 & 5.91 & 4.87 \\ 
&  & average collision rate & 0.19\%  & 0.16\% & 0.12\% & 0.11\% \\ 
&  & average risk reduction & 0.25 & 0.24 & 0.25 & 0.24 \\ \cline{2-7}
& \multirow{3}{2.2cm}{discrete violated cases} & average iteration number & 12.27  & 11.27 & 8.99 & 8.62 \\
&  & average collision rate & 75.30\%  & 79.85\% & 74.10\% & 77.45\% \\ 
&  & average risk reduction & -0.28  & -0.30 & -0.26 & -0.29 \\ 
\bottomrule
\end{tabular}
 \end{threeparttable}
\end{table*}

From Figure~\ref{fig:breakdown_1} we can see that in 60.64\% of the test cases in the ``tabletop with a pole'' environment, the chance constraint is satisfied through the risk-aware p-Chekov's effort, meanwhile in 9.84\% of the test cases this constraint is violated by the continuous-time collision rate but satisfied by the waypoint collision rate. As mentioned previously, edge collision is one of the drawbacks of trajectory discretization, thus we need to balance the computation complexity and plan safety when deciding the number of waypoints. Figure~\ref{fig:breakdown_1} also shows that in a small portion (16.87\%) of the test cases, the deterministic Chekov solutions have already satisfied the chance constraint and no p-Chekov iterations are needed. No cases hit p-Chekov's iteration upper bound, while 12.65\% test cases are failures caused by other reasons than edge collisions. The ``tabletop with a container'' environment shows a similar breakdown, where 11.40\% of the test cases fail because of edge collisions, and 41.35\% fail because of other reasons. We picked some test cases out of this ``other failures'' category to closely inspect the failure reason, and noticed that most of these test cases have either start or goal pose very close to obstacles. This means a lot of these cases might be infeasible because the start or goal collision probability has already violated the chance constraint, which makes the chance-constrained query infeasible. 

Table~\ref{table:ee_01} demonstrates detailed performance of quadrature-based p-Chekov with end-effector pose observations under different levels of noise disturbance. The same 500 test cases in each environment are evaluated, and the chance constraint is also set to 10\%. Compared with Table~\ref{table:00044noise}, we can see that end-effector pose observation model makes it much more difficult for p-Chekov to find feasible solutions that satisfy the chance constraint. This is expected because, in contrast to fully-observable models whose variance estimation will converge, end-effector model doesn't have full information of robot states and the variance estimation will keep growing along the trajectory, leading to p-Chekov's failure in finding feasible solutions in more test cases. Additionally, Table~\ref{table:ee_01} shows a larger difference between continuous-time and discrete-time chance constraint satisfaction performance compared to Table~\ref{table:00044noise}, meaning that edge collisions occur in more test cases. In Table~\ref{table:ee_01}, the overall collision rate is reduced at the expense of average execution trajectory length. Comparing the results for different noise levels, we can see that the deterministic Chekov solutions have similar collision rates but the p-Chekov solutions collide much less with lower noise levels. In the constraint satisfied test cases, p-Chekov is taking many more iterations when the noise level is high, but the average risk reductions with different noise levels are similar. In the ``tabletop with a container'' environment, both the overall collision rates and the chance constraint satisfaction rates under all the three noise levels are only about 50\%. Despite the high overall collision rates, p-Chekov still successfully reduced the collision risk by over 0.45 in the constraint satisfied test cases. 

In the experiments presented in this section, the constraints added to the waypoints where the allocated risk bound is violated include an increase in the collision penalty hit-in distance. Hence the step size of this increase could potentially influence both the conservativeness of the solution trajectory and the number of iterations it takes to find a feasible solution. We compared p-Chekov's performance with four different step sizes in the ``tabletop with a pole'' environment and show the results in Table~\ref{table:ee_02}. The chance constraint is set to 10\% and the standard deviation of noises is 0.0044 rad. Different columns in Table~\ref{table:ee_02} show that using a smaller penalty distance increase step doesn't make a big difference in p-Chekov's performance except for a longer planning time. Therefore, in the experiments in this paper, we set the step size of the collision penalty hit-in distance increase to 0.05 m.

   \begin{figure}
      \centering
      \begin{subfigure}[t]{0.5\textwidth}
      \includegraphics[width=1.08\linewidth]{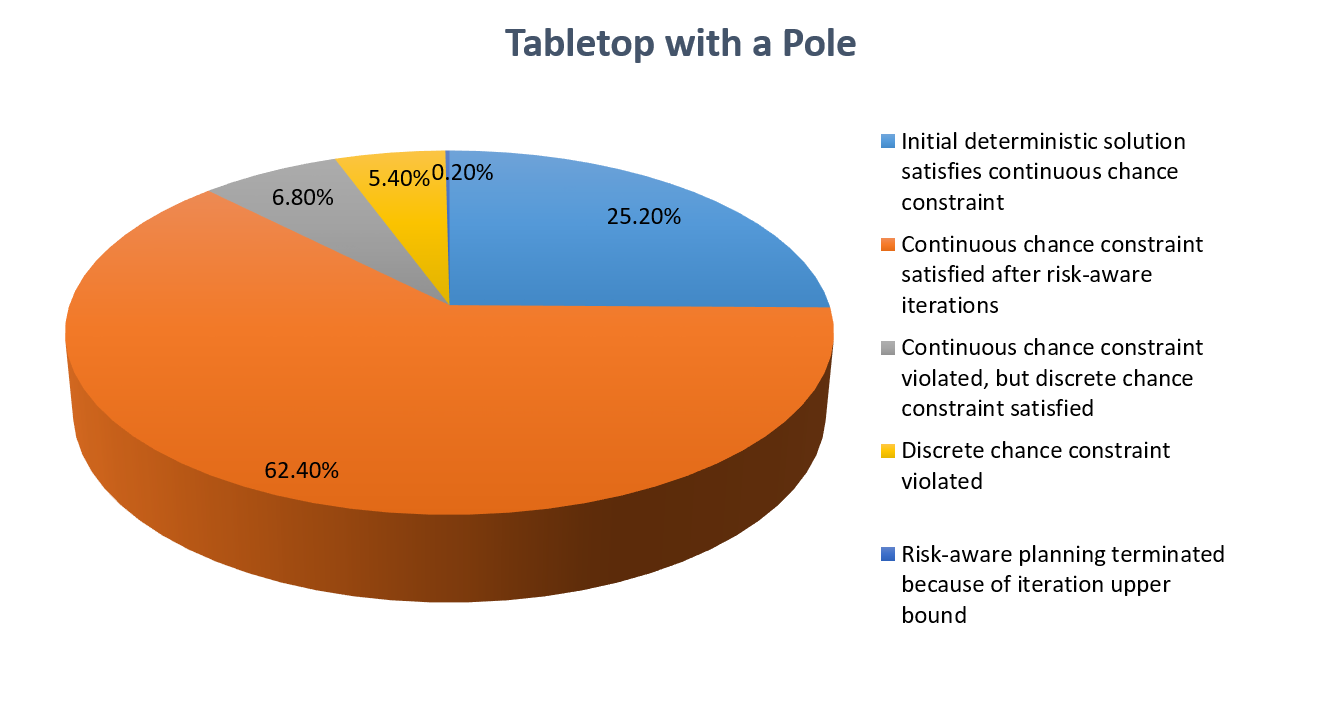}
      \end{subfigure}
      \begin{subfigure}[t]{0.5\textwidth}
      \includegraphics[width=1.08\linewidth]{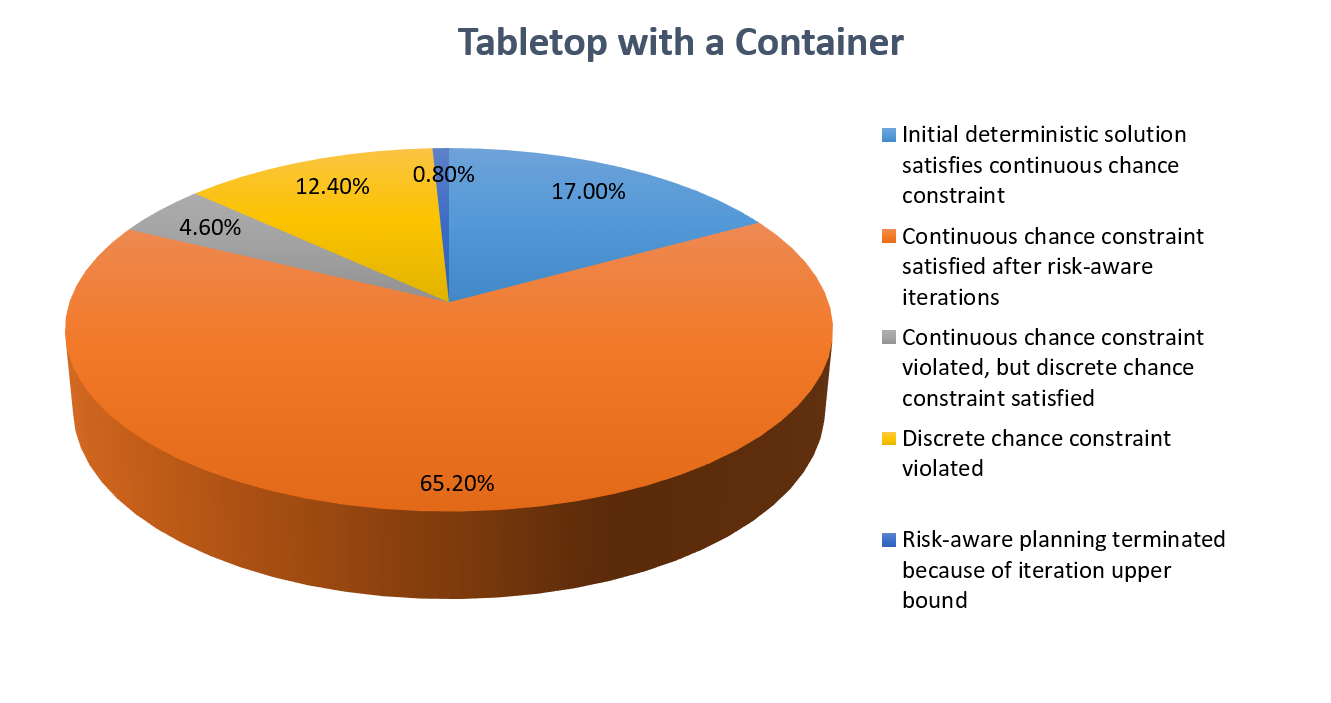}
      \end{subfigure}
      \caption{Quadrature-based p-Chekov statistics breakdown for feasible cases with end-effector observation, 0.0044 noise standard deviation and 10\% chance constraint}
      \label{fig:feasible_1}
   \end{figure}

\begin{table}
\caption{Results in Potentially Feasible Test Cases with Joint Value Observation, Noise Level 0.0044 and Chance Constraint 5\%}
\label{table:feasible_joint_obs}
\footnotesize\sf\centering
\begin{threeparttable}
\begin{tabular}{L{1.57cm}|L{1.2cm}|L{2.94cm}|L{0.84cm} L{0.84cm}}
\toprule
 \multicolumn{3}{c|}{Environment}  &  Tabletop with a Pole  & Tabletop with a Container   \\ 

\hline
\multirow{2}{1.63cm}{\centering Planning Time (s)} 
& \multicolumn{2}{c|}{deterministic Chekov} & 1.28  & 1.29 \\ 
& \multicolumn{2}{c|}{p-Chekov} & 5.40 & 10.44 \\ 

\hline
\multirow{2}{1.65cm}{\centering Overall Collision Rate} 
& \multicolumn{2}{c|}{deterministic Chekov} & 28.98\%  & 43.49\%  \\ 
& \multicolumn{2}{c|}{p-Chekov} & 1.60\% & 6.13\%   \\ 

\hline
\multirow{2}{1.63cm}{\centering Average Path Length (rad)} 
& \multicolumn{2}{c|}{deterministic Chekov} & 0.51  & 0.59  \\ 
& \multicolumn{2}{c|}{p-Chekov} & 0.52  & 0.64  \\

\hline
\multirow{12}{1.63cm}{\centering P-Chekov Performance} 
& \multicolumn{2}{m{4.4cm}|}{continuous chance constraint satisfaction rate} & 98.40\% & 93.80\%  \\ \cline{2-5}
& \multirow{3}{1.3cm}{continuous satisfied cases} & average iteration number & 1.15 & 2.02 \\ 
&  & average collision rate & 0.00\%  & 0.01\%  \\ 
&  & average risk reduction & 0.29 & 0.41 \\ \cline{2-5}
& \multirow{3}{1.3cm}{continuous violated cases} & average iteration number & 4.00  & 8.19  \\
&  & average collision rate & 100.00\%  & 98.68\%  \\ 
&  & average risk reduction & -0.43  & -0.16  \\ \cline{2-5}

& \multicolumn{2}{m{4.4cm}|}{discrete chance constraint satisfaction rate} & 99.00\% & 96.20\%  \\ \cline{2-5}
& \multirow{3}{1.3cm}{discrete satisfied cases} & average iteration number & 1.18 & 2.25 \\ 
&  & average collision rate & 0.00\%  & 0.01\%  \\ 
&  & average risk reduction & 0.19 & 0.31 \\ \cline{2-5}
& \multirow{3}{1.3cm}{discrete violated cases} & average iteration number & 3.20  & 6.37  \\
&  & average collision rate & 100.00\%  & 98.47\%  \\ 
&  & average risk reduction & -0.62  & -0.15  \\ 
\bottomrule
\end{tabular}
 \end{threeparttable}
\end{table}

\begin{table}
\caption{Results in Potentially Feasible Test Cases with End-effector Observation, Noise Level 0.0044 and Chance Constraint 10\%}
\label{table:feasible_ee_obs}
\footnotesize\sf\centering
\begin{threeparttable}
\begin{tabular}{L{1.57cm}|L{1.2cm}|L{2.94cm}|L{0.84cm} L{0.84cm}}
\toprule
 \multicolumn{3}{c|}{Environment}  &  Tabletop with a Pole  & Tabletop with a Container   \\ 

\hline
\multirow{2}{1.63cm}{\centering Planning Time (s)} 
& \multicolumn{2}{c|}{deterministic Chekov} & 1.10  & 1.27 \\ 
& \multicolumn{2}{c|}{p-Chekov} & 19.34 & 31.17 \\ 

\hline
\multirow{2}{1.65cm}{\centering Overall Collision Rate} 
& \multicolumn{2}{c|}{deterministic Chekov} & 27.51\%  & 41.04\%  \\ 
& \multicolumn{2}{c|}{p-Chekov} & 11.39\% & 16.46\%   \\ 

\hline
\multirow{2}{1.63cm}{\centering Average Path Length (rad)} 
& \multicolumn{2}{c|}{deterministic Chekov} & 0.51  & 0.60 \\ 
& \multicolumn{2}{c|}{p-Chekov} & 0.68  & 0.84  \\

\hline
\multirow{12}{1.63cm}{\centering P-Chekov Performance} 
& \multicolumn{2}{m{4.4cm}|}{continuous chance constraint satisfaction rate} & 87.60\% & 82.20\%  \\ \cline{2-5}
& \multirow{3}{1.3cm}{continuous satisfied cases} & average iteration number & 4.14 & 5.19 \\ 
&  & average collision rate & 0.08\%  & 0.11\%  \\ 
&  & average risk reduction & 0.25 & 0.33 \\ \cline{2-5}
& \multirow{3}{1.3cm}{continuous violated cases} & average iteration number & 10.52  & 10.35  \\
&  & average collision rate & 88.50\%  & 88.02\%  \\ 
&  & average risk reduction & -0.44  & -0.13  \\ \cline{2-5}

& \multicolumn{2}{m{4.4cm}|}{discrete chance constraint satisfaction rate} & 94.40\% & 86.80\%  \\ \cline{2-5}
& \multirow{3}{1.3cm}{discrete satisfied cases} & average iteration number & 4.82 & 5.49 \\ 
&  & average collision rate & 0.13\%  & 0.10\%  \\ 
&  & average risk reduction & 0.19 & 0.28 \\ \cline{2-5}
& \multirow{3}{1.3cm}{discrete violated cases} & average iteration number & 6.94  & 10.32  \\
&  & average collision rate & 73.39\%  & 86.59\%  \\ 
&  & average risk reduction & -0.39  & -0.23  \\ 
\bottomrule
\end{tabular}
 \end{threeparttable}
\end{table}

\begin{table*}
\caption{Improvement from Iterative Risk Allocation for Experiments with Both Observation Models\tnote{1}}
\label{table:IRA}
\small\sf\centering
\begin{threeparttable}
\begin{tabular}{L{3.8cm}|L{1.6cm}|L{1.05cm} L{1.2cm}|L{1.2cm} L{1.2cm}|L{1.2cm} L{1.2cm}|L{1.2cm} L{1.2cm}}
\toprule
\multicolumn{2}{c|}{Environment} & \multicolumn{4}{c|}{Tabletop with a Pole}  & \multicolumn{4}{c}{Tabletop with a Container} \\
\hline
\multicolumn{2}{c|}{Test Case Filtering}  &  \multicolumn{2}{c|}{All Cases}  & \multicolumn{2}{c|}{Feasible Cases} &   \multicolumn{2}{c|}{All Cases}  & \multicolumn{2}{c}{Feasible Cases}\\ 
\hline
\multicolumn{2}{c|}{Observation Model\tnote{2}} & Joint & End-Effector & Joint & End-Effector & Joint & End-Effector & Joint & End-Effector \\
\hline
\multirow{2}{3.7cm}{\centering Continuous Chance Constraint Satisfaction Rate\tnote{3}} 
& Without IRA & 85.62\%  & 69.92\% & 96.58\% & 82.16\% & 53.57\% & 41.72\% & 90.94\% & 76.80\% \\ 
& With IRA & 88.44\% & 69.92\% & 97.72\% & 84.21\% & 55.00\% & 43.08\% & 93.44\% & 77.87\% \\ 

\hline
\multirow{2}{3.7cm}{\centering Discrete Chance Constraint Satisfaction Rate\tnote{4}} 
& Without IRA & 88.44\%  & 81.49\% & 97.83\%  & 90.35\% & 64.29\% & 55.33\% & 94.06\% & 82.67\% \\ 
& With IRA & 90.00\% & 82.26\% & 98.86\%   & 91.23\% & 63.81\% & 54.20\% & 95.94\% & 83.47\%  \\ 

\hline
\multirow{2}{3.7cm}{\centering Average Trajectory Length (rad)\tnote{5}} 
& Without IRA & 0.62 & 0.85  & 0.64 & 0.81 & 0.81 & 1.10 & 0.74 & 0.94 \\ 
& With IRA & 0.60 & 0.80 & 0.63  & 0.77 & 0.78 & 1.03 & 0.72 & 0.88 \\

\bottomrule
\end{tabular}
\begin{tablenotes}
\footnotesize
 \item[1] The results shown are averaged from the test cases where the IRA number of iteration is non-zero only. The test cases where all constraints are already active in the planning phase solution are not included.
 \item[2] The chance constraint for experiments with joint configuration observation model is set to 5\%, and the chance constraint for experiments with end-effector pose observation model is set to 10\%.
 \item[3] Percentage of the test cases where the average continuous-time collision rate over 100 noisy executions satisfies the chance constraint.
 \item[4] Percentage of the test cases where the average waypoint collision rate over 100 noisy executions satisfies the chance constraint.
 \item[5] Average length of actual execution trajectories instead of nominal solution trajectories.
 \end{tablenotes}
 \end{threeparttable}
\end{table*}

As previously mentioned, a lot of test queries where p-Chekov fails have their start or goal very close to obstacles. In these cases, feasible solutions might not exist if the collision probability of the start or goal has already exceeded the chance constraint. Therefore, we introduce a pre-processing procedure before running p-Chekov in order to filter out these potentially infeasible test queries. We estimate the collision probability of the start and goal based on the nominal trajectory computed by deterministic Chekov, and discard the test cases where the collision probability of either the start or goal exceeds 1.5 times of the chance constraint. Although it is possible that some of these cases might be feasible since our collision probability estimation approach is conservative, most of them are highly likely to be infeasible compared to other cases where the start and goal has low estimated collision probabilities. We pick 500 test cases that have passed this pre-processing and call them ``feasible cases'' in short, to distinguish from the unfiltered 500 test cases used in previous experiments in this section. Figure~\ref{fig:feasible_1} shows the statistics breakdown for the experiments with the end-effector observation model in the two tabletop environments after filtering out the potentially infeasible test cases. The chance constraint is set to 10\% and the noise standard deviation is 0.0044 rad. Compared with Figure~\ref{fig:breakdown_1}, it is obvious that the chance constraint satisfaction rate has significantly increased. In ``tabletop with a pole'' environment we can see that in 25.20\% of the test cases, the initial deterministic solution has already satisfied the chance constraint, and there are 62.40\% test cases where the chance constraint is satisfied after p-Chekov risk-aware iterations. Only 6.80\% of the test cases fail because of edge collisions, and 5.40\% fail for other reasons. Similarly, the ``tabletop with a container'' environment also shows that 65.20\% of test queries can satisfy the chance constraint after p-Chekov risk-aware iterations, in contrast to the 30.52\% shown in Figure~\ref{fig:breakdown_1}. In this environment, the difference between the experiment results before and after pre-processing is much more noticeable than in the ``tabletop with a pole'' environment, indicating that more cases are infeasible in this complicated environment with narrow spaces. 

Table~\ref{table:feasible_joint_obs} and Table~\ref{table:feasible_ee_obs} compare p-Chekov's performance with joint configuration observation model and with end-effector pose observation model in feasible cases for both environments. Compared with Table~\ref{table:00044noise} to \ref{table:ee_01}, it is noticeable that the results are significantly improved after filtering out potentially infeasible cases. In Table~\ref{table:feasible_joint_obs} we can see that with joint configuration observations, p-Chekov can achieve a chance constraint satisfaction rate of above 90\% in both environments. Especially in the ``tabletop with a container'' environment, p-Chekov shows powerful collision risk reduction ability by having an average risk reduction of 0.41 in satisfied cases with only a small increase in the average execution trajectory length. With end-effector observations, as shown in Table~\ref{table:feasible_ee_obs}, although the performance is not as good compared to Table~\ref{table:feasible_joint_obs}, the constraint satisfaction rates in both environments are still above 80\%. The overall collision risks are significantly reduced compared to deterministic Chekov's solutions, and the average risk reductions in constraint satisfied cases are also very high in both environments. However, in these constraint satisfied cases in Table~\ref{table:feasible_ee_obs}, the collision risk is much lower than the chance constraint level, and the average execution trajectory lengths are much longer compared with deterministic Chekov's solutions. This indicates that p-Chekov's planning phase algorithm can return overly conservative solutions, thus in Section~\ref{execution_results}, we will use the IRA algorithm in p-Chekov's execution phase to improve solution quality.

\subsubsection{Improvement from iterative risk allocation}  \label{execution_results}

This section presents the improvement from using the IRA algorithm introduced in Section \ref{execution_phase_IRA}. Table~\ref{table:IRA} compare the solutions from p-Chekov's planning phase algorithm and from using an IRA procedure after the planning phase in both environments. The comparison considers three main aspects: the percentage of test cases where the continuous-time collision rate satisfies the chance constraint, the percentage of test cases where the waypoint collision rate satisfies the chance constraint, and the average trajectory length over 100 noisy executions. The columns of ``All Cases'' refer to the results of the original 500 test cases, and the columns of ``Feasible Cases'' refer to the results of the 500 test cases after filtering out potentially infeasible cases based on the collision probability of start and end poses. Table~\ref{table:IRA} shows that IRA can slightly improve the chance constraint satisfaction rate, for both continuous-time and discrete-time satisfactions, especially in the relatively difficult ``tabletop with a container'' environment. The improvement on average trajectory length is the main effect of IRA. From Table~\ref{table:IRA} we can see that the solutions with IRA are much shorter, indicating that the trajectory quality is improved without sacrificing the chance constraint satisfaction rate. These results prove that using IRA during execution phase can effectively redistribute risk bounds among different waypoints and improve the solution quality by providing less conservative trajectories that also satisfy the chance constraint. 

\subsection{Learning-based p-Chekov experiment results}   \label{learning_results}

The results from Section~\ref{risk_aware_results} indicate that the planning time of quadrature-based p-Chekov will severely constrain its application in real-time planning tasks that require fast-reaction. In this section, we first compare the training performance of four different classes of machine learning methods in the same two tabletop environments, and then demonstrate the performance of neural network-based p-Chekov, the best performer among the four, with 500 feasible test cases in each environment.

\subsubsection{Comparison between different regression methods}  \label{section:learning}

\begin{table*}
\caption{Best Parameters in Kernel Ridge Regression and Random Forest Regression}
\label{best_param}
\small\sf\centering
\begin{tabular}{L{5.2cm}|L{3.1cm}|L{3.1cm}|L{3.1cm}}
\toprule
Regressor & 8000 training data & 18000 training data & 38000 training data \\ \hline
RBF Kernel Ridge Regression & $\alpha = 0.2$, $\gamma = 0.3$ & $\alpha = 0.2$, $\gamma = 0.3$ & $\alpha = 0.2$, $\gamma = 0.5$ \\ \hline
Polynomial Kernel Ridge Regression & $\alpha = 0.1$, $degree = 5$ & $\alpha = 0.1$, $degree = 6$ & $\alpha = 0.1$, $degree = 6$ \\ \hline
Matern Kernel Ridge Regression & $\alpha = 0.1$, $\nu = 2.29$, $length~scale = 1.5$  & $\alpha = 0.01$, $\nu = 1.14$, $length~scale = 2.5$ & $\alpha = 0.01$, $\nu = 1.44$, $length~scale = 1.66$ \\  \hline
Random Forest Regression & $estimators = 300$, $min~split = 5$, $min~leaf = 3$ & $estimators = 600$, $min~split = 5$, $min~leaf = 3$  & $estimators = 600$, $min~split = 4$, $min~leaf = 3$\\ \bottomrule
\end{tabular}
\end{table*}

In p-Chekov, since the nominal trajectories generated by the deterministic planner are guaranteed to be collision-free without the presence of noise, the nominal configurations inputted into the collision estimator component is more likely to lie in the collision-free configuration space. Therefore, the 60000 samples in each environment include two parts: 20000 have their mean configurations sampled from the entire configuration space (referred to as \emph{Sample Set~1}), and 40000 have their mean configurations sampled purely from the deterministic collision-free configuration space (referred to as \emph{Sample Set~2}). 2000 samples are held out for testing in every experiment no matter how large the training size is. Note that even though the mean configuration is not in collision, the associated collision risk is not necessarily zero if the standard deviation is nonzero. Having more samples taken from the deterministic collision-free space can better represent the practical data p-Chekov faces during online planning. All the 60000 samples from both environments are used when training neural networks, while we only use Sample Set 1 and half of Sample Set 2 to train Scikit Learn regressors because they get very slow when the data size exceeds 40000.

In order to find the best parameters for the regressors, we conduct grid search on kernel ridge regressors and random forest regressors, and use gradient descent on Gaussian process regressors. Table~\ref{best_param} shows the best parameters found for different kernels in kernel ridge regression as well as random forest regression when the training data have different sizes. Since gradient descent for Gaussian process is applied during training to maximize the log marginal likelihood, the best parameters are not shown in Table~\ref{best_param}. All the experiments in this section use the best parameters we found for the corresponding data size.

The comparison between different regression methods on different datasets is shown in Table~\ref{table:reg_results}. In each dataset, 2000 randomly selected data points are used for testing and the rest are used for training. 
Mean squared error (MSE) and $R^2$ score are used to measure the test accuracy for different regressors. Given the predicted value $\hat{y}_i$ and the true value $y_i$ for each test data point, MSE is calculated by:

\begin{equation}
 \text{MSE}(y, \hat{y}) = \frac{1}{n_{samples}} \sum_{i=1}^{n_{samples}} (y_i - \hat{y}_i)^2,
\end{equation}

\noindent and the $R^2$ score is calculated by:

\begin{equation}
 R^2(y, \hat{y}) = 1 - \frac{\sum_{i=1}^{n_{samples}} (y_i - \hat{y}_i)^2}{\sum_{i=1}^{n_{samples}} (y_i - \bar{y}_i)^2},
\end{equation}

\noindent where $\bar{y} = \frac{1}{n_{samples}} \sum_{i=1}^{n_{samples}} y_i$. Here the $R^2$ scores are computed using the test data, and the MSE scores and standard deviations are computed using cross validation on the training data. In terms of the training time performance, Gaussian process regression and Matern kernel ridge regression are the slowest. Gaussian process regression conducts gradient descent to search for best parameters during training, and also outputs distributions instead of single predicted values, which would explain its low training speed. As for Matern kernel ridge regression, the best $\nu$ parameter found by grid search is not one of the default values provided by Scikit Learn, and this would incur a considerably higher computational cost (approximately 10 times higher) since they require to evaluate the modified Bessel function. In contrast, random forest regressor tends to take a very short time to train, and its training time also grows relatively slowly as the size of training data increases. In terms of prediction accuracy, Matern kernel ridge regression and random forest regression have the best performance when the training data include Sample Set 1 data, while polynomial kernel shows the worst performance. RBF kernel ridge regression shows slightly better performance compared to random forest regression when the training data are purely from Sample Set 2. When provided with 38000 training data, the MSE error of random forest regressor is relatively satisfactory, and a number of manually selected test points showed that the prediction is very close to the ``ground truth'' risk value.

\begin{table}
\caption{Comparison of Different Regression Methods}
\label{table:reg_results}
\footnotesize\sf\centering
\begin{tabular}{L{0.6cm}|L{0.72cm}|L{0.74cm}|L{1.36cm} L{0.72cm} L{0.85cm} L{0.82cm} L{0.98cm}}
\toprule
Data & Size & \multicolumn{2}{m{2.18cm}}{\centering Regression Method} & $R^2$ Score & MSE Error & Std of MSE & Training Time (s) \\ \hline
\multirow{10}{0.65cm}{\centering Set 1} & \multirow{5}{0.65cm}{\centering 10000} & \multirow{3}{0.7cm}{\centering Kernel Ridge} & RBF  & 0.792 & 0.0217 & 0.0006 & 5.30 \\ 
& & & Polynomial  & 0.731 & 0.0314 & 0.0010 & 7.37 \\ 
& & & Matern  & 0.801 & 0.0210 & 0.0006 & 40.83 \\ \cline{3-8}
& & \multicolumn{2}{m{2.18cm}}{\centering Random Forest} & 0.799 & 0.0217 & 0.0007 & 1.76 \\ \cline{3-8}
& & \multicolumn{2}{m{2.18cm}}{\centering Gaussian Process} & 0.791 & 0.0221 & 0.0006 & 140.15 \\ \cline{2-8}
& \multirow{5}{0.65cm}{\centering 20000} & \multirow{3}{0.7cm}{\centering Kernel Ridge} & RBF  & 0.842 & 0.0180 & 0.0005 & 31.78 \\ 
& & & Polynomial  & 0.783 & 0.0261 & 0.0007 & 58.21 \\ 
& & & Matern  & 0.851 & 0.0172 & 0.0005 & 311.11 \\ \cline{3-8}
& & \multicolumn{2}{m{2.18cm}}{\centering Random Forest} & 0.854 & 0.0166 & 0.0006 & 4.80 \\ \cline{3-8}
& & \multicolumn{2}{m{2.18cm}}{\centering Gaussian Process} & 0.844 & 0.0181 & 0.0005 & 1108.75 \\ \hline
\multirow{10}{0.65cm}{\centering Set 2} & \multirow{5}{0.65cm}{\centering 10000} & \multirow{3}{0.7cm}{\centering Kernel Ridge} & RBF  & 0.616 & 0.0077 & 0.0003 & 5.29 \\ 
& & & Polynomial  & 0.551 & 0.0097 & 0.0004 & 8.29 \\ 
& & & Matern  & 0.629 & 0.0075 & 0.0003 & 79.25 \\ \cline{3-8}
& & \multicolumn{2}{m{2.18cm}}{\centering Random Forest} & 0.593 & 0.0083 & 0.0003 & 5.48 \\ \cline{3-8}
& & \multicolumn{2}{m{2.18cm}}{\centering Gaussian Process} & 0.626 & 0.0077 & 0.0003 & 166.39  \\ \cline{2-8}
& \multirow{5}{0.65cm}{\centering 20000} & \multirow{3}{0.7cm}{\centering Kernel Ridge} & RBF  & 0.682 & 0.0066 & 0.0002 & 33.27 \\ 
& & & Polynomial  & 0.603 & 0.0085 & 0.0002 & 42.73 \\ 
& & & Matern  & 0.697 & 0.0063 & 0.0002 & 213.28 \\ \cline{3-8}
& & \multicolumn{2}{m{2.18cm}}{\centering Random Forest} & 0.661 & 0.0071 & 0.0002 & 4.23 \\ \cline{3-8}
& & \multicolumn{2}{m{2.18cm}}{\centering Gaussian Process} & 0.689 & 0.0066 & 0.0002 & 1337.29  \\ \hline
\multirow{5}{0.65cm}{\centering Both Sets} & \multirow{5}{0.65cm}{\centering 40000} & \multirow{3}{0.7cm}{\centering Kernel Ridge} & RBF  & 0.847 & 0.0117 & 0.0002 & 435.60  \\ 
& & & Polynomial  & 0.773 & 0.0165 & 0.0002 & 261.03 \\ 
& & & Matern  & 0.855 & 0.0110 & 0.0002 & 1619.93 \\ \cline{3-8}
& & \multicolumn{2}{m{2.18cm}}{\centering Random Forest} & 0.868 & 0.0107 & 0.0002 & 18.08 \\ \cline{3-8}
& & \multicolumn{2}{m{2.18cm}}{\centering Gaussian Process} & 0.849 & 0.0113 & 0.0002 & 8833.86 \\ 
\bottomrule
\end{tabular}
\end{table}

If we compare the results between training on Sample Set 1 and training on Sample Set 2, we can see that Sample Set 2 tests show a smaller MSE error but a lower $R^2$ score. This is because the Sample Set 2 data points are all sampled from the collision-free configuration space, which would tend to have lower collision risk than the in-collision configurations. Therefore, it is reasonable that a lower absolute value leads to a lower MSE error. However, since $R^2$ scores measures the relative error compared to the variance of the original data, it won't decrease as the absolute values of data points decrease. One hypothesis about why the $R^2$ score is lower compared to Sample Set 1 is that Sample Set 2 tends to have ``ground-truth'' collision risk close to 0, and there's not enough variety on the data distribution to ensure that the regressor can capture the data structure. This conclusion shows that although in practical motion planning tasks the configurations that p-Chekov needs to predict collision risk for are more likely to be in the collision-free configuration space, having data from the entire configuration space helps the regressor to learn the data distribution better and achieve higher prediction accuracy.

   \begin{figure}
      \centering
      \begin{subfigure}[t]{0.5\textwidth}
      \includegraphics[width=\linewidth]{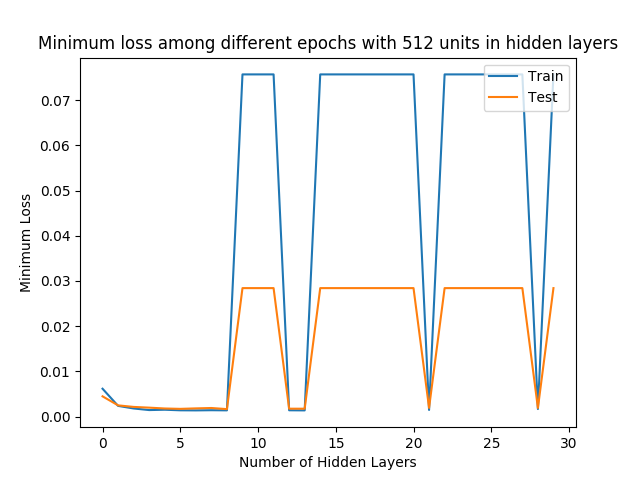}
      \end{subfigure}
      \begin{subfigure}[t]{0.5\textwidth}
      \includegraphics[width=\linewidth]{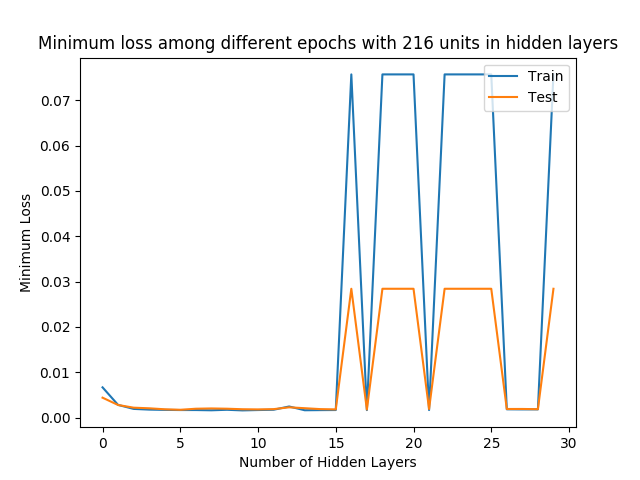}
      \end{subfigure}
      \caption{Minimum loss as a function of hidden layer numbers}
      \label{fig:min_loss}
   \end{figure}  

The neural networks used in this paper are fully connected networks with ReLU activation for the input layer and hidden layers. Adam~\citep{kingma2014adam} optimizer is used, and the batch size is set to 64. Sigmoid is used as the output layer activation function since the output is collision probability. MSE is used as the loss function because this regression problem aims at minimizing the prediction error. All the 60000 data points are used in the neural network experiments: 58000 are for training and 2000 are for testing. In all the neural networks tested in this paper, the number of units in the input layer is kept as 1024, and all the output layer have 1 unit to match the sigmoid output. We compare the networks' performance with different numbers of hidden layers in Figure~\ref{fig:min_loss}, where the top figure has 512 units in each hidden layer and the bottom one has 216. The vertical axis in these two figures shows the minimum training and validation losses within 50 training epochs. From Figure~\ref{fig:min_loss} we can see that when the number of hidden layers with 512 units lies between 0 and 9, the neural networks have relatively stable performance, and the minimum loss has a decreasing trend as the number of layers increases. However, when the number of hidden layer reaches 10, the neural networks start to have trouble minimizing the MSE loss. One of the potential reasons for this phenomenon is that when the neural networks get very deep, the input to the last activation layer, the sigmoid layer, might get very large. Since sigmoid function has very small gradient when the input is large, this could potentially cause the optimizer not being able to properly conduct gradient descent, which then causes high training losses and validation losses. When there are 216 units in hidden layers, the minimum loss curves show similar trends, but the networks have a wider range of hidden layer numbers where the optimization is stable since the layers are narrower. This is potentially related to the fact that when the width of each layer is smaller, it takes the networks more layers to reach saturation where the gradient of activation function approaches zero. The minimum validation loss in the bottom figure of Figure~\ref{fig:min_loss} is 0.0017, when the number of hidden layers is 6, and in the top figure the minimum reaches 0.0016, when the number of hidden layers is 9. Therefore, the optimal network structure among all tested ones has 9 hidden layers with 512 units each, an input layer with 1024 units and an output layer with one sigmoid activation unit.

\begin{table}
\small\sf\centering
\caption{Performance of Neural Network with 9 Hidden Layers with 512 Units Each}
\label{Opt}
\begin{tabular}{L{1.4cm}|L{1.4cm} L{1.4cm} L{1.4cm} L{1.5cm}}
\toprule
Number of Training Epochs & Final Training Loss & Final Validation Loss & Minimum Validation Loss  & Number of Epoch for Minimum Validation Loss \\ \hline
50 & 0.001371 & 0.001698 & 0.001645 & 49 \\ \hline
70 & 0.001031 & 0.001519 & 0.001437 & 63 \\ \hline
100 & 0.000426  & 0.001400 & 0.001302 & 99 \\  \bottomrule
\end{tabular}
\end{table}

\begin{figure}
  \centering
  \includegraphics[width=\linewidth]{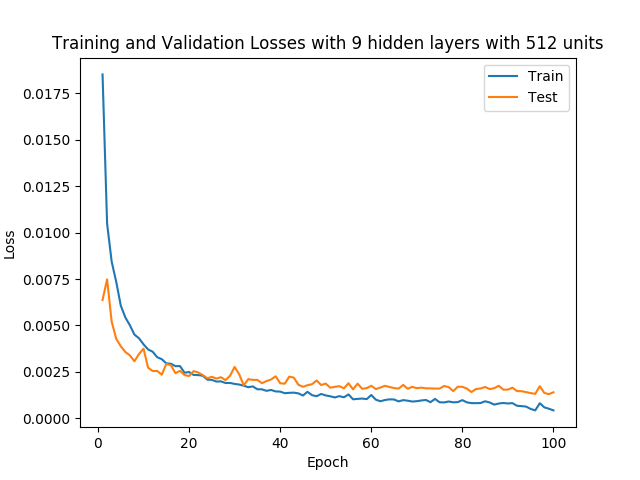}
  \caption{Loss and training epoch relationship for networks with 9 hidden layers with 512 units each}
  \label{fig:epoch_100}
\end{figure}

Table~\ref{Opt} and Figure~\ref{fig:epoch_100} show the performance of the optimal structure network, 9 hidden layers with 512 units each, when more training epochs are provided. Table~\ref{Opt} compares their performance in terms of the training and validation loss after the last epoch's training (the ``Final Training Loss'' and ``Final Validation Loss'' columns), the minimum validation loss among all epochs (the ``Minimum Validation Loss'' column), and the number of epoch where they reach this minimum (the ``Number of Epoch for Minimum Validation Loss'' column). As we can see from Table~\ref{Opt}, both the training loss and the validation loss are decreasing given more training epochs, but the improvement for validation loss is much smaller compared to that of training loss. This means that as we exploit the training data more, although the performance of neural networks will gradually improve, this improvement is more about better fitting the training data structure than generalizing to the entire C-space. Therefore, we would expect very limited improvement or even decreasing validation performance when training more than 100 epochs. Figure~\ref{fig:epoch_100} also shows that the validation loss decreases drastically in the first 30 epochs and then starts to drop slowly, whereas the training loss is still decreasing relatively fast and diverges from the validation loss in the final 30 epochs.

\subsubsection{Neural network learning-based p-Chekov experiment results}  \label{section:NN_results}

   \begin{figure}
      \centering
      \begin{subfigure}[t]{0.5\textwidth}
      \includegraphics[width=1.02\linewidth]{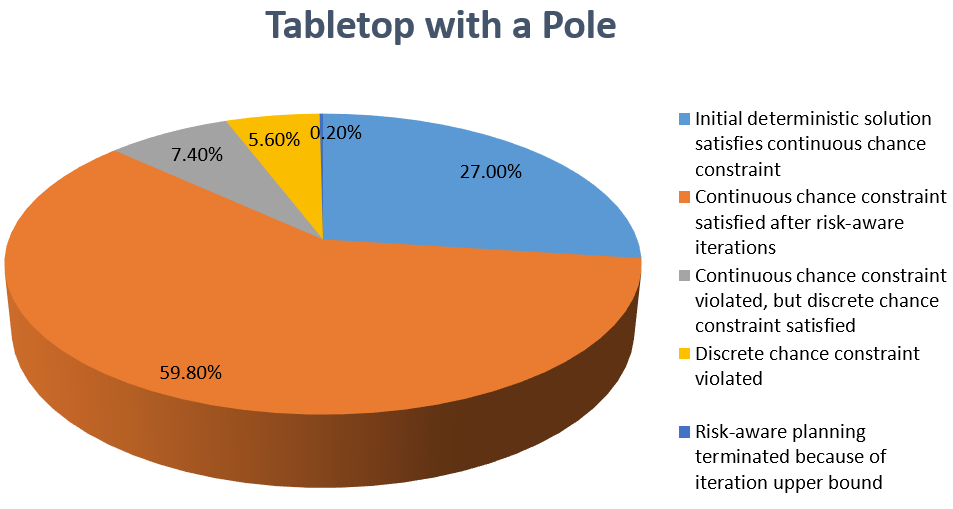}
      \end{subfigure}
      \begin{subfigure}[t]{0.5\textwidth}
      \includegraphics[width=1.02\linewidth]{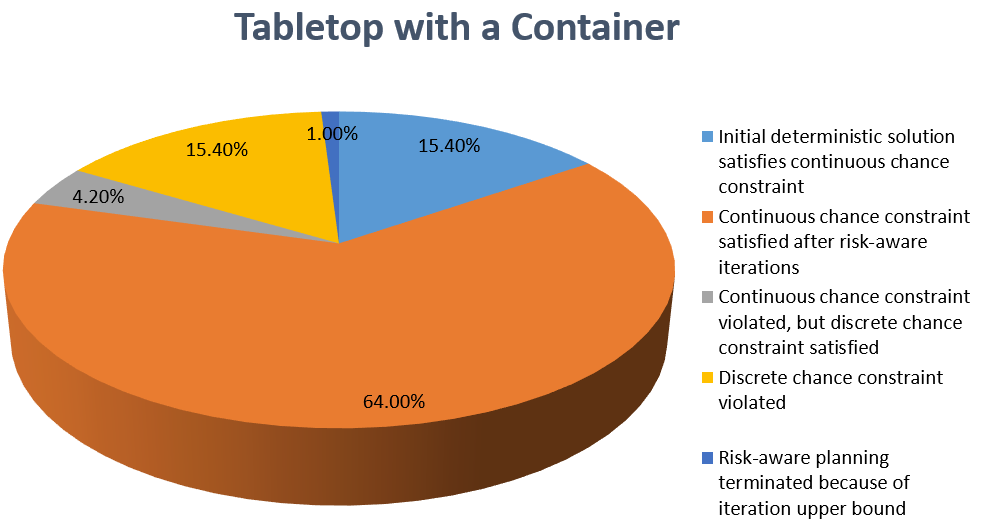}
      \end{subfigure}
      \caption{Learning-based p-Chekov statistics breakdown for feasible cases with end-effector observation, 0.0044 noise standard deviation and 10\% chance constraint}
      \label{fig:NN_feasible_1}
   \end{figure}  

\begin{table}
\caption{Results in Potentially Feasible Test Cases with End-effector Observation, Noise Level 0.0044 and Chance Constraint 10\%}
\label{table:feasible_ee_NN}
\footnotesize\sf\centering
\begin{threeparttable}
\begin{tabular}{L{1.55cm}|L{1.2cm}|L{3cm}|L{0.84cm} L{0.84cm}}
\toprule
 \multicolumn{3}{c|}{Environment}  &  Tabletop with a Pole  & Tabletop with a Container   \\ 

\hline
\multirow{2}{1.63cm}{\centering Planning Time (s)} 
& \multicolumn{2}{c|}{deterministic Chekov} & 1.12  & 1.22 \\ 
& \multicolumn{2}{c|}{p-Chekov} & 8.65 & 10.15 \\ 

\hline
\multirow{2}{1.65cm}{\centering Overall Collision Rate} 
& \multicolumn{2}{c|}{deterministic Chekov} & 31.05\%  & 42.54\%  \\ 
& \multicolumn{2}{c|}{p-Chekov} & 11.82\% & 18.53\%   \\ 

\hline
\multirow{2}{1.63cm}{\centering Average Path Length (rad)} 
& \multicolumn{2}{c|}{deterministic Chekov} & 0.51  & 0.61 \\ 
& \multicolumn{2}{c|}{p-Chekov} & 0.71  & 0.85  \\

\hline
\multirow{12}{1.63cm}{\centering P-Chekov Performance} 
& \multicolumn{2}{m{4.4cm}|}{continuous chance constraint satisfaction rate} & 86.80\% & 79.40\%  \\ \cline{2-5}
& \multirow{4}{1.3cm}{continuous satisfied cases} & average iteration number & 3.49 & 4.41 \\ 
&  & average planning time (s) & 6.39  & 7.96  \\ 
&  & average collision rate & 0.11\%  & 0.12\%  \\ 
&  & average risk reduction & 0.27 & 0.33 \\ \cline{2-5}
& \multirow{4}{1.3cm}{continuous violated cases} & average iteration number & 10.66  & 8.82  \\
&  & average planning time (s) & 22.96  & 18.10  \\ 
&  & average collision rate & 86.19\%  & 85.35\%  \\ 
&  & average risk reduction & -0.28  & -0.09  \\ \cline{2-5}

& \multicolumn{2}{m{4.4cm}|}{discrete chance constraint satisfaction rate} & 94.20\% & 83.60\%  \\ \cline{2-5}
& \multirow{4}{1.3cm}{discrete satisfied cases} & average iteration number & 4.13 & 4.59 \\ 
&  & average planning time (s) & 8.29  & 8.58  \\ 
&  & average collision rate & 0.10\%  & 0.16\%  \\ 
&  & average risk reduction & 0.22 & 0.29 \\ \cline{2-5}
& \multirow{4}{1.3cm}{discrete violated cases} & average iteration number & 9.03  & 9.06  \\
&  & average planning time (s) & 13.54  & 17.69  \\ 
&  & average collision rate & 65.82\%  & 79.65\%  \\ 
&  & average risk reduction & -0.19  & -0.11  \\ 
\hline
\multirow{8}{1.63cm}{\centering Improvement from IRA (for non-zero number of IRA iteration cases)}
& \multirow{4}{1.3cm}{Without IRA} & continuous satisfaction rate & 79.94\% & 72.31\% \\
& & discrete satisfaction rate & 91.22\% & 78.23\% \\
& & average path length & 0.88 & 0.96 \\ \cline{2-5}
& \multirow{4}{1.3cm}{With IRA} & continuous satisfaction rate & 81.82\% & 71.24\% \\
& & discrete satisfaction rate & 93.42\% & 77.42\% \\
& & average path length & 0.86 & 0.97 \\
\bottomrule
\end{tabular}
 \end{threeparttable}
\end{table}

Section~\ref{section:learning} shows that the best performer among all the tested machine learning methods on this collision risk regression problem is the neural network with 9 hidden layers with 512 units in each layer, thus it is used in this section to evaluate the performance of learning-based p-Chekov. Figure~\ref{fig:NN_feasible_1} demonstrates the statistics breakdown of the neural network learning-based p-Chekov experiments with 500 feasible test cases, end-effector pose observation model, noise standard deviation 0.0044 and chance constraint 10\%, and Table~\ref{table:feasible_ee_NN} shows more detailed performance. Note that the pre-processing here is conducted with the learning-based collision estimator, which might select slightly different test cases compared to using the quadrature-based collision estimator because their estimated risk for the same start and goal pose pair might not be exactly the same. The comparison between Figure~\ref{fig:NN_feasible_1} and Figure~\ref{fig:feasible_1} shows that the experiment results from quadrature-based p-Chekov and learning-based p-Chekov have similar structures.  Since the learning-based collision estimation component is less conservative than the quadrature-based one, it filters out fewer difficult test cases, which could explain the relatively lower chance constraint satisfaction rate in Figure~\ref{fig:NN_feasible_1}. Comparing Table~\ref{table:feasible_ee_NN} with Table~\ref{table:feasible_ee_obs}, we can see that although the collision rate performance and the path length performance of the two algorithms are similar, learning-based p-Chekov's planning time is significantly shorter, especially in the more difficult ``tabletop with a container'' environment where it reduced the average planning time by 67\%. From the ``improvement from IRA'' section in table~\ref{table:feasible_ee_NN} we can see that IRA can slightly improve the planning phase solutions in the ``tabletop with a pole'' environment, but it's not able to make improvement in the more complicated ``tabletop with a container'' environment.

\section{Discussion}

This paper presents a fast-reactive motion planning and execution system that can be applied to high-dimensional robotic operations in the presence of uncertainties and generate motion plans that satisfy user-specified chance constraints over collision risks. We first introduce \emph{deterministic Chekov}, an integrated probabilistic roadmap and trajectory optimization framework that features fast online planning for high-dimensional humanoid robots. We evaluate the performance of deterministic Chekov together with five other existing motion planning algorithms in four representative manipulation scenarios, and experiment results show that deterministic Chekov can find high-quality motion plans with a much shorter planning time compared to other existing planners. 

We then describe two different versions of \emph{probabilistic Chekov (p-Chekov)}: the quadrature-based version and the learning-based version. P-Chekov uses deterministic Chekov to generate nominal trajectories, propagates process noises and observation noises along the nominal trajectory in order to estimate the \emph{a priori} probability distribution of robot states, and decomposes the joint chance constraint into allowed collision risk bounds at discrete waypoints. The risk estimation component then predicts the collision risk during execution based on the estimated state distributions, and then compares them with the allocated risk bounds to extract ``conflicts'' where the risk bounds are violated. These conflicts are fed back to deterministic Chekov to guide it to generated safer nominal trajectories. After resolving all the conflicts, the solution trajectory is passed into execution phase and an \emph{Iterative Risk Allocation (IRA)} component will improve the solution through reallocating risk bounds. The only difference between the two versions of p-Chekov is the risk estimation component. Quadrature-based p-Chekov samples from the estimated state distribution according to a quadrature rule during online planning and predicts the collision risk through numerical integration, whereas learning-based p-Chekov samples from the configuration space offline to learn the risk distribution structure and uses the pre-trained regression models to make online risk predictions. Empirical results show that most test cases whose chance constraints are violated by deterministic Chekov's solutions can satisfy their constraints through p-Chekov planning phase's solutions with only a small increase in their trajectory lengths. Through the IRA component, a lot of test cases are able to shorten their trajectory from planning phase solutions. With the partially-observable end-effector pose observation model, quadrature-based p-Chekov shows a slow responding time during the online planning phase, which restricts its application in time-sensitive planning scenarios that require fast reaction. In contrast, in both environments, learning-based Chekov takes only about 6-8~s in the chance-constraint satisfied cases with end-effector observation model, which is a relatively satisfying time for many practical online chance-constrained planning tasks.

Although learning-based p-Chekov shows strong risk-reduction ability and can effectively generate high-quality trajectories that satisfy the chance constraints, it sometimes still spent a long time to search for feasible plans in difficult test cases, and the plans they eventually found often violate the constraints. An interesting future work direction would be to quickly and effectively identify these infeasible test cases where the chance constraint can't be satisfied even though their start and goal poses are not highly risky. In this way, p-Chekov won't need to waste a long time trying to search for feasible solutions for infeasible cases. Another potential future work direction is to incorporate online obstacle avoidance~\citep{park2018fast, orton2019improving, li2019safe} into p-Chekov so that it can handle dynamic obstacles in the execution environment. Additionally, real robot experiments with raw sensor data are also necessary before p-Chekov can be deployed in real-world applications.

\bibliographystyle{SageH}
\bibliography{IJRR}

\begin{thebibliography}{74}
\providecommand{\natexlab}[1]{#1}
\providecommand{\url}[1]{\texttt{#1}}
\providecommand{\urlprefix}{URL }
\expandafter\ifx\csname urlstyle\endcsname\relax
  \providecommand{\doi}[1]{DOI:\discretionary{}{}{}#1}\else
  \providecommand{\doi}{DOI:\discretionary{}{}{}\begingroup
  \urlstyle{rm}\Url}\fi

\bibitem[{Abadi et~al.(2015)Abadi, Agarwal, Barham, Brevdo, Chen, Citro,
  Corrado, Davis, Dean, Devin, Ghemawat, Goodfellow, Harp, Irving, Isard, Jia,
  Jozefowicz, Kaiser, Kudlur, Levenberg, Man\'{e}, Monga, Moore, Murray, Olah,
  Schuster, Shlens, Steiner, Sutskever, Talwar, Tucker, Vanhoucke, Vasudevan,
  Vi\'{e}gas, Vinyals, Warden, Wattenberg, Wicke, Yu and
  Zheng}]{tensorflow2015-whitepaper}
Abadi M, Agarwal A, Barham P, Brevdo E, Chen Z, Citro C, Corrado GS, Davis A,
  Dean J, Devin M, Ghemawat S, Goodfellow I, Harp A, Irving G, Isard M, Jia Y,
  Jozefowicz R, Kaiser L, Kudlur M, Levenberg J, Man\'{e} D, Monga R, Moore S,
  Murray D, Olah C, Schuster M, Shlens J, Steiner B, Sutskever I, Talwar K,
  Tucker P, Vanhoucke V, Vasudevan V, Vi\'{e}gas F, Vinyals O, Warden P,
  Wattenberg M, Wicke M, Yu Y and Zheng X (2015) {TensorFlow}: Large-scale
  machine learning on heterogeneous systems.
\newblock \urlprefix\url{https://www.tensorflow.org/}.
\newblock Software available from tensorflow.org.

\bibitem[{Abramowitz and Stegun(1964)}]{abramowitz1964handbook}
Abramowitz M and Stegun IA (1964) \emph{Handbook of mathematical functions:
  with formulas, graphs, and mathematical tables}, volume~55.
\newblock Courier Corporation.

\bibitem[{Alterovitz et~al.(2007)Alterovitz, Sim{\'e}on and
  Goldberg}]{alterovitz2007stochastic}
Alterovitz R, Sim{\'e}on T and Goldberg KY (2007) The stochastic motion
  roadmap: A sampling framework for planning with markov motion uncertainty.
\newblock In: \emph{Robotics: Science and systems}, volume~3. pp. 233--241.

\bibitem[{Arslan and Tsiotras(2015)}]{arslan2015machine}
Arslan O and Tsiotras P (2015) Machine learning guided exploration for
  sampling-based motion planning algorithms.
\newblock In: \emph{2015 IEEE/RSJ International Conference on Intelligent
  Robots and Systems (IROS)}. IEEE, pp. 2646--2652.

\bibitem[{Atramentov and LaValle(2002)}]{atramentov2002efficient}
Atramentov A and LaValle SM (2002) Efficient nearest neighbor searching for
  motion planning.
\newblock In: \emph{Proceedings 2002 IEEE International Conference on Robotics
  and Automation (Cat. No. 02CH37292)}, volume~1. IEEE, pp. 632--637.

\bibitem[{Axelrod et~al.(2018)Axelrod, Kaelbling and
  Lozano-P{\'e}rez}]{axelrod2018provably}
Axelrod B, Kaelbling LP and Lozano-P{\'e}rez T (2018) Provably safe robot
  navigation with obstacle uncertainty.
\newblock \emph{The International Journal of Robotics Research} 37(13-14):
  1760--1774.

\bibitem[{Bellman(1957)}]{bellman1957dynamic}
Bellman RE (1957) Dynamic programming .

\bibitem[{Bertsekas et~al.(1995)Bertsekas, Bertsekas, Bertsekas and
  Bertsekas}]{bertsekas1995dynamic}
Bertsekas DP, Bertsekas DP, Bertsekas DP and Bertsekas DP (1995) \emph{Dynamic
  programming and optimal control}, volume~1.
\newblock Athena scientific Belmont, MA.

\bibitem[{Blackmore et~al.(2006)Blackmore, Li and
  Williams}]{blackmore2006probabilistic}
Blackmore L, Li H and Williams B (2006) A probabilistic approach to optimal
  robust path planning with obstacles.
\newblock In: \emph{American Control Conference, 2006}. IEEE, pp. 7--pp.

\bibitem[{Blackmore et~al.(2010)Blackmore, Ono, Bektassov and
  Williams}]{blackmore2010probabilistic}
Blackmore L, Ono M, Bektassov A and Williams BC (2010) A probabilistic
  particle-control approximation of chance-constrained stochastic predictive
  control.
\newblock \emph{IEEE transactions on Robotics} 26(3): 502--517.

\bibitem[{Bohlin and Kavraki(2000)}]{bohlin2000path}
Bohlin R and Kavraki LE (2000) Path planning using lazy prm.
\newblock In: \emph{Robotics and Automation, 2000. Proceedings. ICRA'00. IEEE
  International Conference on}, volume~1. IEEE, pp. 521--528.

\bibitem[{Bry and Roy(2011)}]{bry2011rapidly}
Bry A and Roy N (2011) Rapidly-exploring random belief trees for motion
  planning under uncertainty.
\newblock In: \emph{Robotics and Automation (ICRA), 2011 IEEE International
  Conference on}. IEEE, pp. 723--730.

\bibitem[{Burlet et~al.(2004)Burlet, Aycard and Fraichard}]{burlet2004robust}
Burlet J, Aycard O and Fraichard T (2004) Robust motion planning using markov
  decision processes and quadtree decomposition.
\newblock In: \emph{Robotics and Automation, 2004. Proceedings. ICRA'04. 2004
  IEEE International Conference on}, volume~3. IEEE, pp. 2820--2825.

\bibitem[{Campana et~al.(2015)Campana, Lamiraux and
  Laumond}]{campana2015simple}
Campana M, Lamiraux F and Laumond JP (2015) A simple path optimization method
  for motion planning .

\bibitem[{Chen et~al.(2017)Chen, Rickert and Knoll}]{chen2017motion}
Chen C, Rickert M and Knoll A (2017) Motion planning under perception and
  control uncertainties with space exploration guided heuristic search.
\newblock In: \emph{2017 IEEE Intelligent Vehicles Symposium (IV)}. IEEE, pp.
  712--718.

\bibitem[{Chollet et~al.(2015)}]{chollet2015keras}
Chollet F et~al. (2015) Keras.
\newblock \url{https://keras.io}.

\bibitem[{Choset(2005)}]{choset2005principles}
Choset HM (2005) \emph{Principles of robot motion: theory, algorithms, and
  implementation}.
\newblock MIT press.

\bibitem[{Cohen et~al.(2010)Cohen, Chitta and Likhachev}]{cohen2010search}
Cohen BJ, Chitta S and Likhachev M (2010) Search-based planning for
  manipulation with motion primitives.
\newblock In: \emph{Robotics and Automation (ICRA), 2010 IEEE International
  Conference on}. IEEE, pp. 2902--2908.

\bibitem[{Dai et~al.(2018)Dai, Orton, Schaffert, Hofmann and
  Williams}]{dai2018improving}
Dai S, Orton M, Schaffert S, Hofmann A and Williams BC (2018) Improving
  trajectory optimization using a roadmap framework.
\newblock In: \emph{Proceedings of the 2018 IEEE/RSJ International Conference
  on Intelligent Robots and Systems (IROS)}.

\bibitem[{Dai et~al.(2019)Dai, Schaffert, Jasour, Hofmann and
  Williams}]{dai2019chance}
Dai S, Schaffert S, Jasour A, Hofmann A and Williams BC (2019) Chance
  constrained motion planning for high-dimensional robots.
\newblock In: \emph{Proceedings of the 2019 IEEE/RSJ International Conference
  on Robotics and Automation (ICRA)}.

\bibitem[{Eaton(1983)}]{eaton1983multivariate}
Eaton ML (1983) Multivariate statistics: a vector space approach.
\newblock \emph{JOHN WILEY \& SONS, INC., 605 THIRD AVE., NEW YORK, NY 10158,
  USA, 1983, 512} .

\bibitem[{Gelb(1974)}]{gelb1974applied}
Gelb A (1974) \emph{Applied optimal estimation}.
\newblock MIT press.

\bibitem[{Ha et~al.(2018)Ha, Chae and Choi}]{ha2018approximate}
Ha JS, Chae HJ and Choi HL (2018) Approximate inference-based motion planning
  by learning and exploiting low-dimensional latent variable models.
\newblock \emph{IEEE Robotics and Automation Letters} 3(4): 3892--3899.

\bibitem[{Hildebrand(1987)}]{hildebrand1987introduction}
Hildebrand FB (1987) \emph{Introduction to numerical analysis}.
\newblock Courier Corporation.

\bibitem[{Hoeffding et~al.(1948)Hoeffding, Robbins
  et~al.}]{hoeffding1948central}
Hoeffding W, Robbins H et~al. (1948) The central limit theorem for dependent
  random variables.
\newblock \emph{Duke Mathematical Journal} 15(3): 773--780.

\bibitem[{Ichter et~al.(2018)Ichter, Harrison and Pavone}]{ichter2018learning}
Ichter B, Harrison J and Pavone M (2018) Learning sampling distributions for
  robot motion planning.
\newblock In: \emph{2018 IEEE International Conference on Robotics and
  Automation (ICRA)}. IEEE, pp. 7087--7094.

\bibitem[{Janson et~al.(2018)Janson, Schmerling and Pavone}]{janson2018monte}
Janson L, Schmerling E and Pavone M (2018) Monte carlo motion planning for
  robot trajectory optimization under uncertainty.
\newblock In: \emph{Robotics Research}. Springer, pp. 343--361.

\bibitem[{Kalakrishnan et~al.(2011)Kalakrishnan, Chitta, Theodorou, Pastor and
  Schaal}]{kalakrishnan2011stomp}
Kalakrishnan M, Chitta S, Theodorou E, Pastor P and Schaal S (2011) Stomp:
  Stochastic trajectory optimization for motion planning.
\newblock In: \emph{Robotics and Automation (ICRA), 2011 IEEE International
  Conference on}. IEEE, pp. 4569--4574.

\bibitem[{Karaman and Frazzoli(2011)}]{karaman2011sampling}
Karaman S and Frazzoli E (2011) Sampling-based algorithms for optimal motion
  planning.
\newblock \emph{The International Journal of Robotics Research} 30(7):
  846--894.

\bibitem[{Kingma and Ba(2014)}]{kingma2014adam}
Kingma DP and Ba J (2014) Adam: A method for stochastic optimization.
\newblock \emph{arXiv preprint arXiv:1412.6980} .

\bibitem[{Koenig and Likhachev(2005)}]{koenig2005fast}
Koenig S and Likhachev M (2005) Fast replanning for navigation in unknown
  terrain.
\newblock \emph{IEEE Transactions on Robotics} 21(3): 354--363.

\bibitem[{Kurniawati et~al.(2008)Kurniawati, Hsu and
  Lee}]{kurniawati2008sarsop}
Kurniawati H, Hsu D and Lee WS (2008) Sarsop: Efficient point-based pomdp
  planning by approximating optimally reachable belief spaces.
\newblock In: \emph{Robotics: Science and systems}, volume 2008. Zurich,
  Switzerland.

\bibitem[{LaValle(1998)}]{lavalle1998rapidly}
LaValle SM (1998) Rapidly-exploring random trees: A new tool for path planning
  .

\bibitem[{Lee et~al.(2013)Lee, Duan, Patil, Schulman, McCarthy, Van Den~Berg,
  Goldberg and Abbeel}]{lee2013sigma}
Lee A, Duan Y, Patil S, Schulman J, McCarthy Z, Van Den~Berg J, Goldberg K and
  Abbeel P (2013) Sigma hulls for gaussian belief space planning for imprecise
  articulated robots amid obstacles.
\newblock In: \emph{2013 IEEE/RSJ International Conference on Intelligent
  Robots and Systems}. IEEE, pp. 5660--5667.

\bibitem[{Lenz et~al.(2015)Lenz, Rickert and Knoll}]{lenz2015heuristic}
Lenz D, Rickert M and Knoll A (2015) Heuristic search in belief space for
  motion planning under uncertainties.
\newblock In: \emph{2015 IEEE/RSJ International Conference on Intelligent
  Robots and Systems (IROS)}. IEEE, pp. 2659--2665.

\bibitem[{Li and Shah(2019)}]{li2019safe}
Li S and Shah JA (2019) Safe and efficient high dimensional motion planning in
  space-time with time parameterized prediction.
\newblock In: \emph{Proceedings of the 2019 IEEE/RSJ International Conference
  on Robotics and Automation (ICRA)}.

\bibitem[{Liaw et~al.(2002)Liaw, Wiener et~al.}]{liaw2002classification}
Liaw A, Wiener M et~al. (2002) Classification and regression by randomforest.
\newblock \emph{R news} 2(3): 18--22.

\bibitem[{Liu and Ang(2014)}]{liu2014incremental}
Liu W and Ang MH (2014) Incremental sampling-based algorithm for risk-aware
  planning under motion uncertainty.
\newblock In: \emph{Robotics and Automation (ICRA), 2014 IEEE International
  Conference on}. IEEE, pp. 2051--2058.

\bibitem[{Luders et~al.(2010)Luders, Kothari and How}]{luders2010chance}
Luders B, Kothari M and How J (2010) Chance constrained rrt for probabilistic
  robustness to environmental uncertainty.
\newblock In: \emph{AIAA guidance, navigation, and control conference}. p.
  8160.

\bibitem[{Luders et~al.(2013)Luders, Karaman and How}]{luders2013robust}
Luders BD, Karaman S and How JP (2013) Robust sampling-based motion planning
  with asymptotic optimality guarantees.
\newblock In: \emph{AIAA Guidance, Navigation, and Control (GNC) Conference}.
  p. 5097.

\bibitem[{Luenberger(1979)}]{luenberger1979introduction}
Luenberger DG (1979) \emph{Introduction to dynamic systems: theory, models, and
  applications}, volume~1.
\newblock Wiley New York.

\bibitem[{Luna et~al.(2013)Luna, {\c{S}}ucan, Moll and
  Kavraki}]{luna2013anytime}
Luna R, {\c{S}}ucan IA, Moll M and Kavraki LE (2013) Anytime solution
  optimization for sampling-based motion planning.
\newblock In: \emph{Robotics and Automation (ICRA), 2013 IEEE International
  Conference on}. IEEE, pp. 5068--5074.

\bibitem[{Luo et~al.(2019)Luo, Bai, Hsu and Lee}]{luo2019importance}
Luo Y, Bai H, Hsu D and Lee WS (2019) Importance sampling for online planning
  under uncertainty.
\newblock \emph{The International Journal of Robotics Research} 38(2-3):
  162--181.

\bibitem[{Mukadam et~al.(2018)Mukadam, Dong, Yan, Dellaert and
  Boots}]{mukadam2018continuous}
Mukadam M, Dong J, Yan X, Dellaert F and Boots B (2018) Continuous-time
  gaussian process motion planning via probabilistic inference.
\newblock \emph{The International Journal of Robotics Research} 37(11):
  1319--1340.

\bibitem[{Murphy(2012)}]{murphy2012machine}
Murphy KP (2012) \emph{Machine learning: a probabilistic perspective}.
\newblock MIT press.

\bibitem[{Ono and Williams(2008{\natexlab{a}})}]{ono2008efficient}
Ono M and Williams B (2008{\natexlab{a}}) An efficient motion planning
  algorithm for stochastic dynamic systems with constraints on probability of
  failure.

\bibitem[{Ono and Williams(2008{\natexlab{b}})}]{ono2008iterative}
Ono M and Williams BC (2008{\natexlab{b}}) Iterative risk allocation: A new
  approach to robust model predictive control with a joint chance constraint.
\newblock In: \emph{Decision and Control, 2008. CDC 2008. 47th IEEE Conference
  on}. IEEE, pp. 3427--3432.

\bibitem[{Ono et~al.(2013)Ono, Williams and Blackmore}]{ono2013probabilistic}
Ono M, Williams BC and Blackmore L (2013) Probabilistic planning for continuous
  dynamic systems under bounded risk.
\newblock \emph{Journal of Artificial Intelligence Research} 46: 511--577.

\bibitem[{Orton et~al.(2019)Orton, Dai, Schaffert, Hofmann and
  Williams}]{orton2019improving}
Orton M, Dai S, Schaffert S, Hofmann A and Williams BC (2019) Improving
  incremental planning performance through overlapping replanning and
  execution.
\newblock In: \emph{Proceedings of the 2019 IEEE/RSJ International Conference
  on Robotics and Automation (ICRA)}.

\bibitem[{Owen(2014)}]{owen2014monte}
Owen A (2014) Monte carlo theory, methods and examples (book draft).

\bibitem[{Pan et~al.(2012)Pan, Chitta and Manocha}]{pan2012fcl}
Pan J, Chitta S and Manocha D (2012) Fcl: A general purpose library for
  collision and proximity queries.
\newblock In: \emph{2012 IEEE International Conference on Robotics and
  Automation}. IEEE, pp. 3859--3866.

\bibitem[{Pan et~al.(2017)Pan, Chitta and Manocha}]{pan2017probabilistic}
Pan J, Chitta S and Manocha D (2017) Probabilistic collision detection between
  noisy point clouds using robust classification.
\newblock In: \emph{Robotics Research}. Springer, pp. 77--94.

\bibitem[{Pan and Manocha(2016)}]{pan2016fast}
Pan J and Manocha D (2016) Fast probabilistic collision checking for
  sampling-based motion planning using locality-sensitive hashing.
\newblock \emph{The International Journal of Robotics Research} 35(12):
  1477--1496.

\bibitem[{Park et~al.(2012)Park, Pan and Manocha}]{park2012itomp}
Park C, Pan J and Manocha D (2012) Itomp: Incremental trajectory optimization
  for real-time replanning in dynamic environments.
\newblock In: \emph{ICAPS}.

\bibitem[{Park et~al.(2018)Park, Park and Manocha}]{park2018fast}
Park C, Park JS and Manocha D (2018) Fast and bounded probabilistic collision
  detection for high-dof trajectory planning in dynamic environments.
\newblock \emph{IEEE Transactions on Automation Science and Engineering} 15(3):
  980--991.

\bibitem[{Park et~al.(2015)Park, Rabe, Sharma, Scheurer, Zimmermann and
  Manocha}]{park2015parallel}
Park C, Rabe F, Sharma S, Scheurer C, Zimmermann UE and Manocha D (2015)
  Parallel cartesian planning in dynamic environments using constrained
  trajectory planning.
\newblock In: \emph{Humanoid Robots (Humanoids), 2015 IEEE-RAS 15th
  International Conference on}. IEEE, pp. 983--990.

\bibitem[{Patil et~al.(2014)Patil, Duan, Schulman, Goldberg and
  Abbeel}]{patil2014gaussian}
Patil S, Duan Y, Schulman J, Goldberg K and Abbeel P (2014) Gaussian belief
  space planning with discontinuities in sensing domains.
\newblock In: \emph{2014 IEEE International Conference on Robotics and
  Automation (ICRA)}. IEEE, pp. 6483--6490.

\bibitem[{Patil et~al.(2015)Patil, Kahn, Laskey, Schulman, Goldberg and
  Abbeel}]{patil2015scaling}
Patil S, Kahn G, Laskey M, Schulman J, Goldberg K and Abbeel P (2015) Scaling
  up gaussian belief space planning through covariance-free trajectory
  optimization and automatic differentiation.
\newblock In: \emph{Algorithmic foundations of robotics XI}. Springer, pp.
  515--533.

\bibitem[{Patil et~al.(2012)Patil, Van Den~Berg and
  Alterovitz}]{patil2012estimating}
Patil S, Van Den~Berg J and Alterovitz R (2012) Estimating probability of
  collision for safe motion planning under gaussian motion and sensing
  uncertainty.
\newblock In: \emph{Robotics and Automation (ICRA), 2012 IEEE International
  Conference on}. IEEE, pp. 3238--3244.

\bibitem[{Pedregosa et~al.(2011)Pedregosa, Varoquaux, Gramfort, Michel,
  Thirion, Grisel, Blondel, Prettenhofer, Weiss, Dubourg, Vanderplas, Passos,
  Cournapeau, Brucher, Perrot and Duchesnay}]{scikit-learn}
Pedregosa F, Varoquaux G, Gramfort A, Michel V, Thirion B, Grisel O, Blondel M,
  Prettenhofer P, Weiss R, Dubourg V, Vanderplas J, Passos A, Cournapeau D,
  Brucher M, Perrot M and Duchesnay E (2011) Scikit-learn: Machine learning in
  {P}ython.
\newblock \emph{Journal of Machine Learning Research} 12: 2825--2830.

\bibitem[{Pfeiffer et~al.(2017)Pfeiffer, Schaeuble, Nieto, Siegwart and
  Cadena}]{pfeiffer2017perception}
Pfeiffer M, Schaeuble M, Nieto J, Siegwart R and Cadena C (2017) From
  perception to decision: A data-driven approach to end-to-end motion planning
  for autonomous ground robots.
\newblock In: \emph{2017 IEEE International Conference on Robotics and
  Automation (ICRA)}. IEEE, pp. 1527--1533.

\bibitem[{Rasmussen and Williams(2006)}]{rasmussen2006gaussian}
Rasmussen C and Williams C (2006) \emph{Gaussian Processes for Machine
  Learning}.
\newblock MIT Press.

\bibitem[{RethinkRobotics(2012)}]{BaxterRobot}
RethinkRobotics (2012) Baxter.
\newblock \urlprefix\url{http://www.rethinkrobotics.com/baxter/}.

\bibitem[{Schulman et~al.(2014)Schulman, Duan, Ho, Lee, Awwal, Bradlow, Pan,
  Patil, Goldberg and Abbeel}]{schulman2014motion}
Schulman J, Duan Y, Ho J, Lee A, Awwal I, Bradlow H, Pan J, Patil S, Goldberg K
  and Abbeel P (2014) Motion planning with sequential convex optimization and
  convex collision checking.
\newblock \emph{The International Journal of Robotics Research} 33(9):
  1251--1270.

\bibitem[{Schulman et~al.(2013)Schulman, Ho, Lee, Awwal, Bradlow and
  Abbeel}]{schulman2013finding}
Schulman J, Ho J, Lee AX, Awwal I, Bradlow H and Abbeel P (2013) Finding
  locally optimal, collision-free trajectories with sequential convex
  optimization.
\newblock In: \emph{Robotics: science and systems}, volume~9. Citeseer, pp.
  1--10.

\bibitem[{Stentz(1994)}]{stentz1994optimal}
Stentz A (1994) Optimal and efficient path planning for partially-known
  environments.
\newblock In: \emph{Robotics and Automation, 1994. Proceedings., 1994 IEEE
  International Conference on}. IEEE, pp. 3310--3317.

\bibitem[{Sun et~al.(2015)Sun, Patil and Alterovitz}]{sun2015high}
Sun W, Patil S and Alterovitz R (2015) High-frequency replanning under
  uncertainty using parallel sampling-based motion planning.
\newblock \emph{IEEE Transactions on Robotics} 31(1): 104--116.

\bibitem[{Sun et~al.(2016)Sun, Torres, Van Den~Berg and
  Alterovitz}]{sun2016safe}
Sun W, Torres LG, Van Den~Berg J and Alterovitz R (2016) Safe motion planning
  for imprecise robotic manipulators by minimizing probability of collision.
\newblock In: \emph{Robotics Research}. Springer, pp. 685--701.

\bibitem[{Thrun et~al.(2005)Thrun, Burgard and Fox}]{thrun2005probabilistic}
Thrun S, Burgard W and Fox D (2005) \emph{Probabilistic robotics}.
\newblock MIT press.

\bibitem[{Van Den~Berg et~al.(2011)Van Den~Berg, Abbeel and
  Goldberg}]{van2011lqg}
Van Den~Berg J, Abbeel P and Goldberg K (2011) Lqg-mp: Optimized path planning
  for robots with motion uncertainty and imperfect state information.
\newblock \emph{The International Journal of Robotics Research} 30(7):
  895--913.

\bibitem[{Van Den~Berg et~al.(2012)Van Den~Berg, Patil and
  Alterovitz}]{van2012motion}
Van Den~Berg J, Patil S and Alterovitz R (2012) Motion planning under
  uncertainty using iterative local optimization in belief space.
\newblock \emph{The International Journal of Robotics Research} 31(11):
  1263--1278.

\bibitem[{Wang et~al.(2016)Wang, Chen, Lau and Ren}]{wang2016motion}
Wang H, Chen J, Lau HY and Ren H (2016) Motion planning based on learning from
  demonstration for multiple-segment flexible soft robots actuated by
  electroactive polymers.
\newblock \emph{IEEE Robotics and Automation Letters} 1(1): 391--398.

\bibitem[{Xiao et~al.(2020)Xiao, Dufek and Murphy}]{xiao2020robot}
Xiao X, Dufek J and Murphy RR (2020) Robot risk-awareness by formal risk
  reasoning and planning.
\newblock \emph{IEEE Robotics and Automation Letters} 5(2): 2856--2863.

\bibitem[{Zucker et~al.(2013)Zucker, Ratliff, Dragan, Pivtoraiko, Klingensmith,
  Dellin, Bagnell and Srinivasa}]{zucker2013chomp}
Zucker M, Ratliff N, Dragan AD, Pivtoraiko M, Klingensmith M, Dellin CM,
  Bagnell JA and Srinivasa SS (2013) Chomp: Covariant hamiltonian optimization
  for motion planning.
\newblock \emph{The International Journal of Robotics Research} 32(9-10):
  1164--1193.

\end{thebibliography}

\end{document}